\documentclass[10pt,twocolumn,letterpaper]{article}

\usepackage{cvpr}
\usepackage{times}
\usepackage{epsfig}
\usepackage{graphicx}
\usepackage{amsmath}
\usepackage{amssymb}
\usepackage{calc}

\usepackage{multirow}

\graphicspath{{fig/}}

\usepackage[table]{xcolor}
\usepackage{subcaption}
\usepackage{soul}
\usepackage{booktabs}
\usepackage{enumitem}
\usepackage{overpic}
\setlist{nosep}

\makeatletter
\@namedef{ver@everyshi.sty}{}
\makeatother
\usepackage{tikz}
\usetikzlibrary{calc}

\newif\ifdraft
\draftfalse
\drafttrue

\definecolor{orange}{rgb}{1,0.5,0}
\definecolor{violet}{RGB}{70,0,170}

\ifdraft
 \newcommand{\PF}[1]{{\color{red}{\bf PF: #1}}}
 
 \newcommand{\LC}[1]{{\color{blue}{\bf LC: #1}}}
 
 \newcommand{\MK}[1]{{\color{green}{\bf MK: #1}}}
 
\else
 \newcommand{\PF}[1]{}
 
 \newcommand{\LC}[1]{}
 
 \newcommand{\MK}[1]{}
 
\fi

\newcommand{\comment}[1]{}
\newcommand{\parag}[1]{\vspace{-3mm}\paragraph{#1}}

\newcommand{\Pre}{P}
\newcommand{\Rec}{R}
\newcommand{\PP}{\mathrm{PP}}
\newcommand{\AP}{\mathrm{AP}}
\newcommand{\TP}{\mathrm{TP}}
\newcommand{\IoU}{\mathrm{IoU}}
\newcommand{\FOne}{\mathrm{F1}}

\newcommand{\SEGM}{{\bf CCQ}}

\newcommand{\TLTS}{{\bf TLTS}}
\newcommand{\APLS}{{\bf APLS}}
\newcommand{\length}{L}
\newcommand{\path}{p}
\newcommand{\gt}{\mathrm{gt}}
\newcommand{\est}{\mathrm{est}}
\newcommand{\gtpath}{{\path}^{\gt}}
\newcommand{\estpath}{{\path}^{\est}}
\newcommand{\paths}{\mathcal{P}}

\newcommand{\JOLD} {{\bf JUNC}}
\newcommand{\JNEW}{{\bf NEWJ}}

\newcommand{\ngt}{n_{\mathrm{gt}}}

\newcommand{\junctionmark}{_{J}}
\newcommand{\jprecision}{{\Pre}\junctionmark}
\newcommand{\jrecall}{{\Rec}\junctionmark}
\newcommand{\jPP}{{\PP}\junctionmark}
\newcommand{\jAP}{{\AP}\junctionmark}
\newcommand{\jTP}{{\TP}\junctionmark}

\newcommand{\FGU}{{F^{-}_\mathrm{gt}}}
\newcommand{\FPU}{{F^{-}_\mathrm{est}}}

\newcommand{\jdist}{d}
\newcommand{\jmaxdist}{\jdist^\mathrm{max}}
\newcommand{\jmcost}{c}

\newcommand{\jcoeff}{\alpha}

\newcommand{\jMatches}{M}

\newcommand{\order}{o}

\newcommand{\PNEW}{{\bf NEWP}}
\newcommand{\pathmark}{_{P}}
\newcommand{\pprecision}{{\Pre}\pathmark}
\newcommand{\precall}{{\Rec}\pathmark}

\newcommand{\pPath}{{\pi}}
\newcommand{\pPaths}{{\Pi}}

\newcommand{\segment}{s}

\newcommand{\Segments}{\mathcal{S}}

\newcommand{\pprob}{\mathrm{P}}

\newcommand{\GOLD}{{\bf GRAPH}}
\newcommand{\GNEW}{{\bf NEWG}}
\newcommand{\graphmark}{_{G}}
\newcommand{\gprecision}{{\Pre}\graphmark}
\newcommand{\grecall}{{\Rec}\graphmark}
\newcommand{\gPP}{{\PP}\graphmark}
\newcommand{\gAP}{{\AP}\graphmark}
\newcommand{\gTP}{{\TP}\graphmark}

\newcommand{\RTracer}{{\it RoadTracer}}
\newcommand{\Segm}{{\it Segmentation}}
\newcommand{\SegPath}{{\it Seg-Path}}
\newcommand{\DRoad}{{\it DeepRoad}}
\newcommand{\RCNN}{{\it RCNNUNet}}
\newcommand{\MultiB}{{\it MultiBranch}}
\newcommand{\LinkN}{{\it LinkNet}}

\usepackage[pagebackref=true,breaklinks=true,letterpaper=true,colorlinks,bookmarks=false]{hyperref}

\cvprfinalcopy 

\ifcvprfinal\fi
\begin{document}

\title{Towards Reliable Evaluation of Road Network Reconstructions}

\author{%
        \noindent\parbox[c]{0.29\textwidth}{\centering Leonardo Citraro} \parbox{0.29\textwidth}{\centering Mateusz Kozi\'nski} \parbox{0.29\textwidth}{\centering Pascal Fua}\\
	{Computer Vision Laboratory, \'Ecole Polytechnique F\'ed\'erale de Lausanne, Switzerland} \\
        \noindent\parbox[c]{0.29\textwidth}{\centering\tt\small leonardo.citraro@epfl.ch} \parbox{0.29\textwidth}{\centering\tt\small mateusz.kozinski@epfl.ch} \parbox{0.29\textwidth}{\centering\tt\small pascal.fua@epfl.ch}
}

\maketitle


\begin{abstract}
Existing performance measures rank delineation algorithms inconsistently, which makes it difficult to decide which one is best in any given situation. We show that these inconsistencies stem from design flaws that make the metrics insensitive to whole classes of errors. To provide more reliable evaluation, we design three new metrics that are far more consistent even though they use very different approaches to comparing ground-truth and reconstructed road networks. We use both synthetic and real data to demonstrate this and advocate the use of these corrected metrics as a tool to gauge future progress. 

\end{abstract}


\section{Introduction}
\label{sec:introduction}

Reconstruction of road networks from aerial images is an old computer vision problem. It has been tackled almost since the inception of the field~\cite{Bajcsy76a,Vanderbrug76,Quam78,Fischler81b}. Yet, it is still open and is addressed by many recent papers~\cite{Mnih10,Mnih13,Cheng17,Mattyus17,Li18f,Bastani18,Mosinska18,Chu19,Batra19,Mosinska19,Yang19}. One roadblock, however, is that the metrics used to measure performance are unreliable and inconsistent. A method that performs well according to one popular metric can perform poorly according to another, which we will demonstrate. This makes interpretation and therefore further progress difficult. 

This situation arises from the fact that the quality of a reconstructed road graph does not only depend on the spatial accuracy of the road centerline predictions but also on the connectivity these centerlines define. The first is relatively easy to measure while the second is much more difficult and there is no generally accepted way of doing so. This is because comparing the predicted topology to the ground truth one amounts to solving a complex graph-matching problem for which no efficient algorithm exists.

\begin{figure}
\includegraphics[trim={0 5.7cm 0 0},clip]{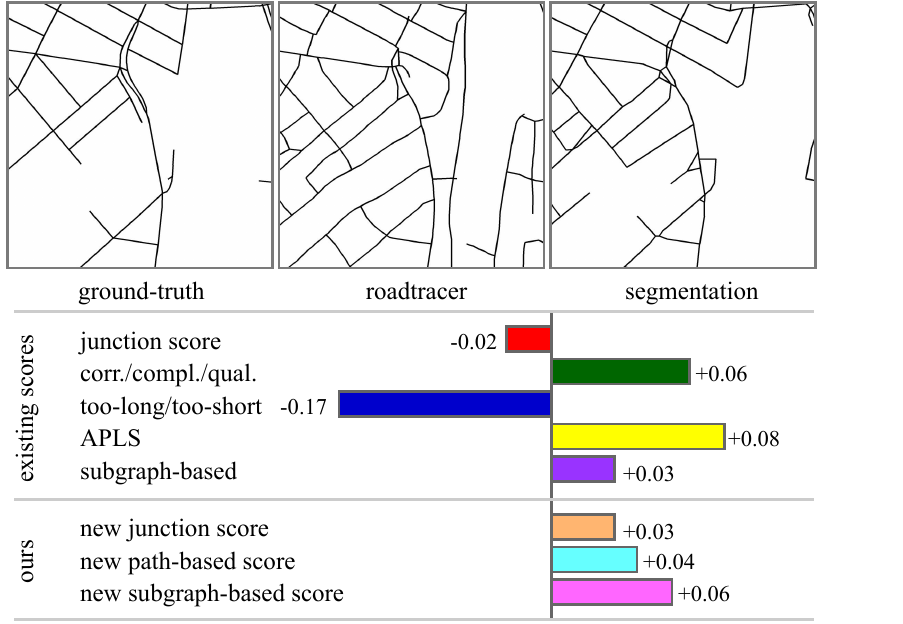}
\vspace{-7mm}
\caption{
\emph{Top}: Crops of a road network of Montreal and its reconstructions from aerial images by Roadtracer~\cite{Bastani18} and Segmentation. 
We show that existing metrics for evaluating such reconstructions are unreliable. We propose more reliable ones.
\emph{Bottom}: Differences of the metrics. Bars that extend to the right favor Segmentation, ones that extend to the left favor Roadtracer~\cite{Bastani18}.
\label{teaser}
}
\includegraphics[trim={0 0.9cm 0 7.8cm},clip]{fig/teaser_rt_seg_split.pdf}
\vspace{-7mm}
\caption{
Injecting a controlled number of errors into ground truth networks reveals insensitivity of existing metrics to certain types of errors 
and helps us design new, sensitive metrics.
}
\end{figure}

Current topology-aware metrics are therefore approximations that fall into three main categories, those that compare the junctions of the two graphs, those that compare the lengths of the paths connecting random pairs of junctions, and those that match small subgraphs. We will show that, unfortunately,  these metrics correlate poorly with the number of topological discrepancies---missed road branches, unwarranted or missing connections---between the two graphs. This is problematic because, in practice, an operator will ultimately have to fix these delineations by hand before they can be used and the number of such mistakes correlates directly with the cost of doing so. 

In this paper, we therefore propose three new topologically-aware metrics that measure much more effectively the true quality of a delineation:
\begin{itemize}

\item {\bf Path-Based.} This metric compares pairs of paths in the two graphs that connect the same endpoints, while enforcing a one-to-one correspondence between the paths in the two graphs so that an edge representing a short road segment in one cannot be associated to two different edges in the other. As a result, it catches errors caused by finding two roads where there is only one or, conversely, missing one of two closely spaced roads. Existing path-based metrics tend to miss such errors.

\item {\bf Junction-Based.} This metric compares topologies of two graphs in terms of the number of connections incident on junctions.  Unlike current ones,  it correctly balances the impact of different kinds of errors, particularly that of road breaks that are rarely handled correctly. 

\item {\bf Graph-Based.} This metric compares small sub-graphs connected to randomly selected center points. Unlike current  ones, it samples these center points in both the predicted and ground truth networks and enforces one-to-one correspondence between the subgraphs' nodes. This makes it sensitive to false positives that current graph-based metrics miss.

\end{itemize}
Not only are our new metrics better at quantifying the number of mistakes that an algorithm makes, they are also consistent with each other. 
This is significant because the three metrics are computed in very different ways. 
Thus, when they all agree, one can be confident of their conclusions. 
Our contributions are therefore
\begin{itemize}
\item An in-depth analysis of existing metrics that exposes their lack of sensitivity to certain types of errors and the resulting lack of consistency when using them to compare different algorithms. 
\item Three new measures, free from this problem, that we advocate for future algorithm evaluation. 
\item A benchmark dataset for systematic evaluation of the sensitivity of such scores to specific types of errors.
\end{itemize}
%


\section{Existing Metrics}
\label{sec:existingMetrics}

Let us assume that a road network is represented by a graph.
Each of its nodes is attributed with a pair of coordinates.
Nodes play a double role: they represent road intersections and outline curvy road segments.
Pairs of nodes are connected with edges that model straight road segments.

We are given a \emph{predicted} graph and a \emph{ground truth} graph, whose similarity we want to assess.
Comparing these two networks that are similar, but not identical, is non-trivial. Doing this in a graph-theoretic way can be viewed as an NP-complete graph matching problem~\cite{Wegener05}. In practice, a number of metrics have been developed for this purpose. They can be classified into four main categories, depending on whether they are pixel-based, junction-based, path-based, or subgraph-based. 
We now review these four classes of existing metrics and argue that they {\it all} ignore particular type of errors. 
We will confirm this in the experiment section.

\subsection{Pixel-Based Metrics}
\label{sec:pixelBased}

Road delineation can be understood as foreground/background segmentation problem. 
The quality of the segmentation can be evaluated in terms of  \emph{precision} $\Pre=\frac{| \TP | } {| \PP |}$ and \emph{recall} $\Rec=\frac{|\TP|}{| \AP |}$, where  $\PP$ is the set of pixels predicted to be foreground, $\AP$ is the set of pixel labeled as foreground, and $\TP  = \PP \cap \AP$.  
When a single number is preferred, either the \emph{intersection-over-union} $\IoU=\frac{\TP}{\PP\cup\AP}$, or the \emph{f1-score} $\FOne=2/(\frac{1}{\Pre}+\frac{1}{\Rec}) $ is used.

\parag{Correctness/Completeness/Quality (\SEGM{}).} 
To account for the fact that the position of the pixels estimated to be foreground might be slightly off, the definitions of precision and recall were relaxed in~\cite{Wiedemann98} to allow small shifts in pixel locations.
The relaxed precision was called \emph{correctness}, relaxed recall \emph{completeness} and \emph{quality} was the equivalent of intersection over union.

\parag{Discussion.}
\SEGM{} are adequate to gauge segmentation quality but do not capture connectivity of the foreground pixels.
This makes them insensitive to topological errors, which is why the path- and junction-based metrics described below have become popular. 

\subsection{Path-Based Metrics}
\label{sec:pathBased}

The idea behind path-based metrics is that if two graphs are similar, so should paths connecting any pair of their nodes via a sequence of edges. 
Edges that appear in one graph and not the other result in measurably different paths. 
There are two main ways to measure such differences. 

\parag{Too Long / Too Short (\TLTS{}).} In~\cite{Wegner13}, it was proposed to compare the length of the shortest path between randomly-chosen but corresponding pairs of nodes in the predicted and ground-truth networks. A path in the predicted graph is classified as \emph{correct} if its length is within 5\% of that of the path in the ground-truth graph, and as \emph{too-long}, or \emph{too-short} otherwise.
A path is marked \emph{infeasible} if its end points are not connected in the other network.  
The percentage of \emph{correct} paths is used to assess the quality of a delineation and the other percentages serve to characterize the errors. 

\parag{Average Path Length Similarity (\APLS{}).} 
The alternative to counting too long/short paths is aggregating path differences.
This has been proposed first for evaluating road network reconstructions from GPS tracks~\cite{Karagiorgou12,Ahmed15a} and, more recently, for image-based reconstructions~\cite{VanEtten18}, in form of the Average Path Length Similarity score
\begin{equation}
1-\frac{1}{|\paths|}\sum_{(\gtpath,\estpath)\in\paths}\min\left\{1, \frac{|\length_{\gtpath}-\length_{\estpath}|}{\length_{\gtpath}}\right\} \; ,
\end{equation}
where $\length_{\estpath}$ and $\length_{\gtpath}$ are the path length in the estimated and ground truth graphs, respectively. 
The set $\paths$ is obtained by sampling pairs of points in one graph, retrieving the corresponding pairs in the other graph, and computing the shortest paths between them.

\paragraph{Discussion.}

\TLTS{} and \APLS{} are better at capturing topological differences than the pixel-based scores, but they suffer from a major flaw. 
Since paths are sampled and matched independently, 
multiple paths from a graph can be matched to a single path in the other graph.
This makes the scores insensitive to errors that consist in predicting one road where many closely spaced roads exist and to predicting more than one road where there is just one, as shown in Fig.~\ref{fig:new_path_expl}.

\subsection{Junction-Based Metric (\JOLD{})}
\label{sec:junctionBased}

The path-based metrics capture the topological similarity indirectly. 
A more direct approach, proposed in~\cite{Bastani18}, 
relies on comparing the degree of corresponding nodes with at least three incident edges, called junctions. 
The correspondences are established greedily by matching closest nodes.

For each matched ground-truth junction $v$, the per-junction recall $f_{v,\text{correct}}$ is taken to be the fraction of edges incident on $v$ that are also captured around the corresponding predicted junction. 
For each predicted junction $u$, its false discovery rate (one-minus-precision) $f_{u,\text{error}}$ is the fraction of its edges that do not appear around the corresponding ground truth junction. 
For unmatched junctions, $f_{v,\text{correct}}=0$ and $f_{u,\text{error}}=1$, respectively. These per-junction scores are then aggregated 
\begin{align}
n_\text{correct} &=\sum_v f_{v,\text{correct}} \;, & & & n_\text{error}&=\sum_u f_{u,\text{error}} \; , \nonumber \\
\hspace{0.1cm} F_{\text{correct}} &= \frac{n_\text{correct}}{\ngt} \; , & \text{and} & & 
F_\text{error} &= \frac{n_\text{error}}{n_\text{error}+n_\text{correct}} \; , \hspace{0.1cm} \nonumber
\end{align}
where $\ngt$ is the number of ground-truth junctions. 

\paragraph{Discussion}
The main issue with \JOLD{} is that it only accounts for nodes with three or more incident edges.
This disregards what happens at road end points and makes the metric insensitive to interruptions in predicted networks. 
Moreover, the score penalizes missing $k-2$ out of $k$ incident edges more than any other number, a choice that is difficult to justify in terms of any possible application. 
The top of Fig.~\ref{fig:new_junction_expl} illustrates this problem: an edge is missing from a junction with three incident edges, which results in $n_\text{correct}=\frac{0}{3}$ instead of $\frac{2}{3}$. 

\subsection{Subgraph-Based Metric (\GOLD{})}
\label{sec:subgraphBased}

In~\cite{Biagioni12}, it is suggested to compare the sets of locations accessible by travelling a predefined distance away from corresponding points in two graphs.
To this end, a starting location is randomly selected in the ground truth network, and its closest point in the predicted network is identified. 
Then, local subgraphs are extracted by exploring the graphs away from the starting locations.
To compute the score, virtual control points are inserted at regular intervals into these subgraphs.
A control point is considered matched, or true positive, if it lies sufficiently close to some control point in the other network.
Unmatched control points in the predicted, and annotated subgraphs are treated as false positives and false negatives, respectively. 
Sampling and matching of local subgraphs is repeated many times, and precision and recall are computed from the total counts of matched and unmatched control points.

\paragraph{Discussion}
As the starting point is always sampled from the ground truth network,
the false positive roads that are sufficiently far from any ground truth road are not covered by control points.
In consequence, \GOLD{} is not sensitive to such errors.
Moreover, since multiple control points of the ground truth network can be matched to the same control point of the prediction, 
errors consisting in predicting just one instead of two closely spaced roads go unnoticed.
%


\section{New Metrics}
\label{sec:newMetrics}

In Section~\ref{sec:existingMetrics}, we identified weaknesses of existing metrics that make them insensitive to whole classes of errors. 
Here, we introduce new metrics, inspired by current ones, but without their blind spots.
We will demonstrate the effectiveness of all the scores in Section~\ref{sec:experiments}.

\subsection{Path-Based Metric (\PNEW{})}
\label{sec:pathBasedNew}


\begin{figure}[h!]
  \centering
  \setlength{\tabcolsep}{0pt}
  \begin{tabular}{@{}l@{}}
    \includegraphics[trim={0, 6.0cm, 0, 0},clip]{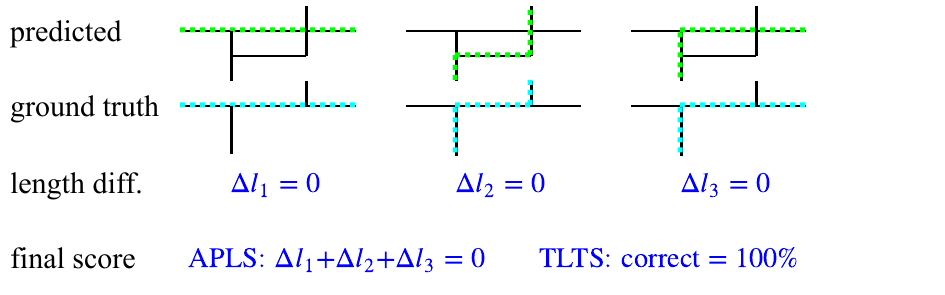} \\
    \multicolumn{1}{>{\centering\arraybackslash}p{\columnwidth}}{\small (a) The existing path length statistics \TLTS{}, \APLS{}. Both sampled paths (green) and their matches (cyan) overlap. As a result, the scores do not capture the difference between the networks.
} \\
    \\
    \includegraphics[trim={0, 1.2cm, 0, 3.8cm},clip]{new_path_expl_v2.pdf} \\
    \multicolumn{1}{>{\centering\arraybackslash}p{\columnwidth}}{\small (b) Our new path-based score (\PNEW{}). Paths do not overlap and the score captures the difference between the networks. 
} \\
  \end{tabular}
  \caption{
A comparison of (a) the existing path-based statistics and (b) our new path score.
Three paths are sampled from the predicted network (overlayed in green), and matched to the ground truth network (the matching chains are highlighted in cyan). 
The unmatched parts of the paths are highlighted in red. 
Removed parts of the networks are shown in dotted gray.
See section~\ref{sec:pathBased} for the definition of the \APLS{} and \TLTS{} and section~\ref{sec:pathBasedNew} for \PNEW{}.
\label{fig:new_path_expl} 
 }
\end{figure}

In section~\ref{sec:pathBased} we argued that the \TLTS{}~\cite{Wegner13} and \APLS{}~\cite{VanEtten18} are insensitive to erroneous paths when they are close to valid ones. 
We therefore introduce a new path-based metric \PNEW{}. 
It involves computing $\precall$, which can loosely be interpreted as path recall,  and $\pprecision$, which plays the role of path precision.
In contrast to earlier metrics, we do not sample or match paths independently. 
Instead, we ensure that no two paths sampled from a graph share the same edges, 
and that any two sampled paths are matched to two disjoint sets of edges in the other graph. 
This makes $\pprecision$ sensitive to the presence of many similar predicted paths, only one of which is valid. 

More precisely, we developed  an iterative path sampling and matching scheme.
To compute recall,  
in each iteration we sample a path from the ground truth network and match it to the predicted network. 
Using the match, we compute our measure of connectivity as described in the next paragraph.
To ensure that no two paths share the same edges, we remove the sampled path from the ground-truth network. 
To guarantee that no edge from the predicted network is matched to two different paths, we also remove the matched edges from the predicted network.
We iterate until one of the networks has no more edges. 
Fig.~\ref{fig:new_path_expl} illustrates this behavior. 
Precision is computed similarly, but the roles of the networks are exchanged.

Matching a path $\pPath$ to a graph consists in identifying a chain of connected nodes with a minimum Euclidean distance to the path.
However, if the path is not connected in the graph, it may be mapped to more than one connected chain, where each chain is disconnected from the others. 
This mapping induces a partitioning of the path $\pPath$ into segments $\Segments(\pPath)$, such that each $\segment\in\Segments(\pPath)$ maps to a different chain. 
If the $\pPath$ exists in the graph without disconnections, $\Segments(\pPath)$ contains only one segment. 
In case of disconnections $|\Segments(\pPath)|>1$ and $|\Segments(\pPath)|=0$ if $\pPath$ does not exist in the graph. 

To compute $\pprecision$ and $\precall$, we imagine sampling a sub-path of $\pPath$ by randomly selecting its two end-points independently and with uniform probability along the path.
The sub-path is connected in the target network only if its both end-points lie on the same path segment $\segment$.
The probability of such event is $\pprob_{\pPath}=\frac{\sum_{\segment\in\Segments} l(\segment)^2 }{l(\pPath)^2}$, where $l(.)$ denotes path length.
We define path recall as the average of these probabilities over all paths $\pPath\in\pPaths$ sampled from the ground truth network $\precall=\frac{1}{|\pPaths|}\sum_{\pPath\in\pPaths} \pprob_\pPath$. 
The path precision $\pprecision$ is computed according to the same formula, but with paths sampled from the predicted network and matched to the ground truth one. 

\subsection{Junction-Based Metric (\JNEW{})}
\label{sec:junctionBasedNew}


\begin{figure}
  \centering
  \setlength{\tabcolsep}{0pt}
  \begin{tabular}{@{}l@{}}
    \includegraphics[trim={0, 5.0cm, 1cm, 0.3cm},clip]{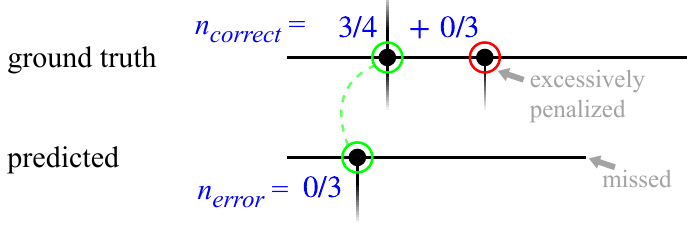} \\
    \multicolumn{1}{c}{\small (a) the existing junction score \JOLD{}} \\
    \\
    \includegraphics[trim={0, 1.0cm, 0, 3.7cm},clip]{new_junction_expl_v2.pdf} \\
    \multicolumn{1}{c}{\small (b) the new junction score \JNEW{} } \\
  \end{tabular}
  \caption{
A comparison of the existing junction score (a) to our junction score (b). 
Feature points are marked as black dots, matches in green and unmatched features in red.
For readability, we only consider the features on the horizontal lines, and assume the vertical lines continue indefinitely.
Candidate and actual edge matches are depicted by hollow nodes.
See section~\ref{sec:junctionBased} for the definition of \JOLD{} and section~\ref{sec:junctionBasedNew} for the definition \JNEW{}.
\label{fig:new_junction_expl}
}
\end{figure}

As discussed in section~\ref{sec:junctionBased}, the junction score \JOLD{}~\cite{Bastani18} is insensitive to road interruptions and excessively penalizes junctions with missing $k-2$ out of $k$ incident edges.  
To address these shortcomings, we propose a new junction score $\JNEW{}$, including junction precision $\jprecision$ and recall $\jrecall$.
As for \JOLD{}, computing $\JNEW{}$ involves matching feature nodes in the two networks and comparing the numbers of edges incident on them. 
Unlike in \JOLD{}, where the feature set consists exclusively of junctions -- nodes with at least three incident edges -- we take both junctions and endpoints as features. 
Moreover, we enable matching features of one graph not only to features of the other graph, but also to their closest points on edges.
This gives our metric the desired sensitivity to unwarranted road breaks and prevents excessively penalizing specific cases of missing edges.
Fig.~\ref{fig:new_junction_expl} illustrates the differences between \JOLD{} and $\JNEW{}$. 

We denote a match by $(i,j)$, where $i$ belongs to the ground truth and $j$ to the predicted graph,
and both are features, or one of them is a feature, and the other is its closest point on an edge.
We perform greedy matching with the cost of a match $\jmcost_{ij} = \jcoeff \jdist_{ij}+| \order_i-\order_j |$, where $\order_i$ is the number of edges incident on $i$ if $i$ is a feature and by convention $\order_i=2$ if $i$ is a point on an edge. 
$\jdist_{ij}$ is the distance between $i$ and $j$.
$\jcoeff$ is a parameter of the score.
We only allow a feature to participate in one match, but 
we do not enforce any constraint on the edges.
We only consider matches $(i,j)$ such that $i$ and $j$ are within a predefined small distance $\jmaxdist$.

We denote the set of matches $\jMatches$, 
the sets of unmatched ground truth features
by $\FGU$ and the set of unmatched predicted features by $\FPU$.
We estimate the number of true positive incident edges as $\jTP=\sum_{(i,j)\in\jMatches} \min\{\order_i,\order_j\}$, 
the number of predicted edges as $\jPP=\sum_{(i,j)\in\jMatches} \order_j+\sum_{j\in\FPU} \order_j$, 
and the number of ground truth edges as $\jAP=\sum_{(i,j)\in\jMatches} \order_i + \sum_{i\in\FGU} \order_i$.
We compute the precision and recall as $\jprecision=\frac{\jTP}{\jPP}$ and $\jrecall=\frac{\jTP}{\jAP}$.

\subsection{Subgraph-Based Metric (\GNEW{})}
\label{sec:graphBasedNew}
\begin{figure}
\centering
\includegraphics[trim=0 1cm 0.9cm 0, clip]{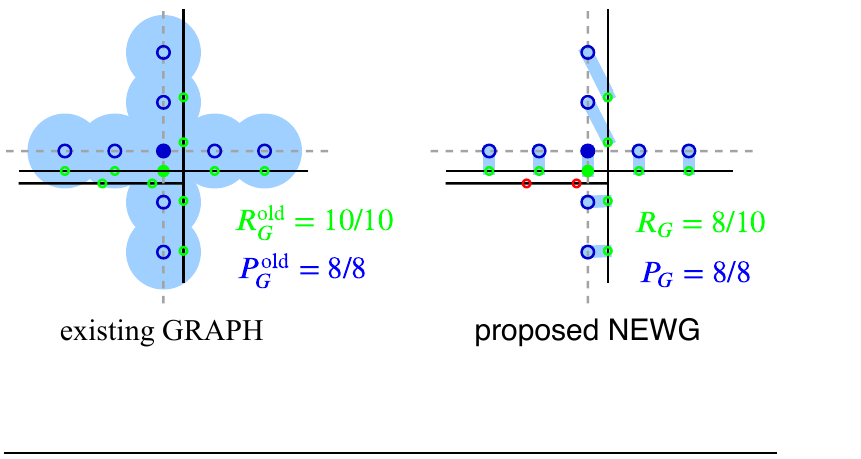}
\caption{
The difference between the existing subgraph-based score \GOLD{} and our subgraph-based score \GNEW{}.
A single starting point is shown in both the predicted network (as filled blue circle) and ithe ground truth networks (as a filled green circle). 
The networks are drawn with dashed gray and solid black lines, respectively.
All the control points in the ground truth network (hollow green circles) are within a matching distance (visualized with light blue disks) from the control points in the predicted network (hollow blue circles). 
This makes the existing score insensitive to the missing road.
In our score, the matching is one-to-one (visualized by light blue lines).
In result, some of the control points remain unmatched (marked in red), which gives the score sensitivity to the missing road.
\label{fig:new_graph_expl}
}
\end{figure}

In section~\ref{sec:subgraphBased} we have exposed the lack of sensitivity of the local graph comparison \GOLD{}~\cite{Biagioni12} to false positive roads that are far away from ground truth roads and to errors consisting in missing one of several closely spaced roads.
In order to fix this lack of sensitivity, we propose a new score \GNEW{}.
Like \GOLD{}, \GNEW{} is based on comparing sets of graph locations accessible by travelling a short distance in the graph from a randomly selected starting point.
However, unlike in the old score, we sample the starting points both in the ground-truth and predicted graphs, which makes the score sensitive to false positives.
In order to avoid that two sufficiently close roads in one graph are matched to a single road in the other graph, we make the matching of the accessible locations one-to-one.
This makes the score capture errors consisting in predicting a single road where several closely spaced roads exist.
This difference is illustrated in Figure~\ref{fig:new_graph_expl}.

To compute the score, we iteratively sample a starting point in one of the graphs.
We then find its closest point in the other graph.
Using breadth-first graph traversal, we crop out subgraphs accessible by travelling a predefined distance from the starting points.
Control points are inserted in equal intervals during the traversal.
We then perform a one-to-one matching of control points from the two graphs by the Hungarian algorithm, with the cost of matching two points equal to the Euclidean distance between them.
Only points within a predefined distance are matched.
Calculation of the score is based on the number of matched and unmatched control points.
We define subgraph-based precision as 
$
\gprecision=\frac{\gTP}{\gPP}
$
 and subgraph-based recall as 
$
\grecall=\frac{\gTP}{\gAP}
$,
where $\gTP$ is the total number of matched control points, $\gPP$ is the number of control points in the predicted graph and $\gAP$ is the number of control points in the ground truth graph.


\section{Experiments}
\label{sec:experiments}

\begin{figure*}[h!]
  \centering
  \setlength{\tabcolsep}{2pt}
  \setlength{\fboxsep}{0pt}
  \setlength{\fboxrule}{0.5pt}
	  \begin{tabular}{@{}ccccccc@{}}
                & {\small interruptions} & 
                  {\small overconnections} & 
                  {\small perturbations} & 
                  {\small doubled roads} & 
                  {\small \shortstack{doubled roads \\ ground-truth}} & 
                  {\small \shortstack{false positives far\\ away from true roads}} \\ 
		\raisebox{1.0cm}{\rotatebox[origin=t]{90}{\small`ground-truth'}} &  
		\fbox{\includegraphics[width=.15\textwidth]{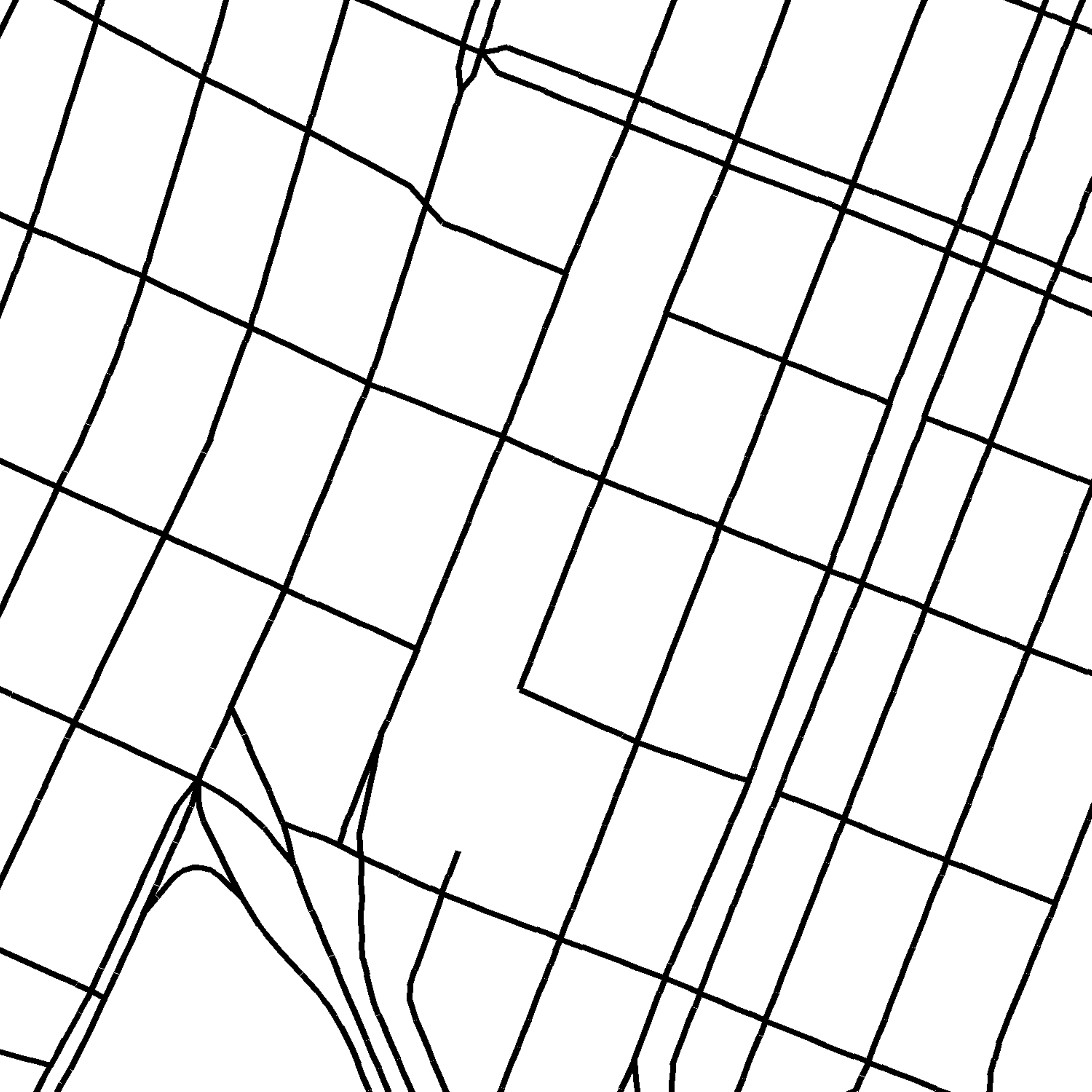}} &  
		\fbox{\includegraphics[width=.15\textwidth]{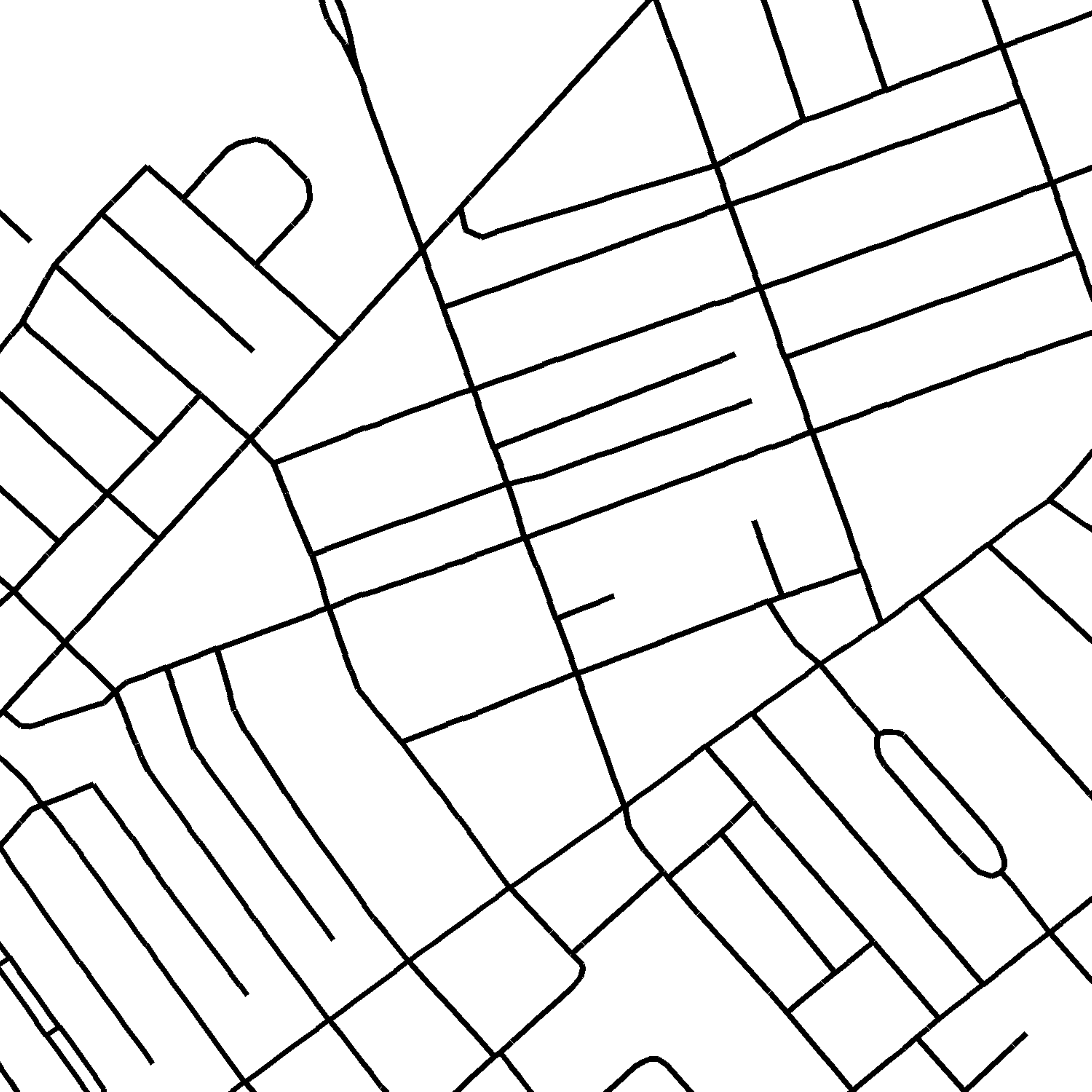}} &  
		\fbox{\includegraphics[width=.15\textwidth]{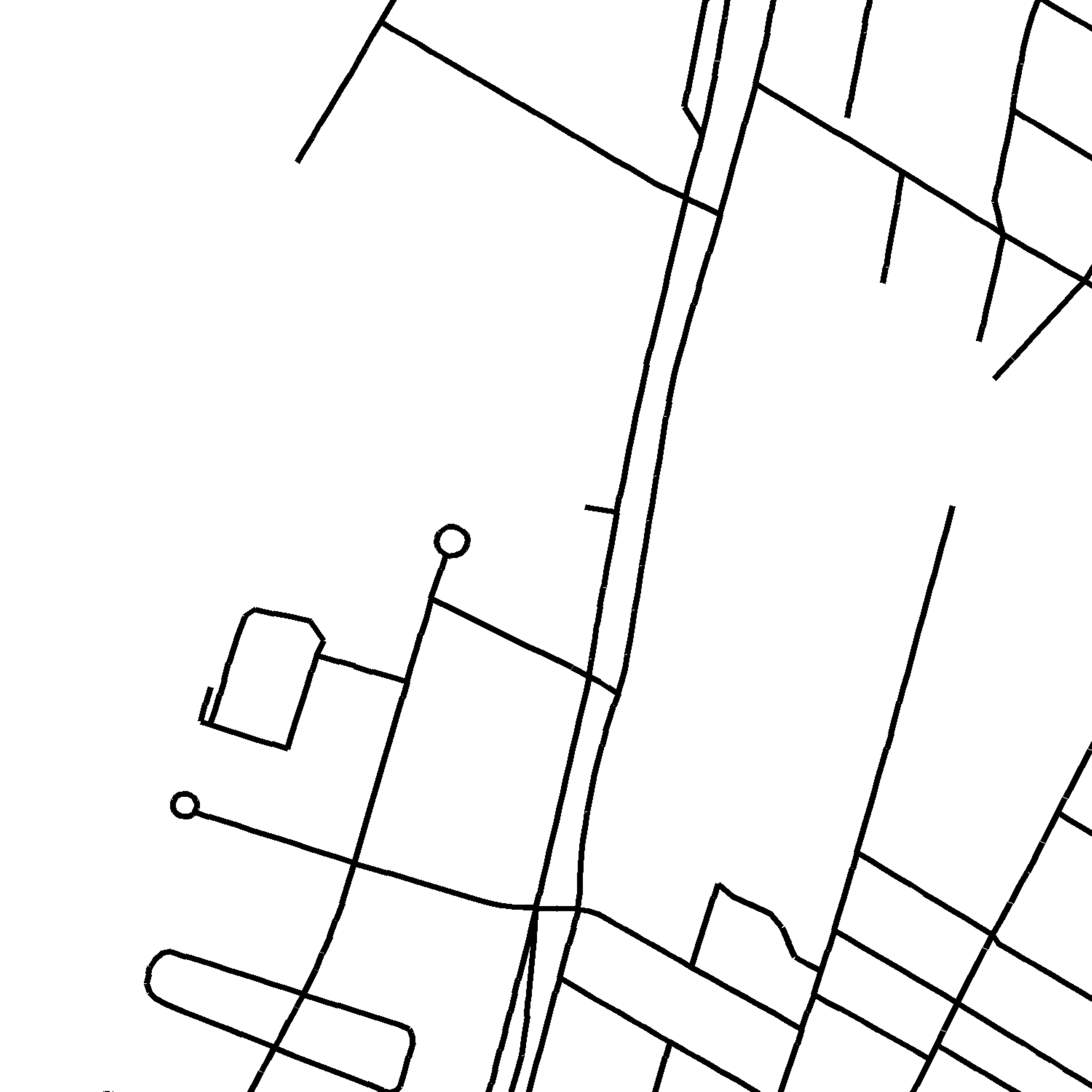}}  &
		\fbox{\includegraphics[width=.15\textwidth]{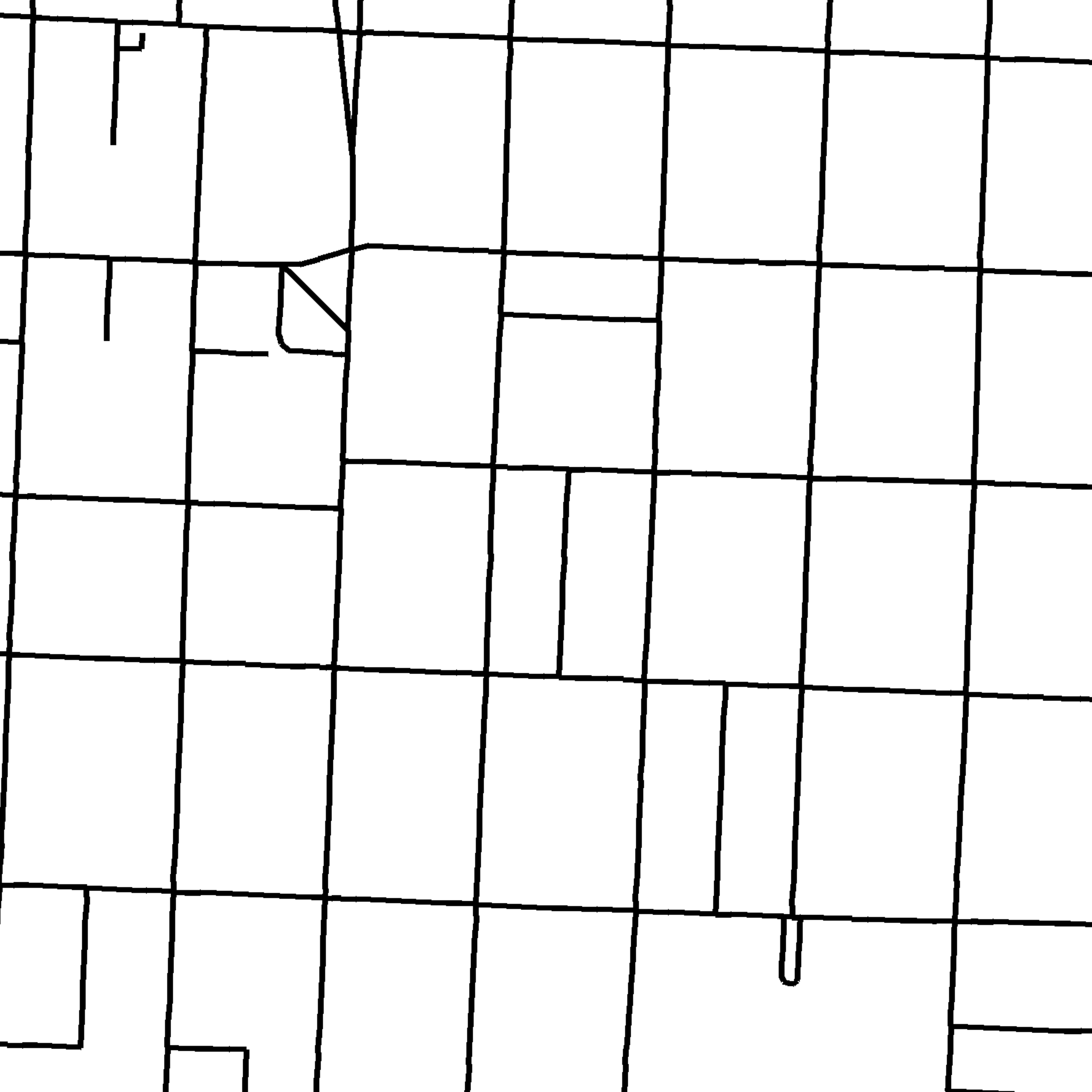}}  &
		\fbox{\includegraphics[width=.15\textwidth]{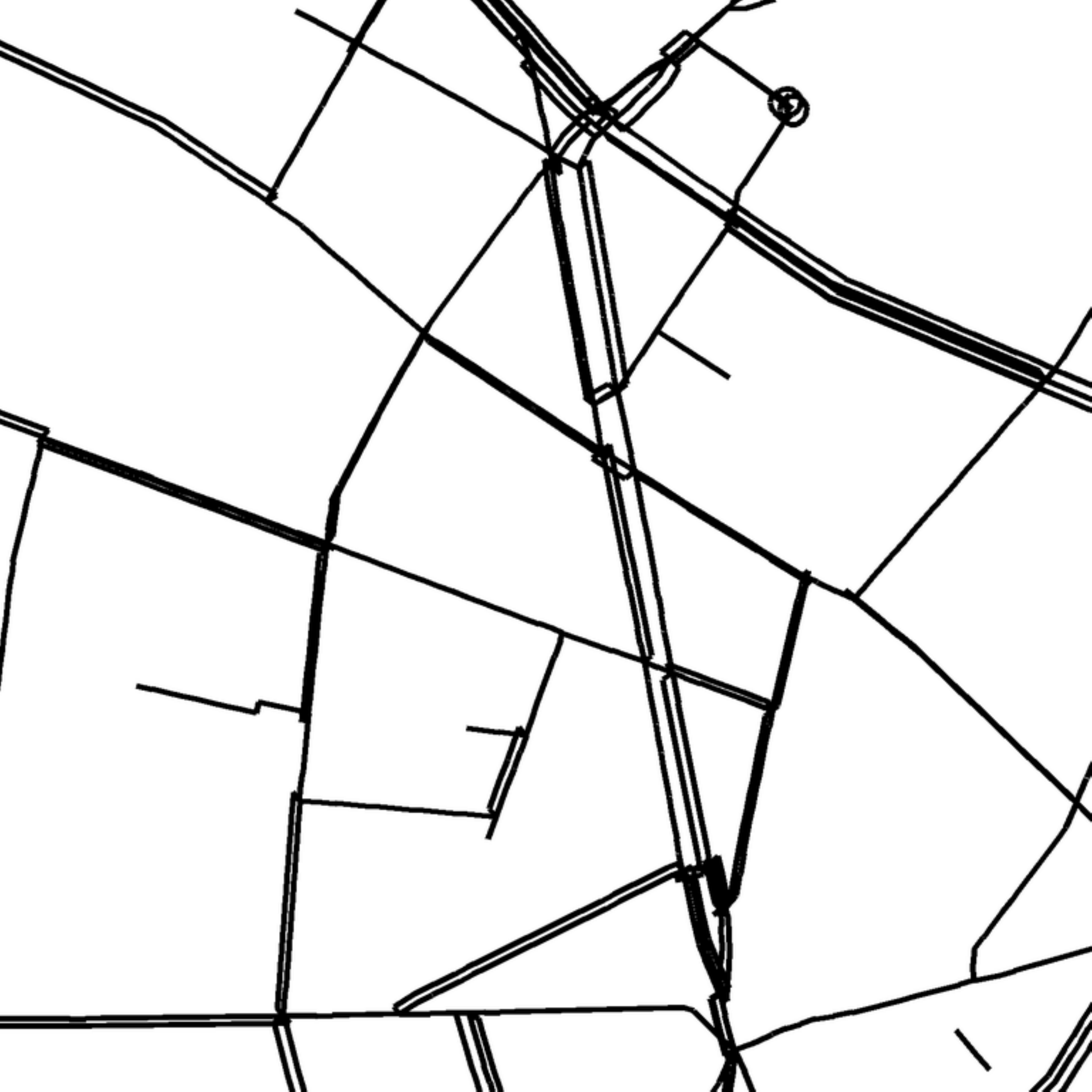}}  &		
		\fbox{\includegraphics[width=.15\textwidth]{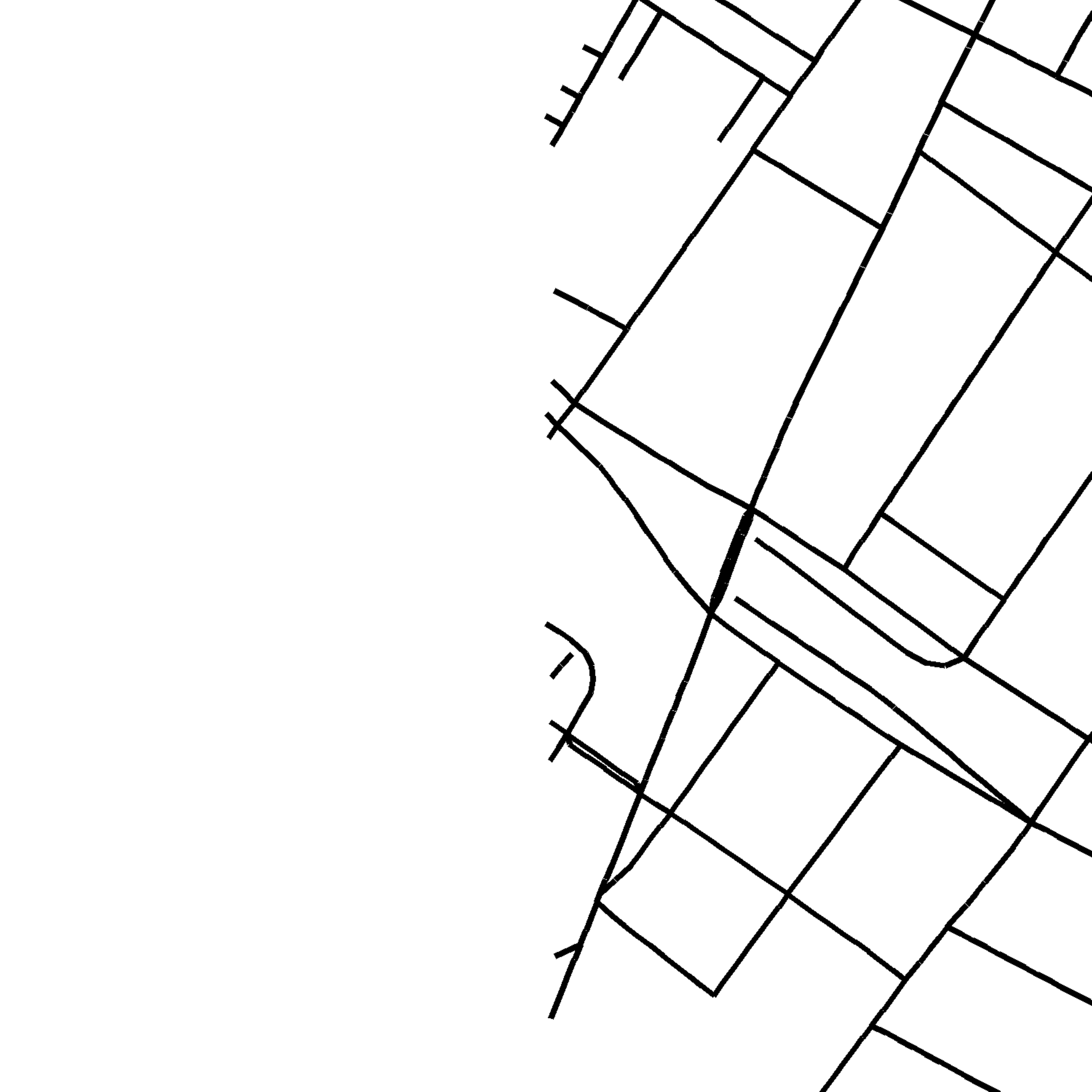}}  \\
                 
		\raisebox{1.0cm}{\rotatebox[origin=t]{90}{\small`prediction'}} &   
		\fbox{\includegraphics[width=.15\textwidth]{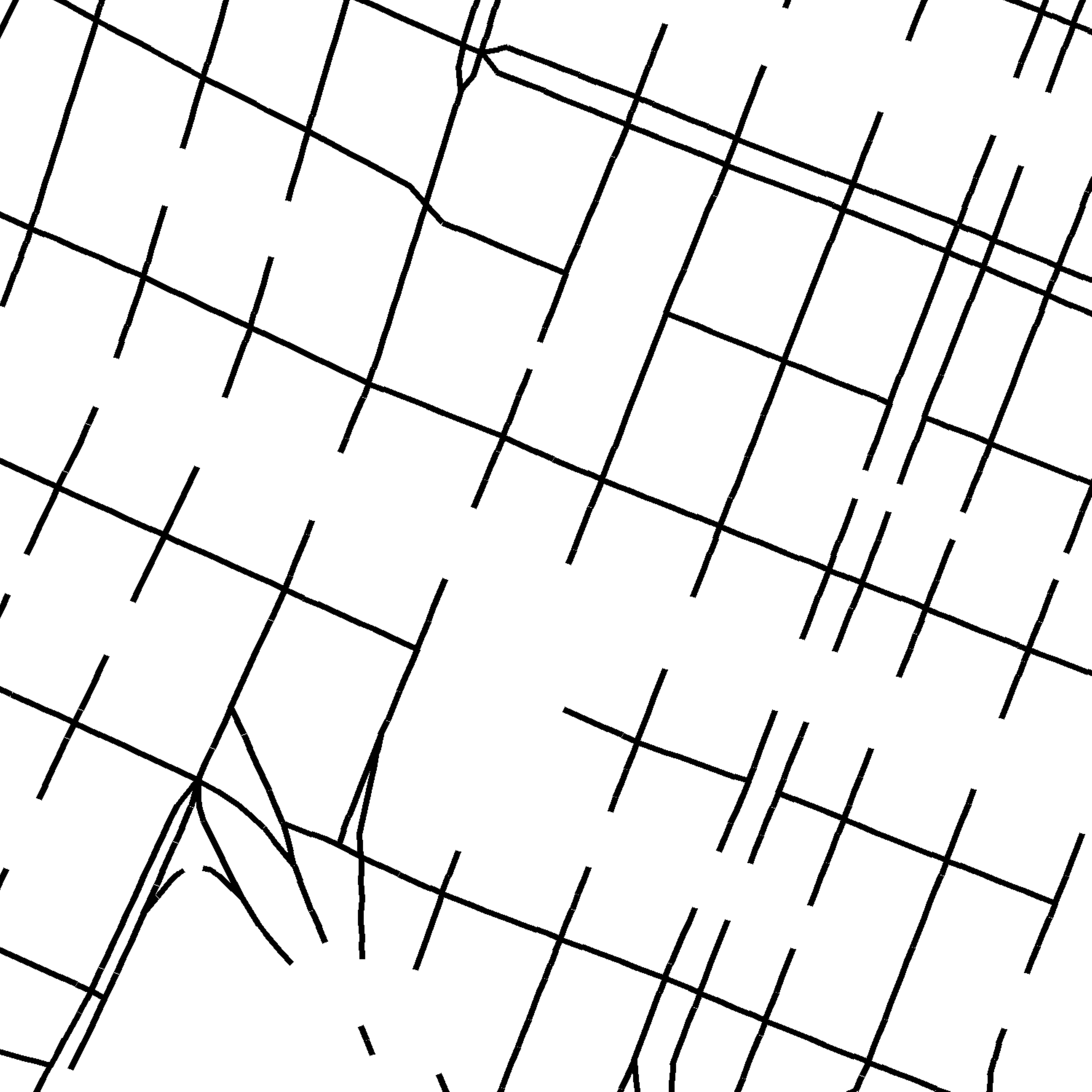}} &  
		\fbox{\includegraphics[width=.15\textwidth]{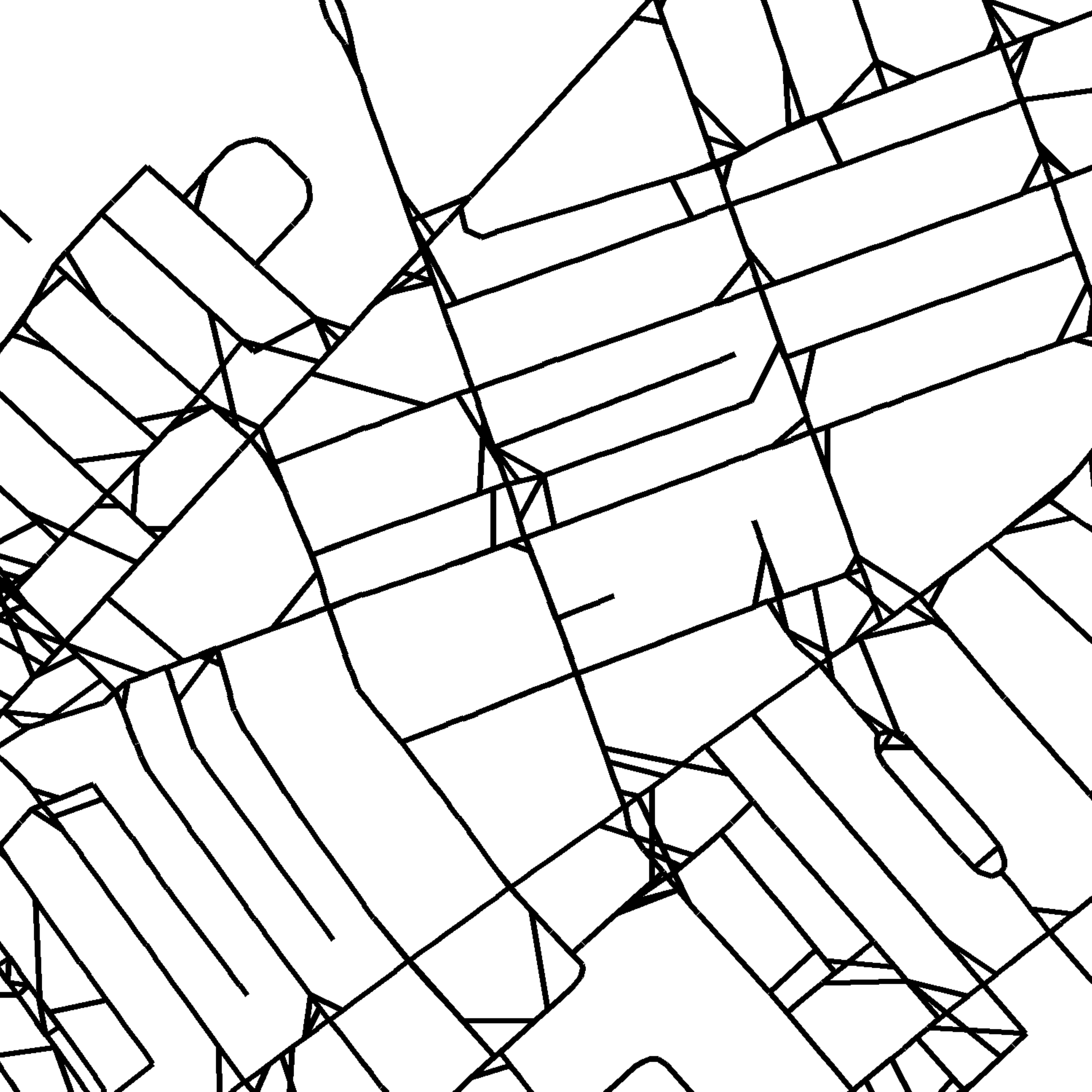}} &  
		\fbox{\includegraphics[width=.15\textwidth]{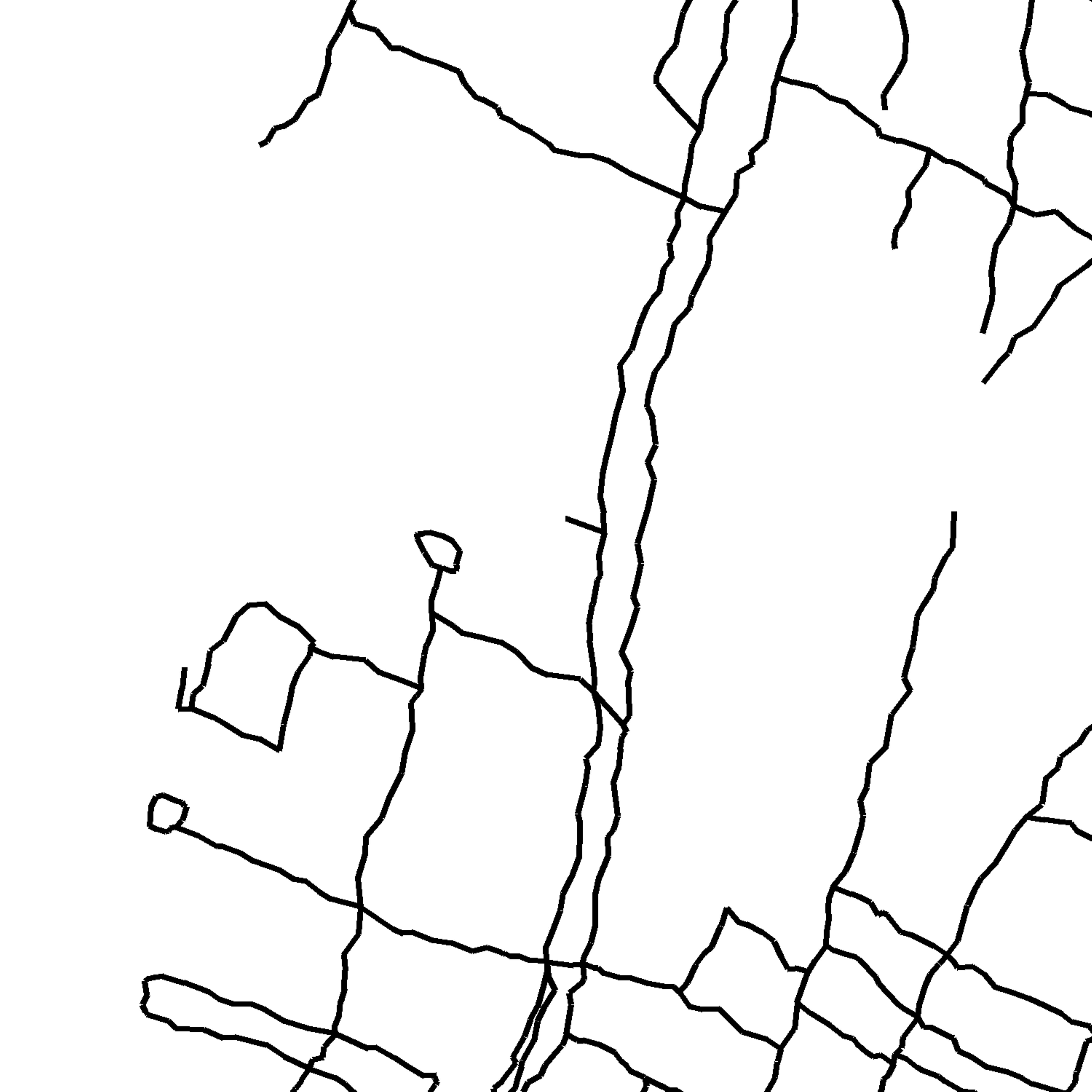}} &  
		\fbox{\includegraphics[width=.15\textwidth]{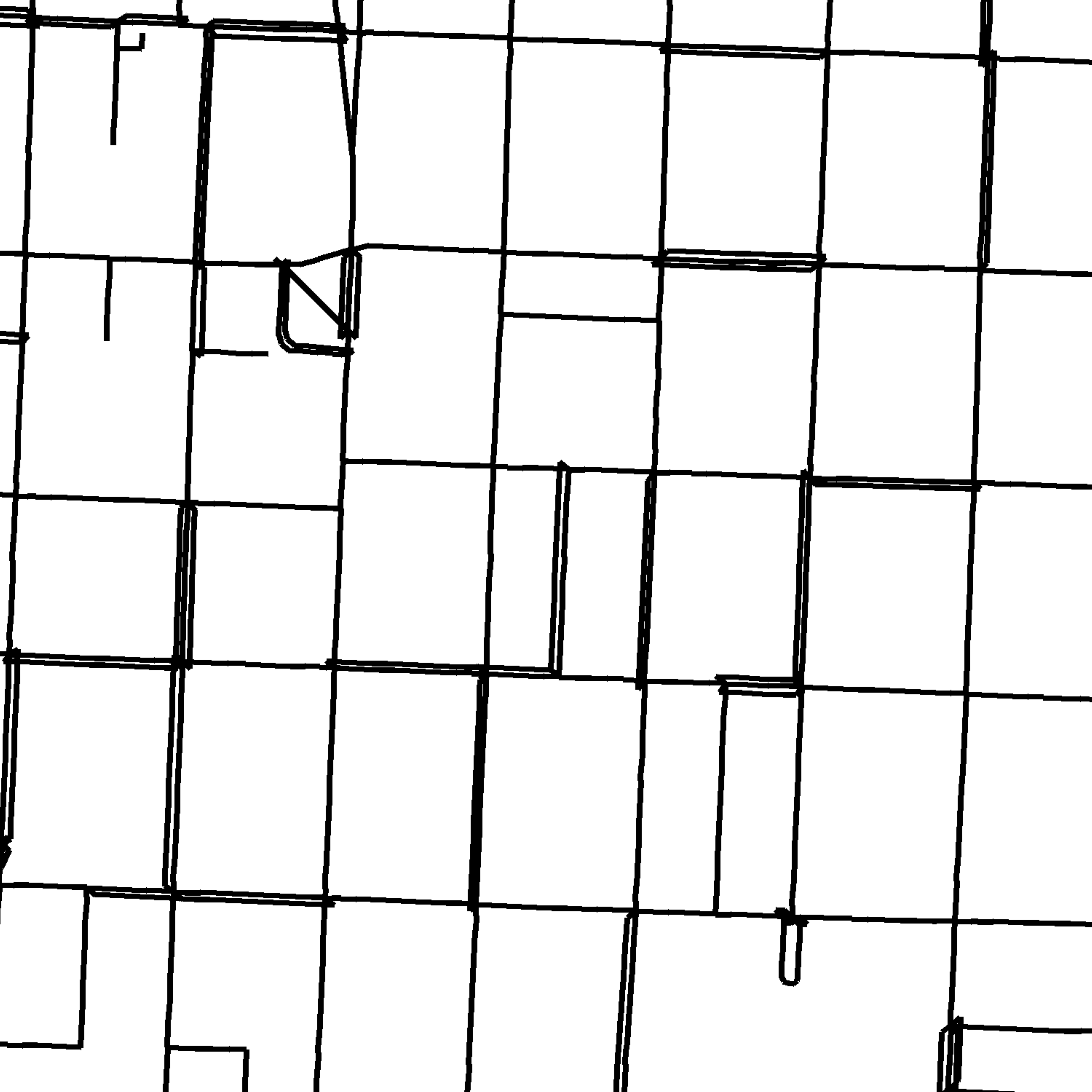}} & 
		\fbox{\includegraphics[width=.15\textwidth]{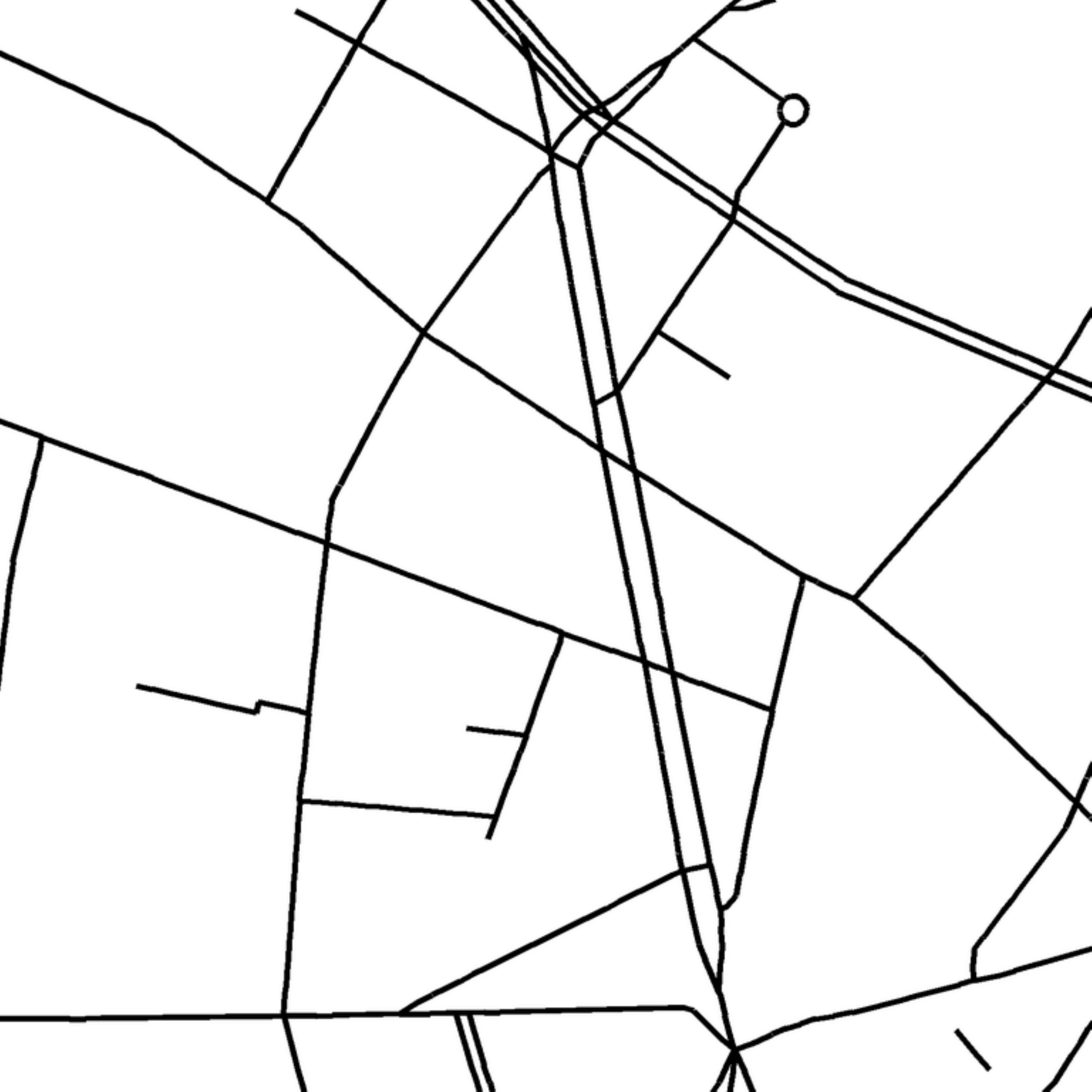}} &		 
		\fbox{\includegraphics[width=.15\textwidth]{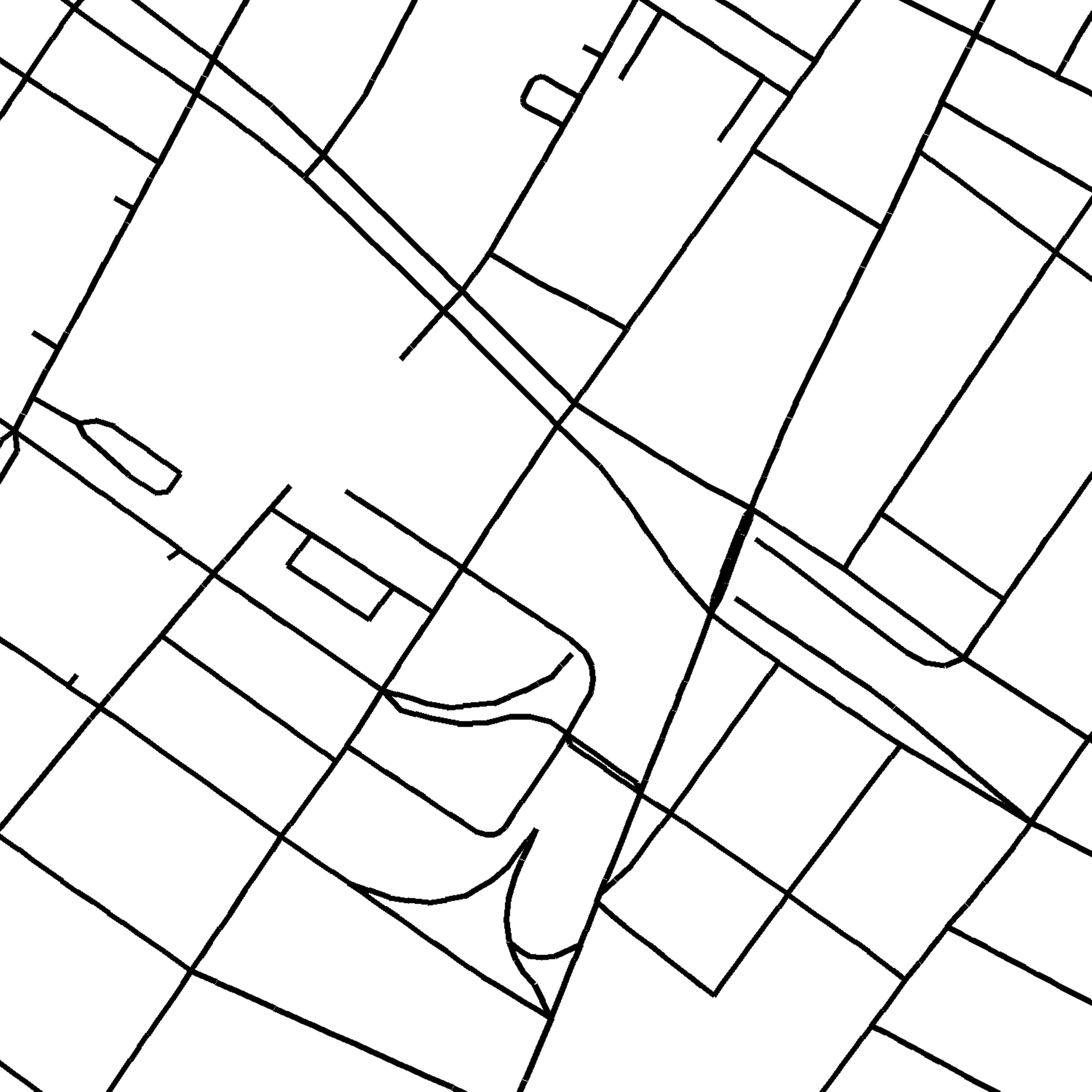}} \\
	  \end{tabular}
  \caption{\small 
Example pairs of road networks from the benchmark dataset. 
\label{fig:benchmark}}
  
\end{figure*}


\makeatletter
\newcommand\HUGE{\@setfontsize\Huge{30}{30}}
\makeatother

\begin{figure*}[h!]
  \centering
  \setlength{\tabcolsep}{1pt}
  \resizebox{\linewidth}{!}{%
      \setlength{\fboxsep}{1pt}
      \setlength{\fboxrule}{0pt}
	  \begin{tabular}{@{}l@{\hskip -12mm}c c c c c c @{}}	
                \multicolumn{1}{c}{} &
                \multicolumn{1}{c}{\HUGE \;\quad interruptions} &
                \HUGE \: overconnections &
                \HUGE \quad perturbations &
                \HUGE \quad doubled roads &
                \HUGE  \shortstack{doubled roads \\ ground-truth} &
                \HUGE \: \shortstack{false positives far\\ away from true roads} \\

                \\
                \multicolumn{7}{c}{\multirow{2}{*}{\HUGE \qquad existing scores}}\\
                \cmidrule(r{3.5cm}){2-4}
                \cmidrule(l{3.0cm}){5-7}
                \\

		\raisebox{4.5cm}{\rotatebox[origin=t]{90}{\HUGE  \TLTS{}}} &
		\fbox{\includegraphics{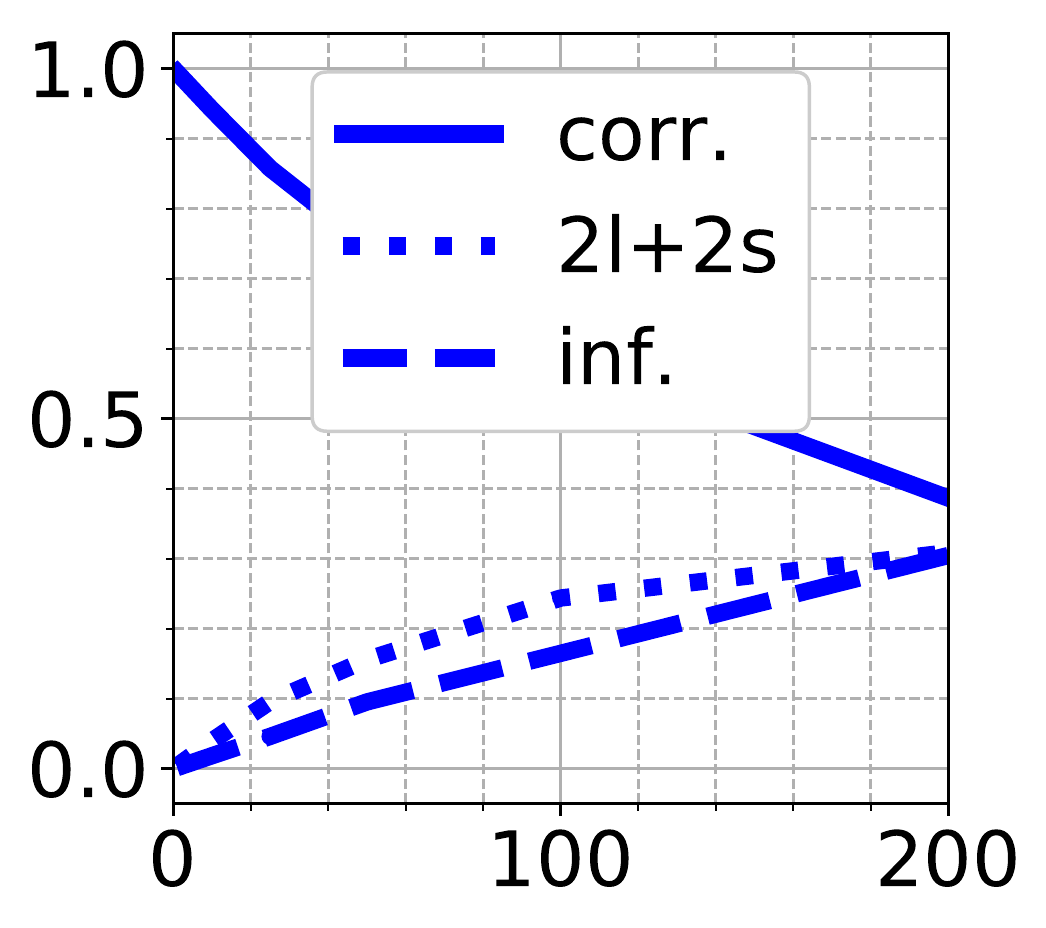}} & 
		\fbox{\includegraphics{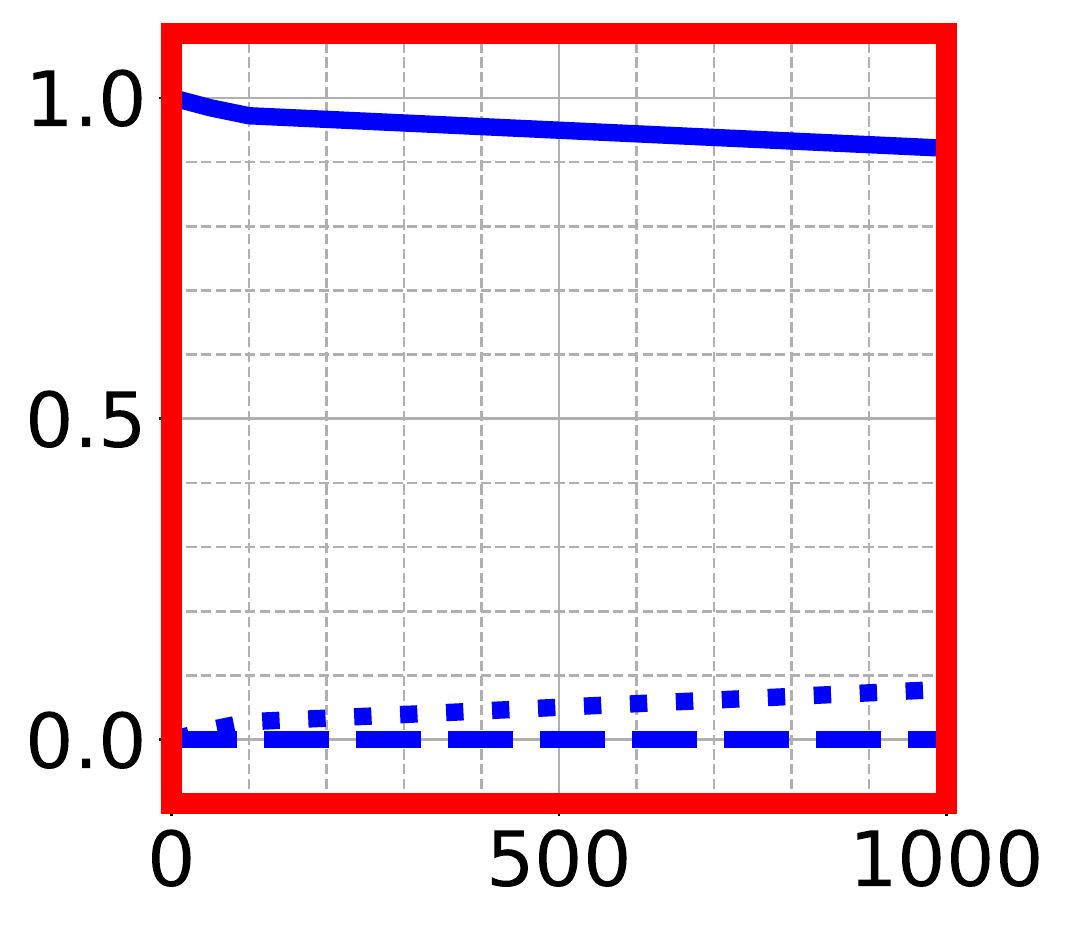}} &
		\fbox{\includegraphics{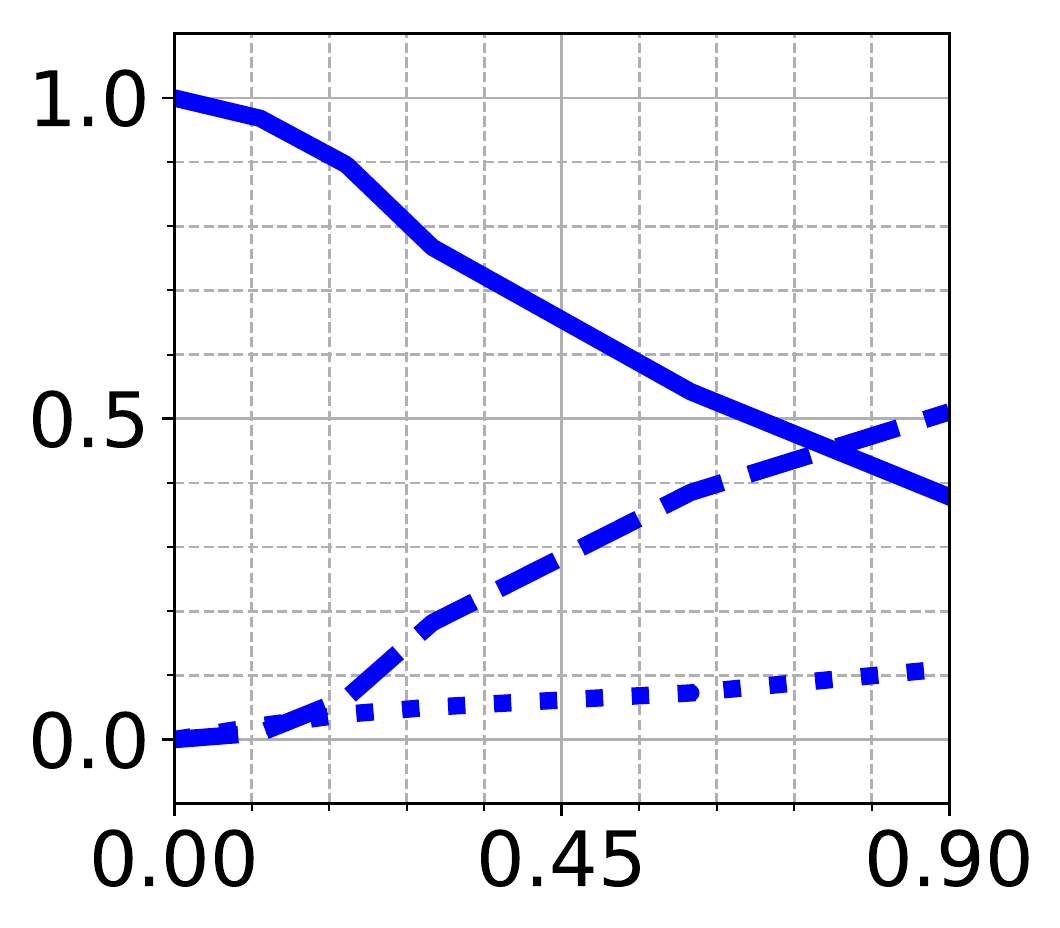}} &
		\fbox{\includegraphics{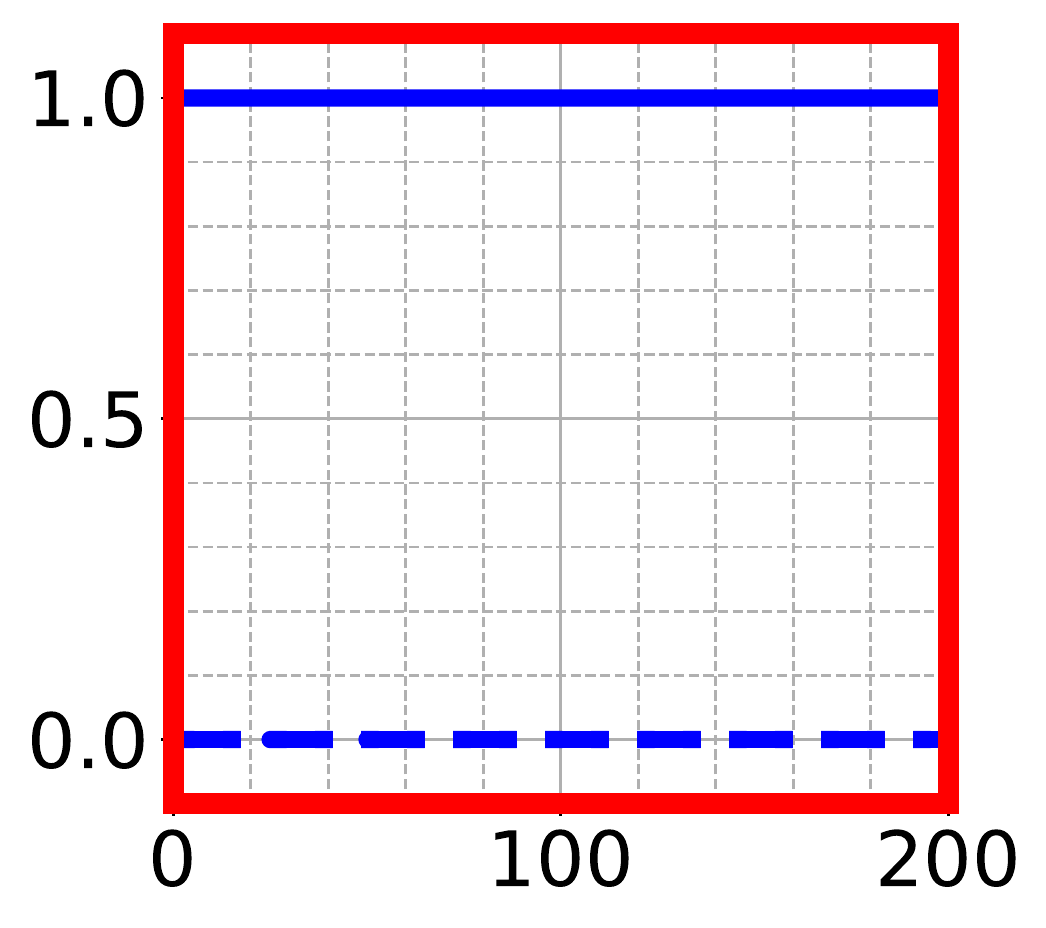}} &
		\fbox{\includegraphics{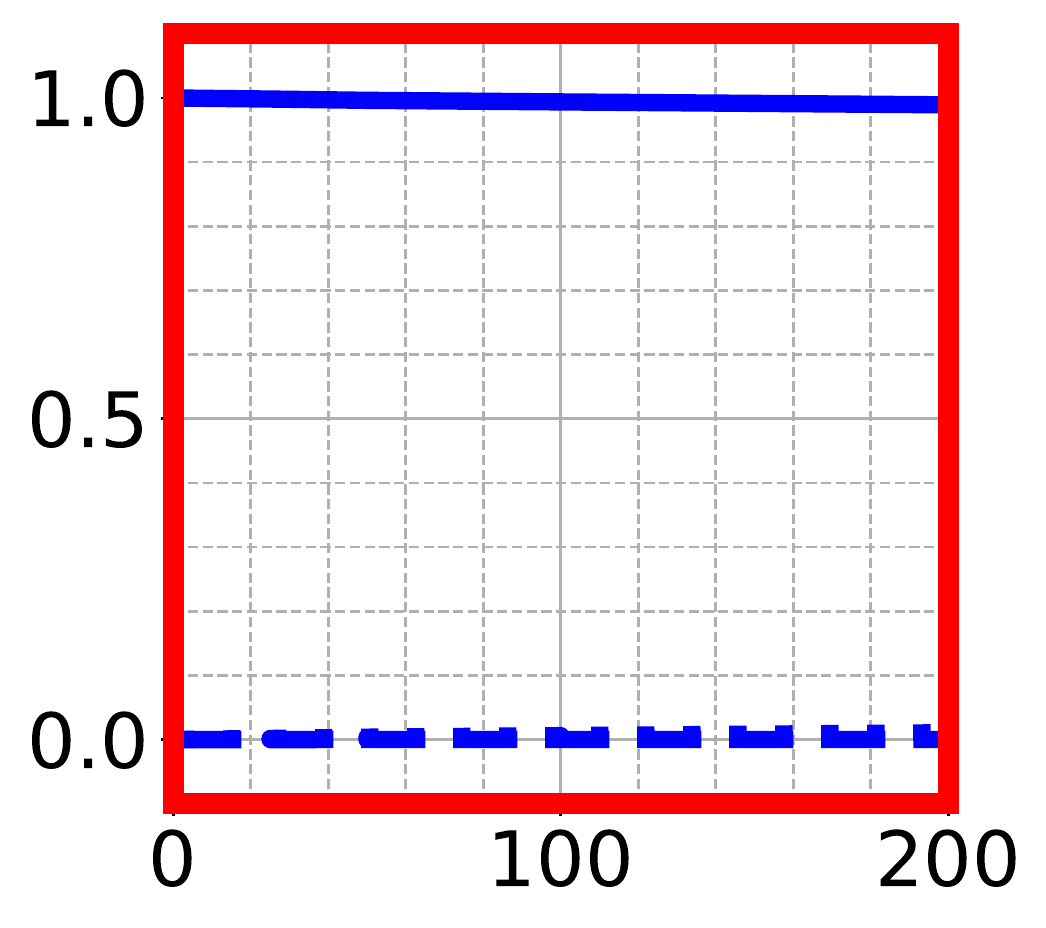}} &
		\fbox{\includegraphics{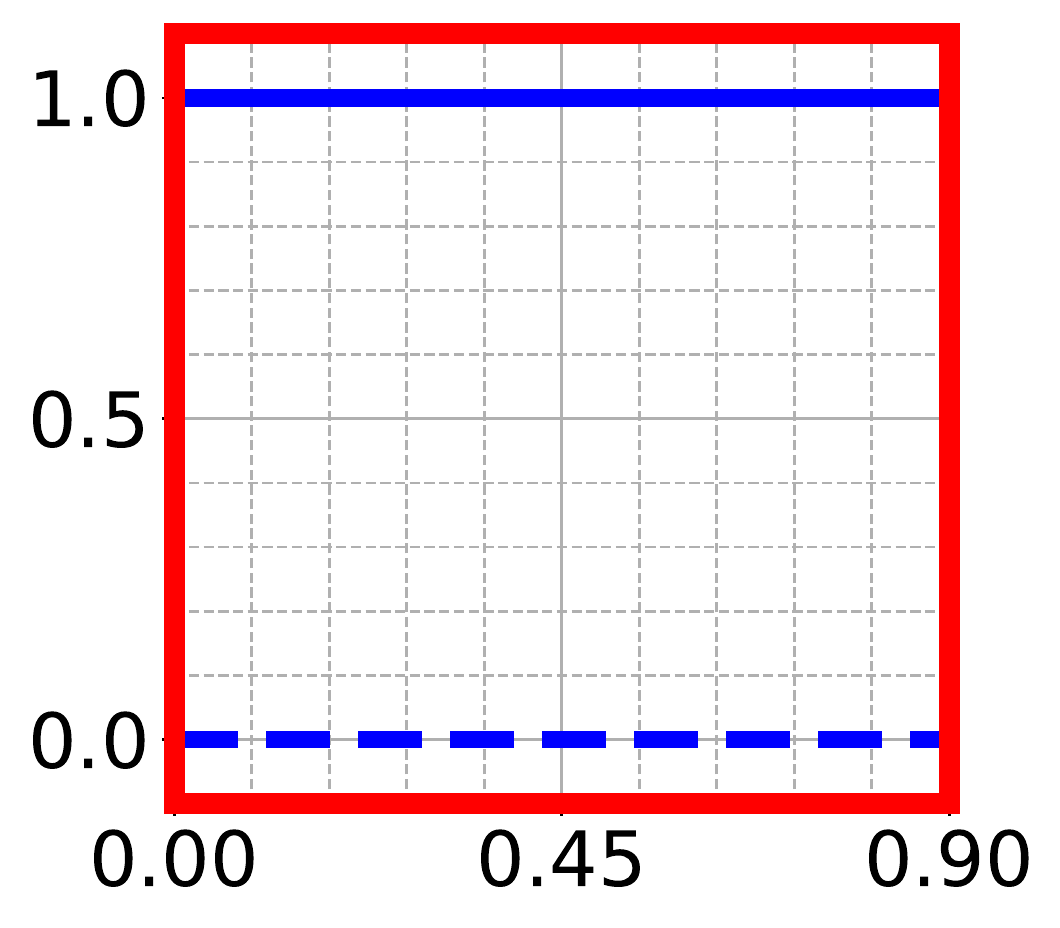}} \\

		\raisebox{4.5cm}{\rotatebox[origin=t]{90}{\HUGE  \APLS{}}} &
		\fbox{\includegraphics{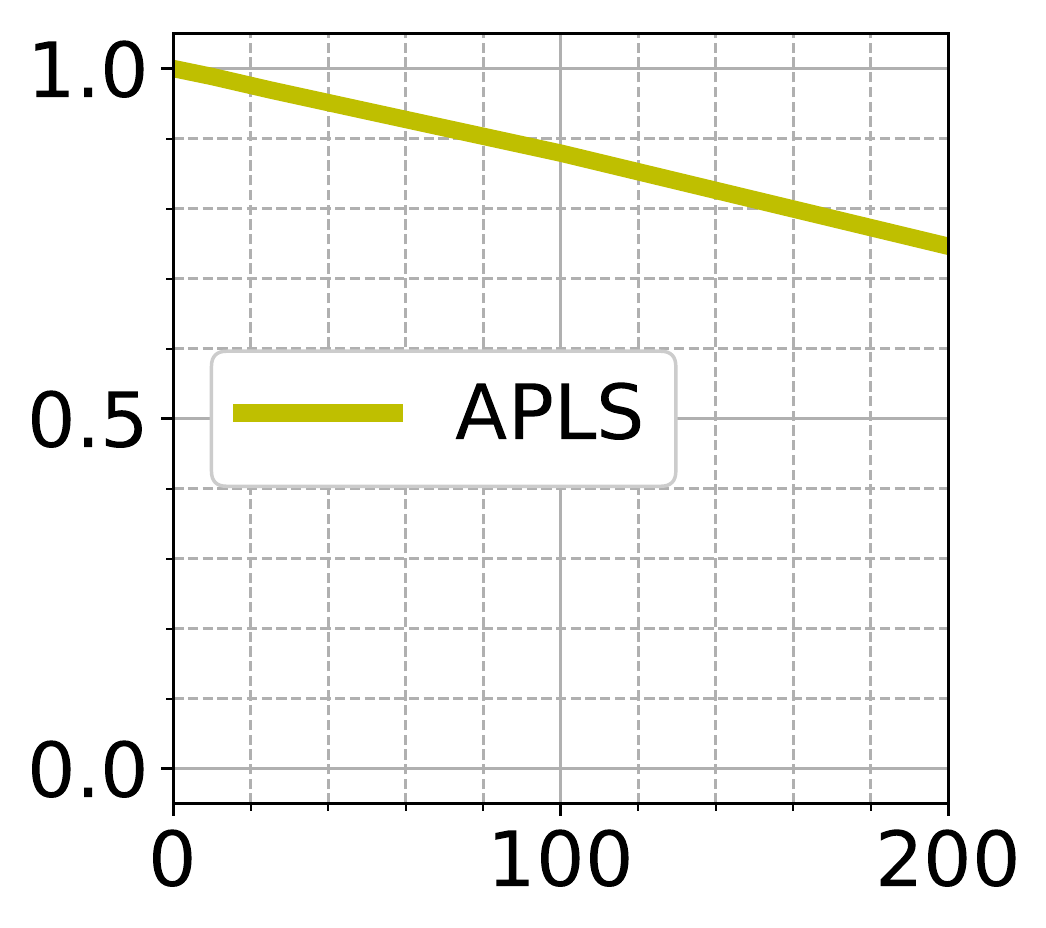}} & 
		\fbox{\includegraphics{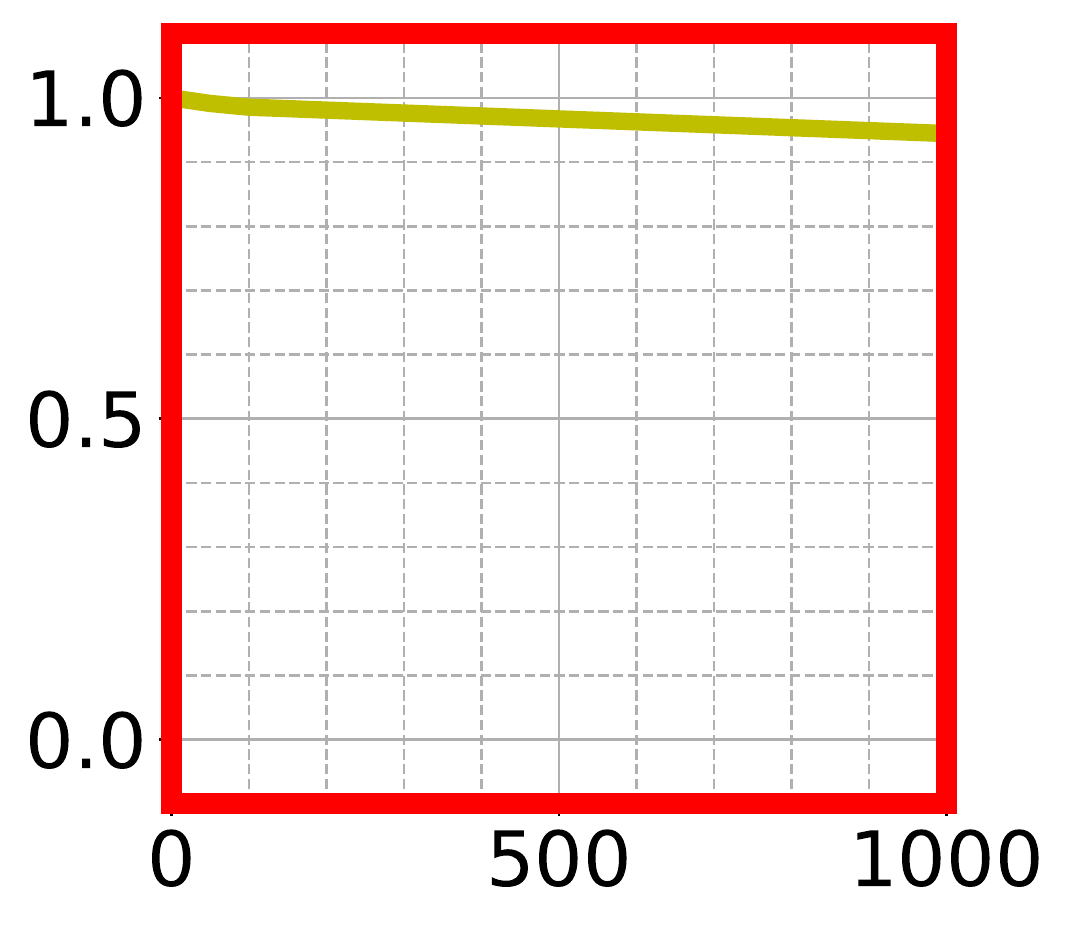}} &
		\fbox{\includegraphics{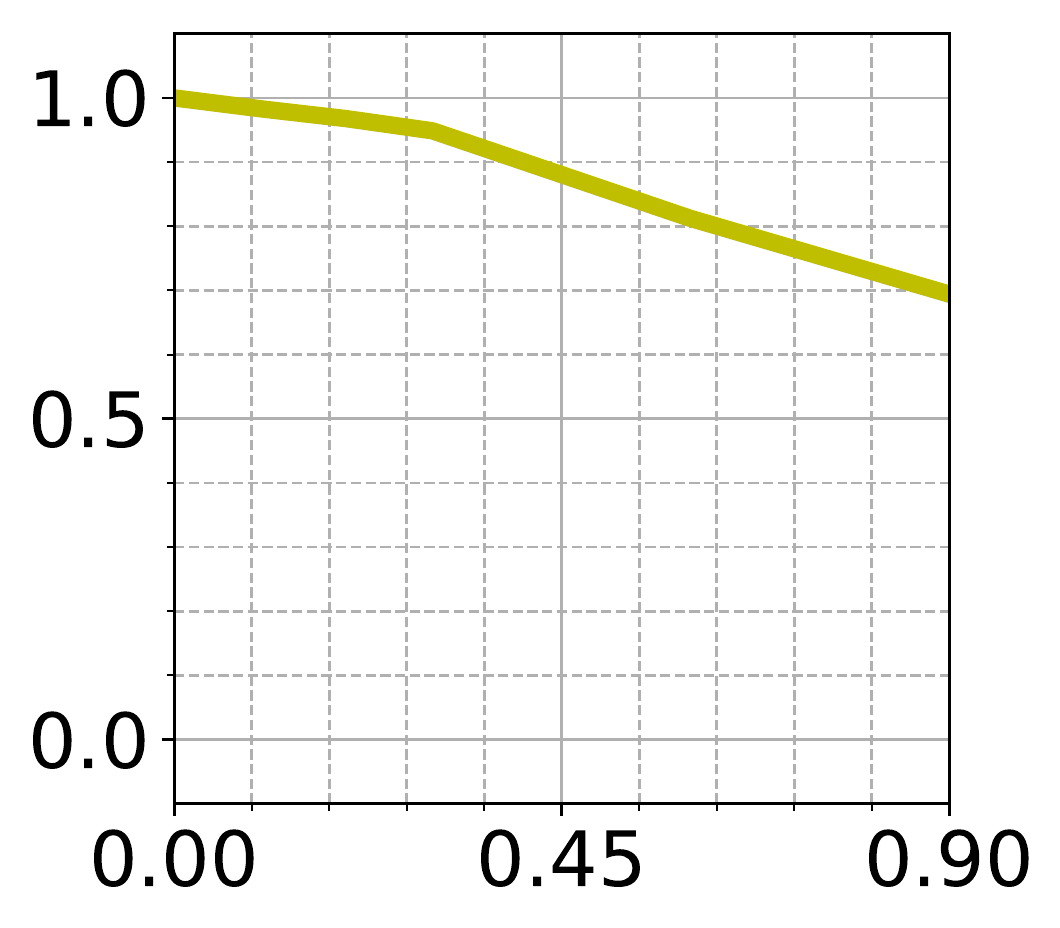}} &
		\fbox{\includegraphics{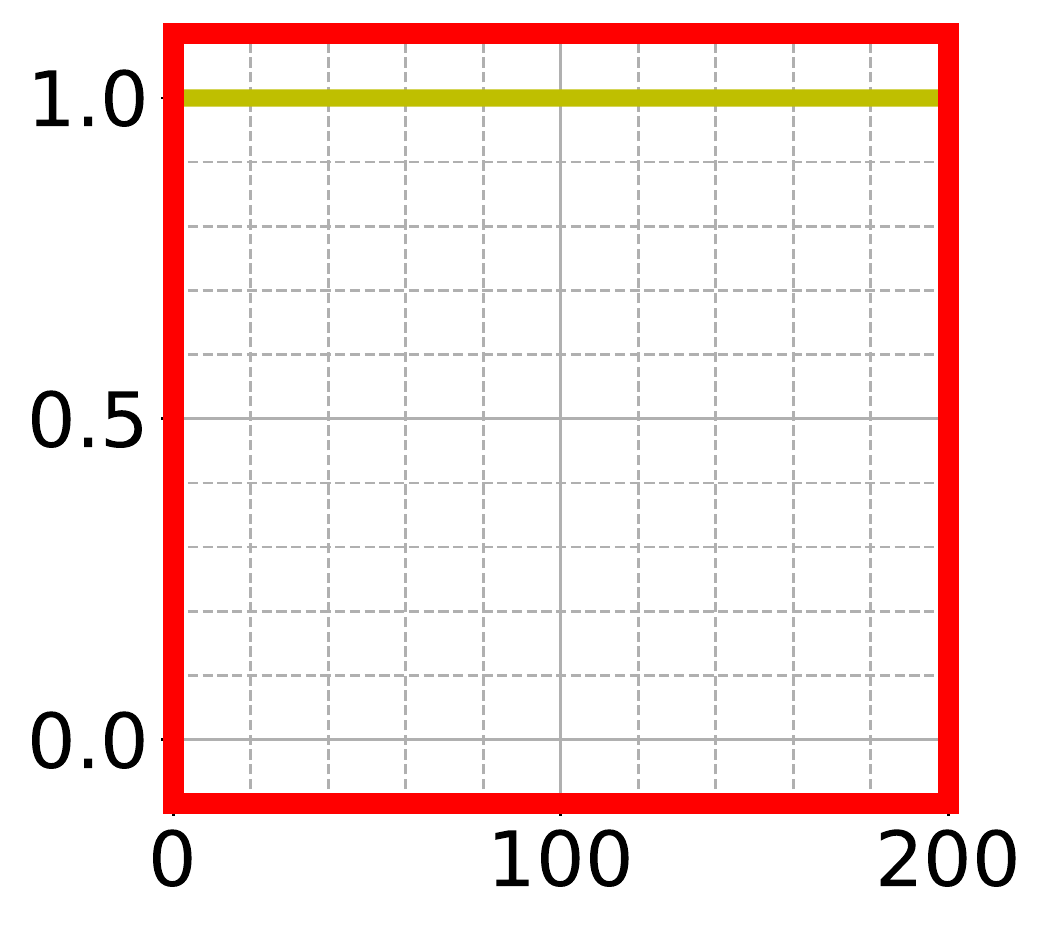}} &
		\fbox{\includegraphics{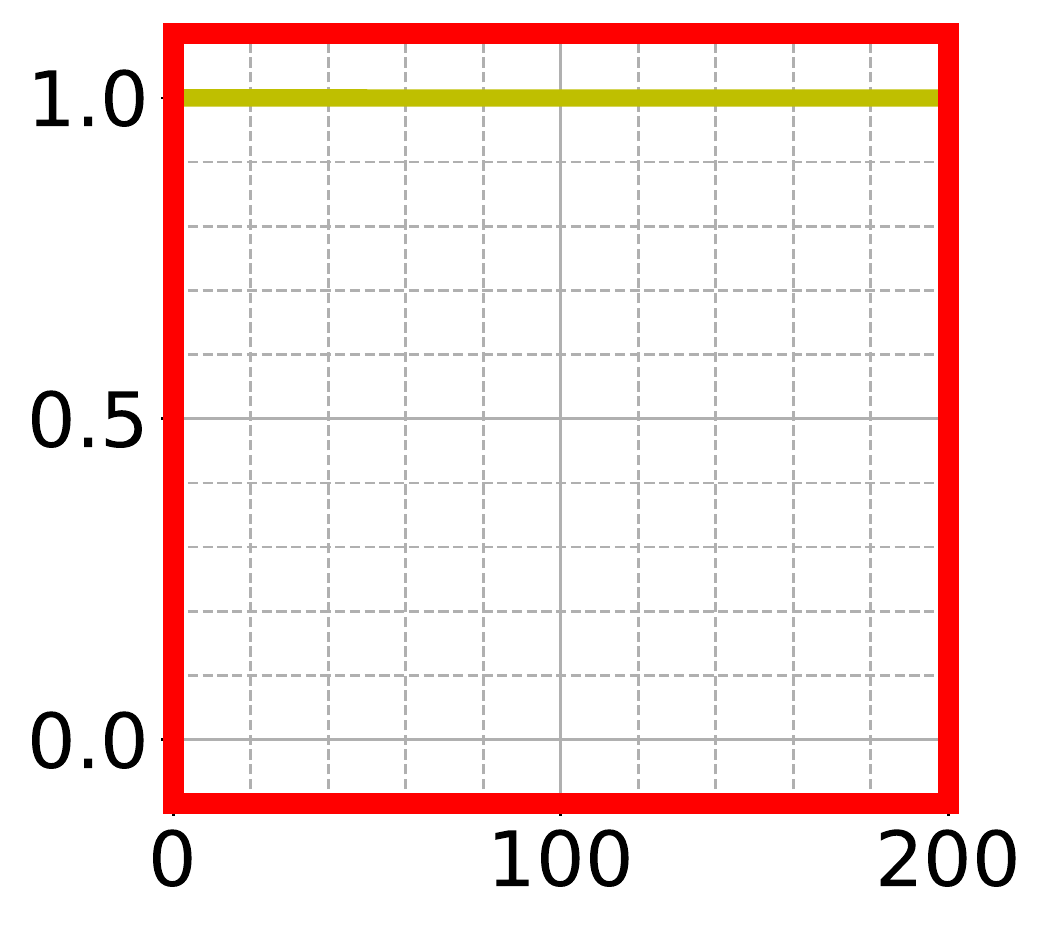}} &
		\fbox{\includegraphics{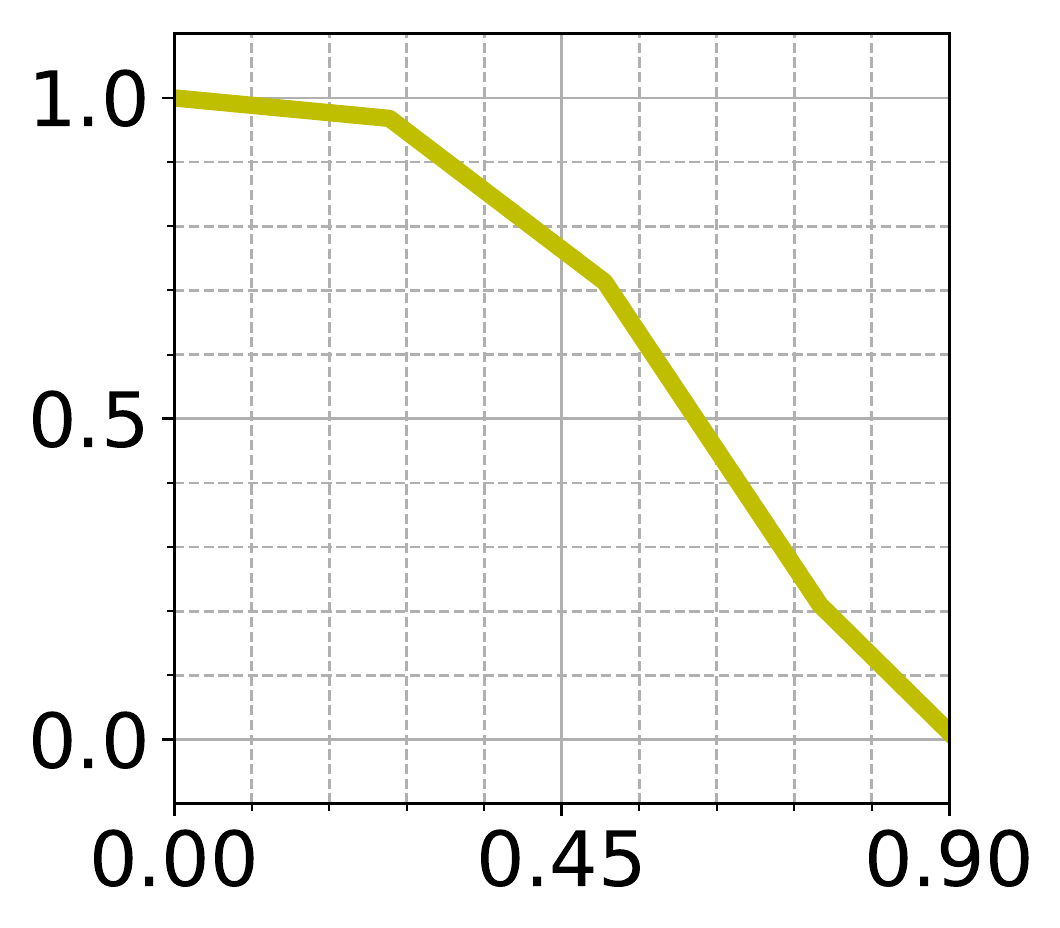}} \\	
		
		\raisebox{4.5cm}{\rotatebox[origin=t]{90}{\HUGE  \GOLD{}}} &
		\fbox{\includegraphics{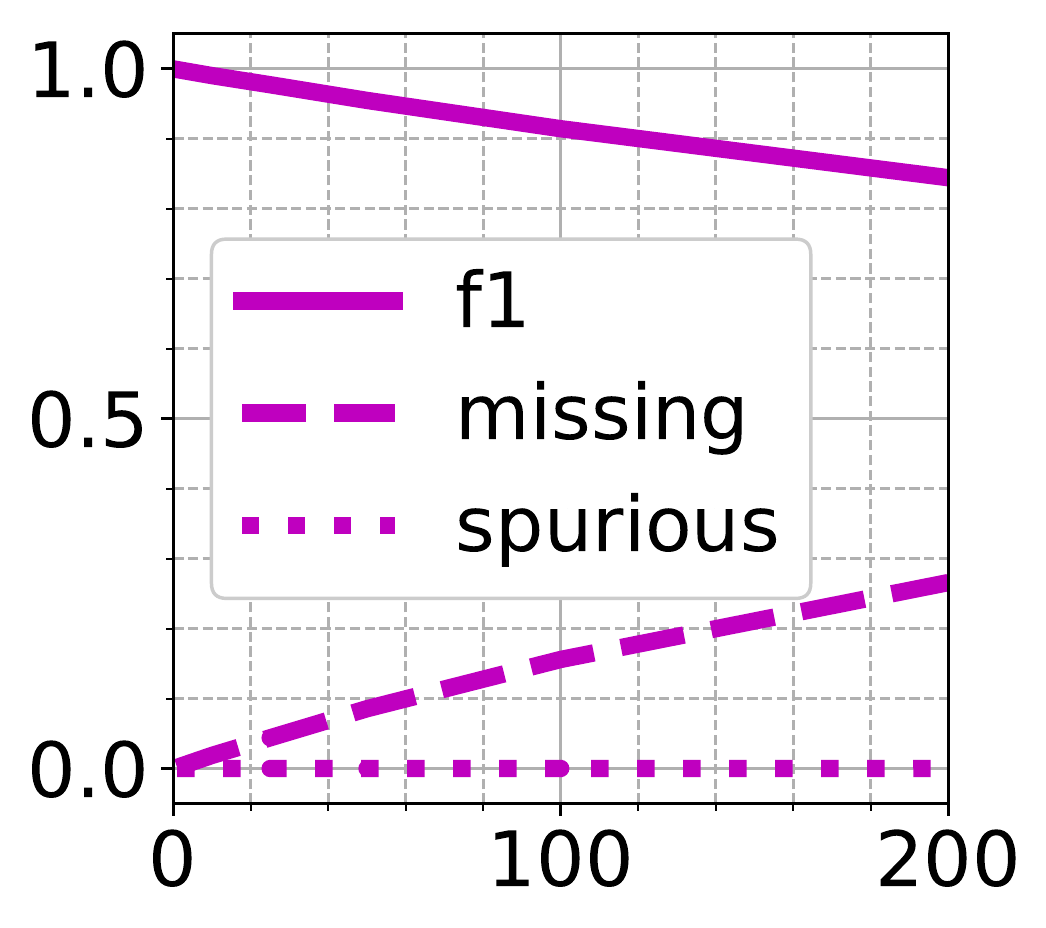}} & 
		\fbox{\includegraphics{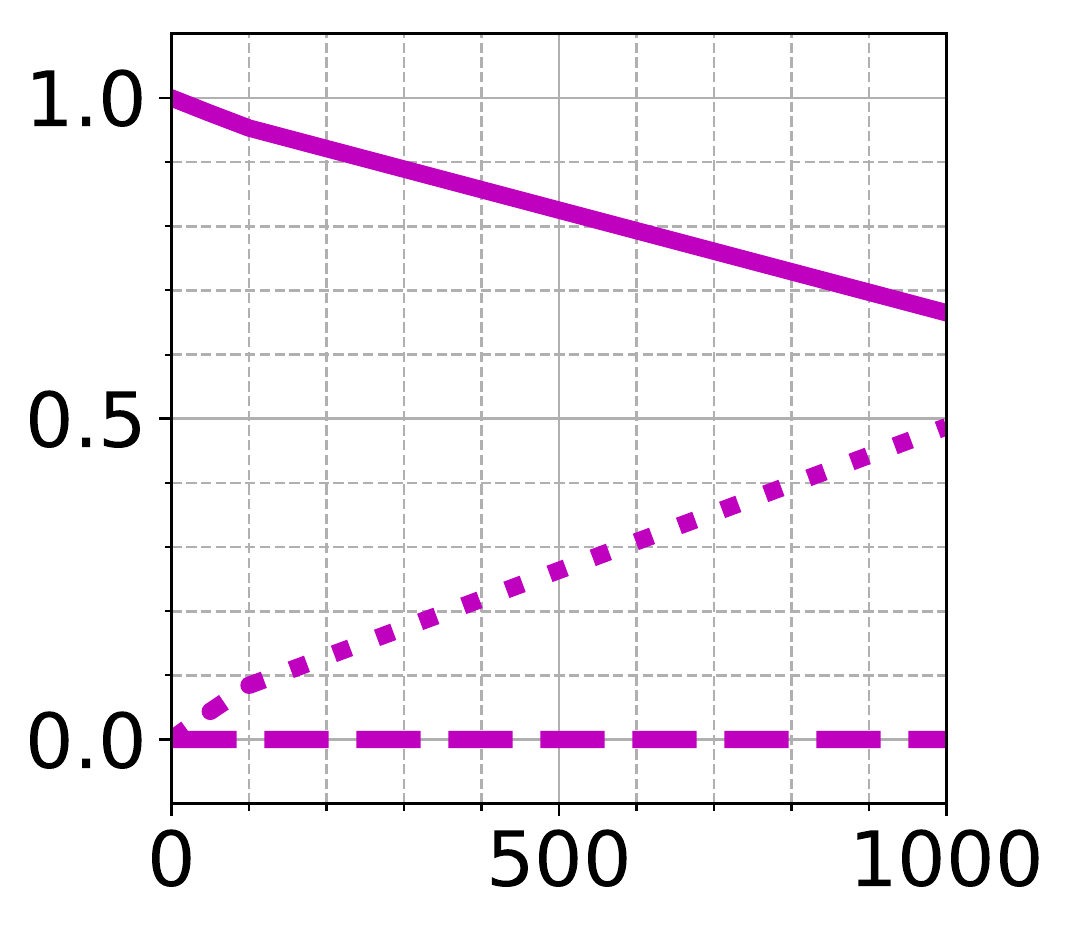}} &
		\fbox{\includegraphics{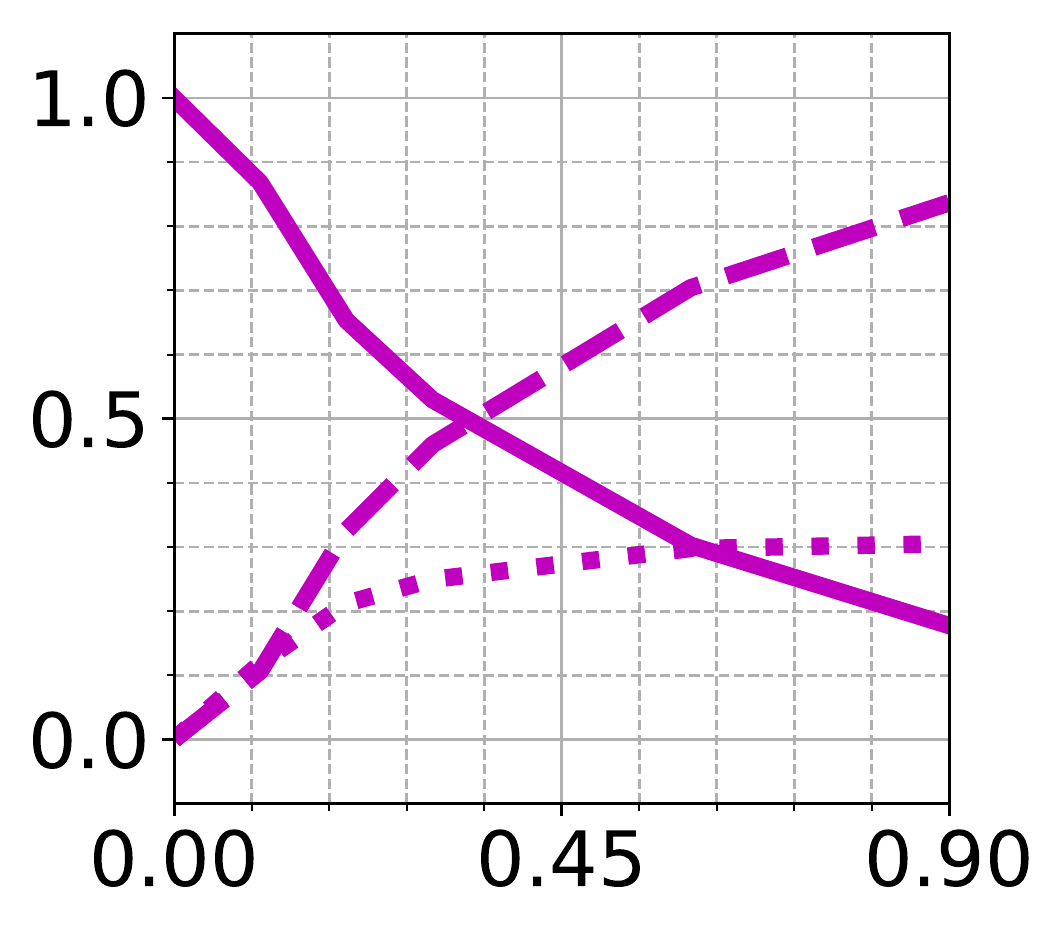}} &
		\fbox{\includegraphics{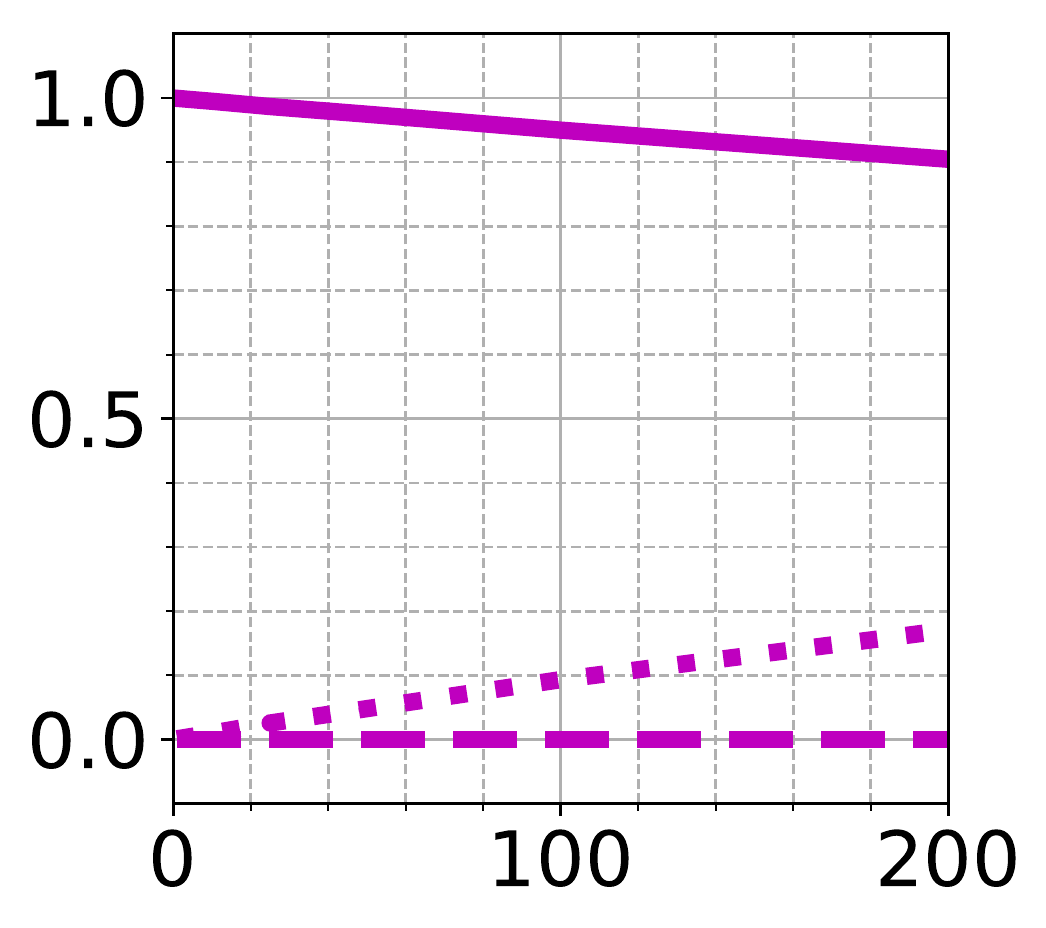}} &
		\fbox{\includegraphics{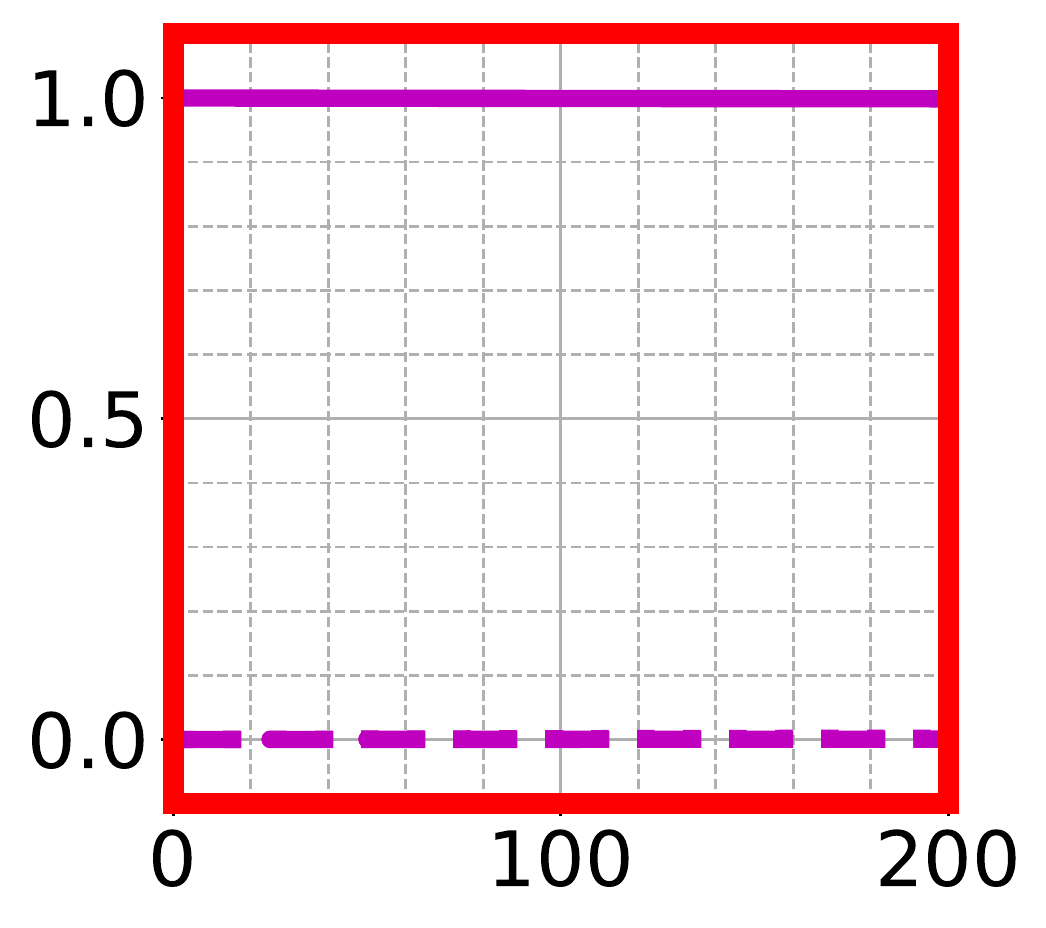}} &
		\fbox{\includegraphics{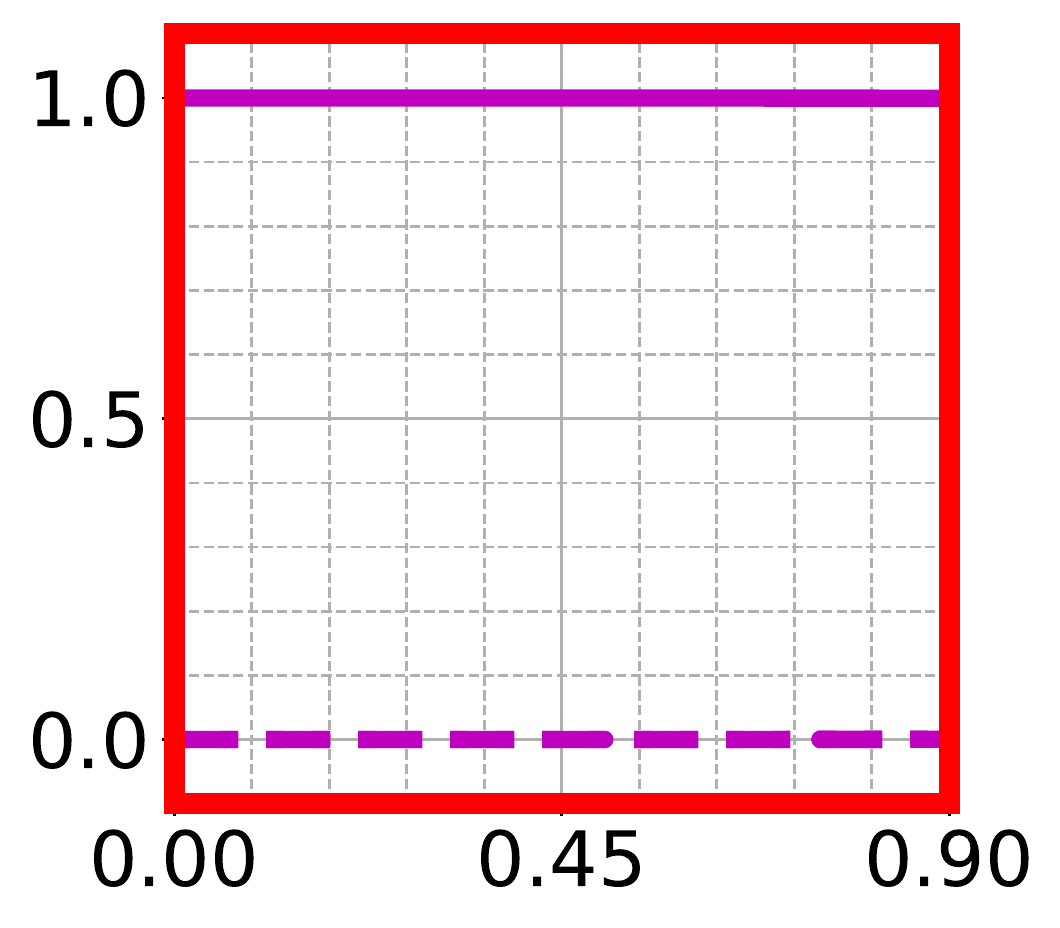}} \\				
		
		\raisebox{4.5cm}{\rotatebox[origin=t]{90}{\HUGE  \JOLD{}}} &
		\fbox{\includegraphics{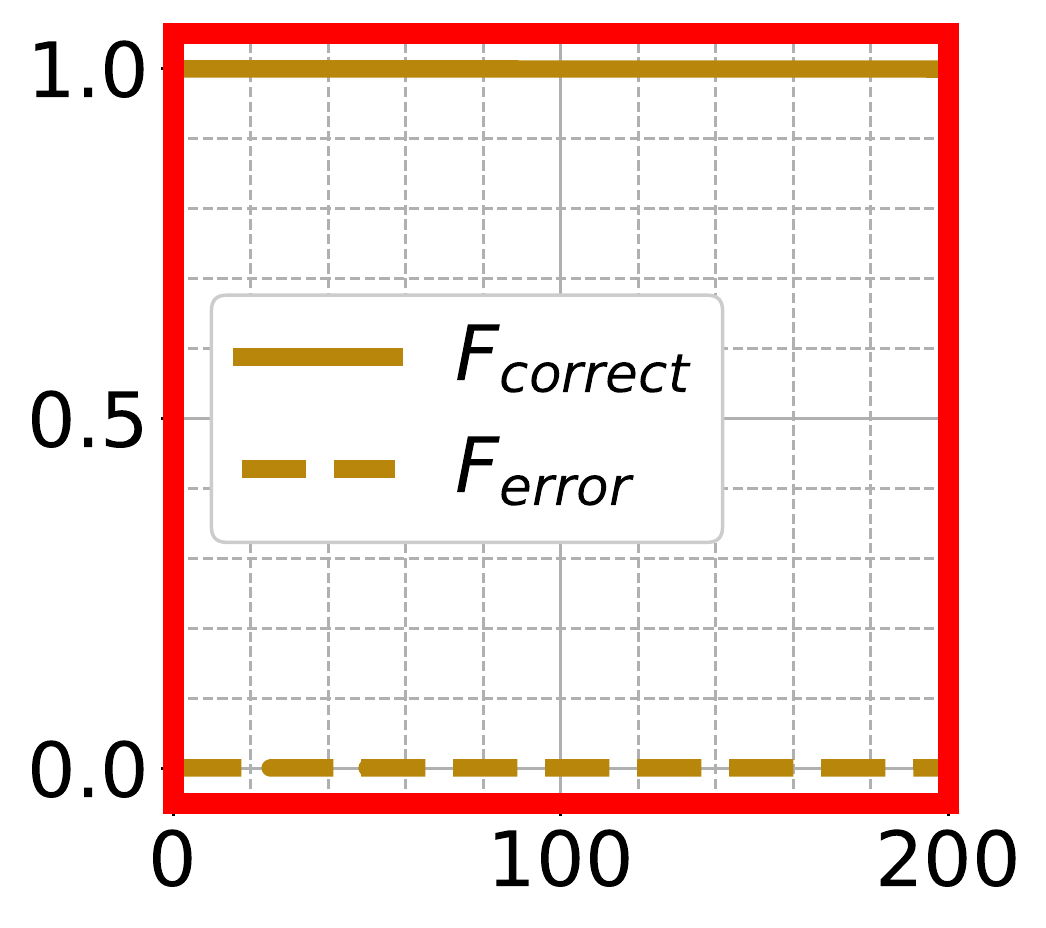}} & 
		\fbox{\includegraphics{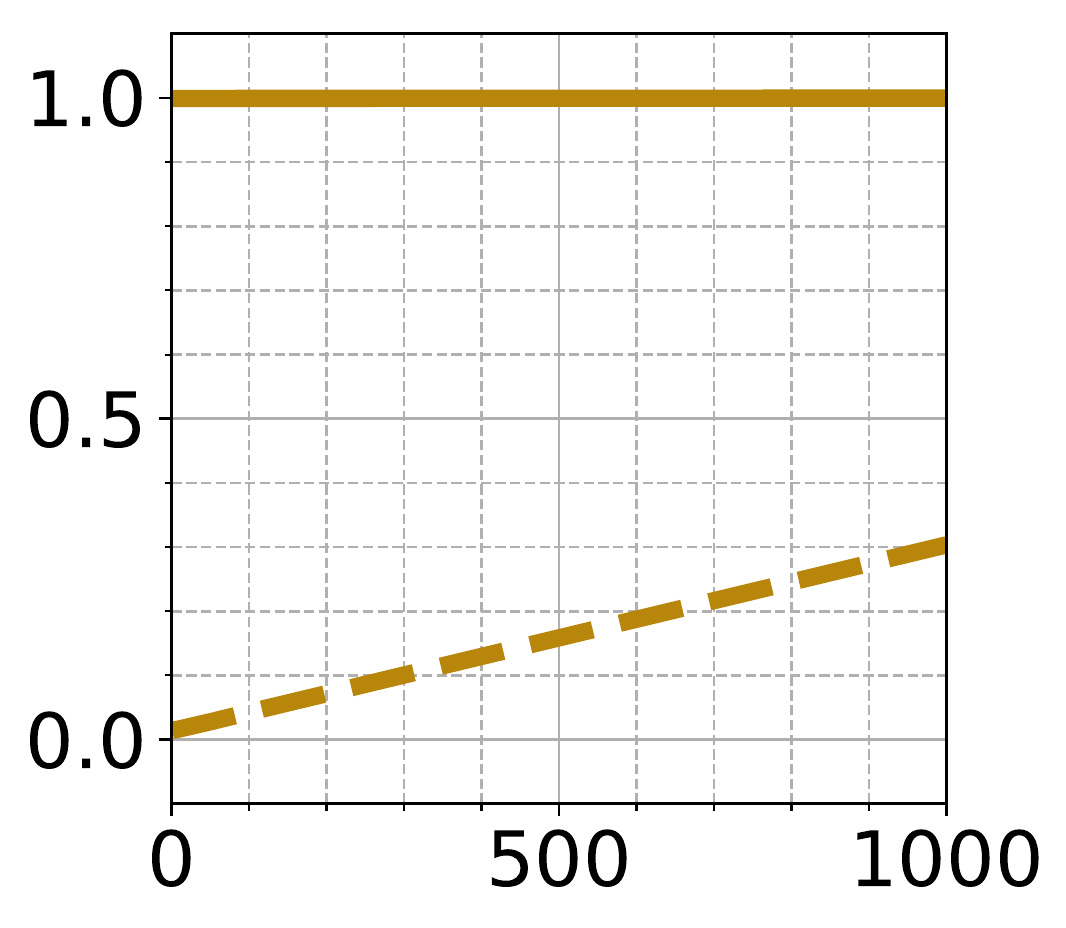}} &
		\fbox{\includegraphics{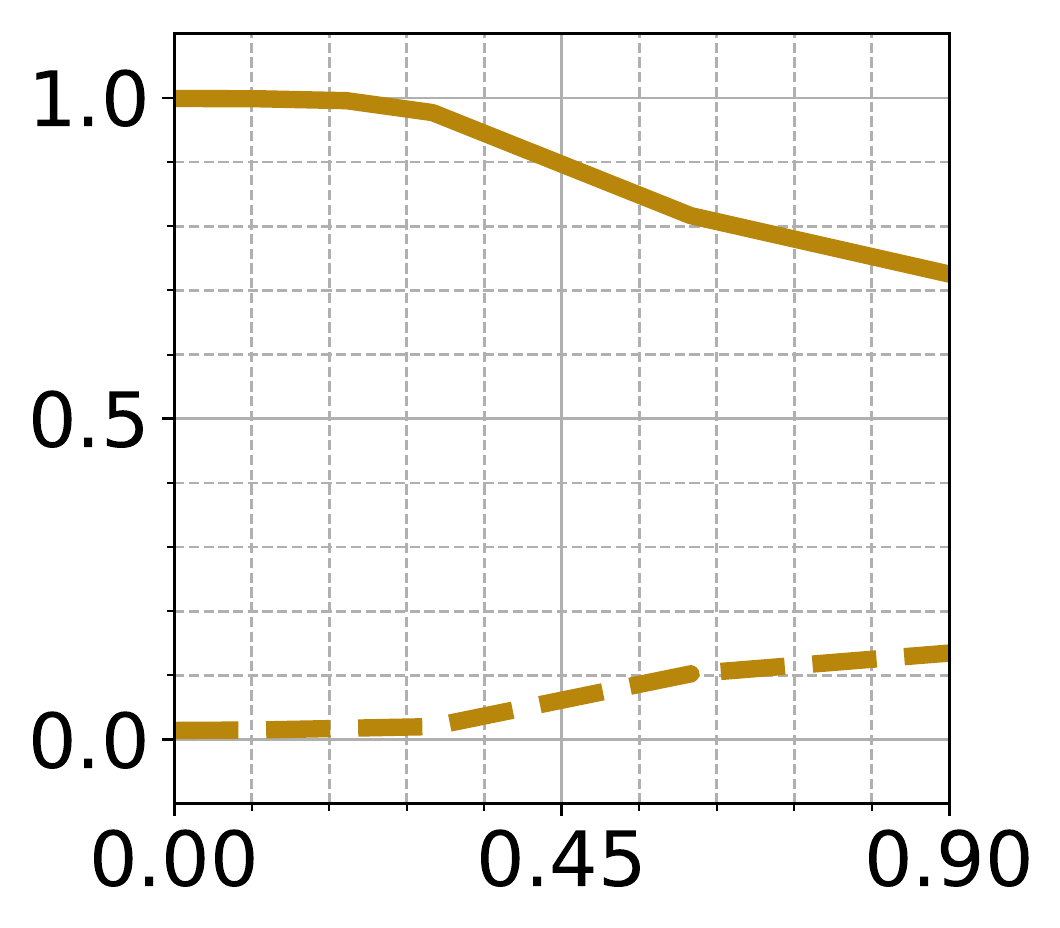}} &
		\fbox{\includegraphics{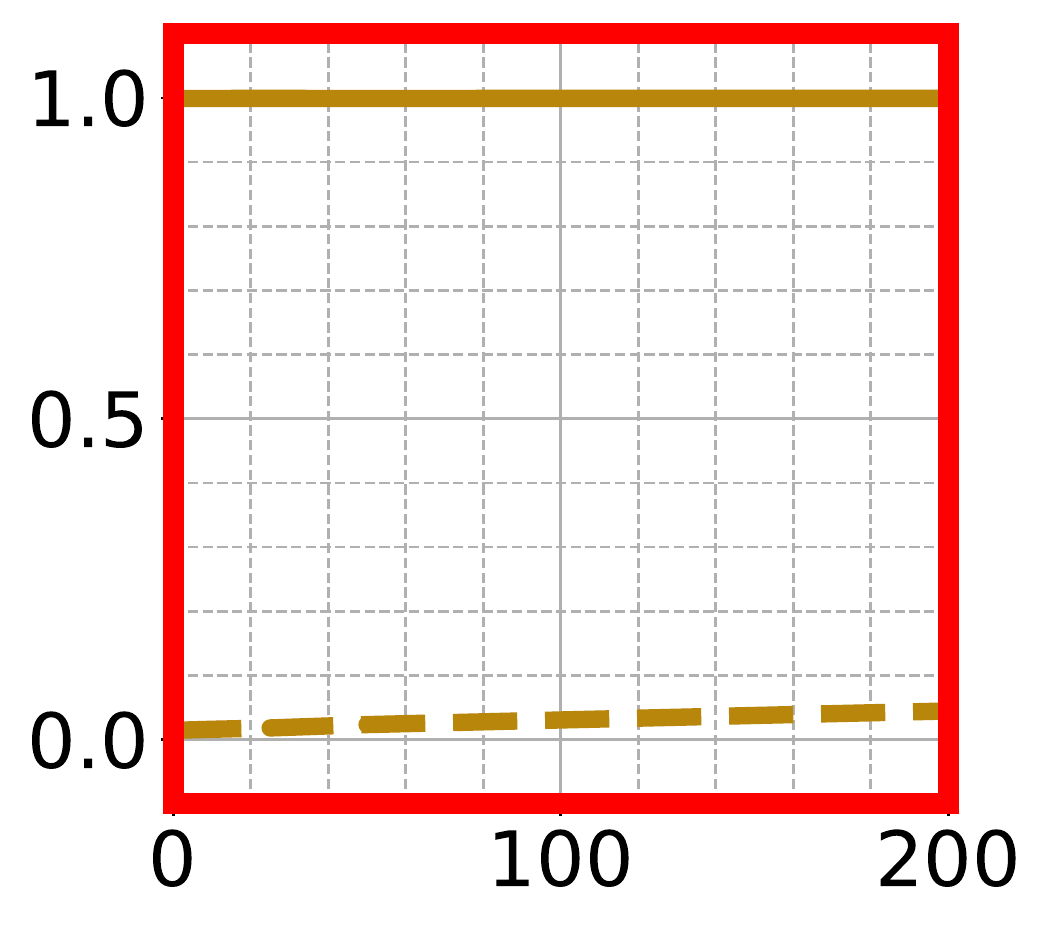}} &
		\fbox{\includegraphics{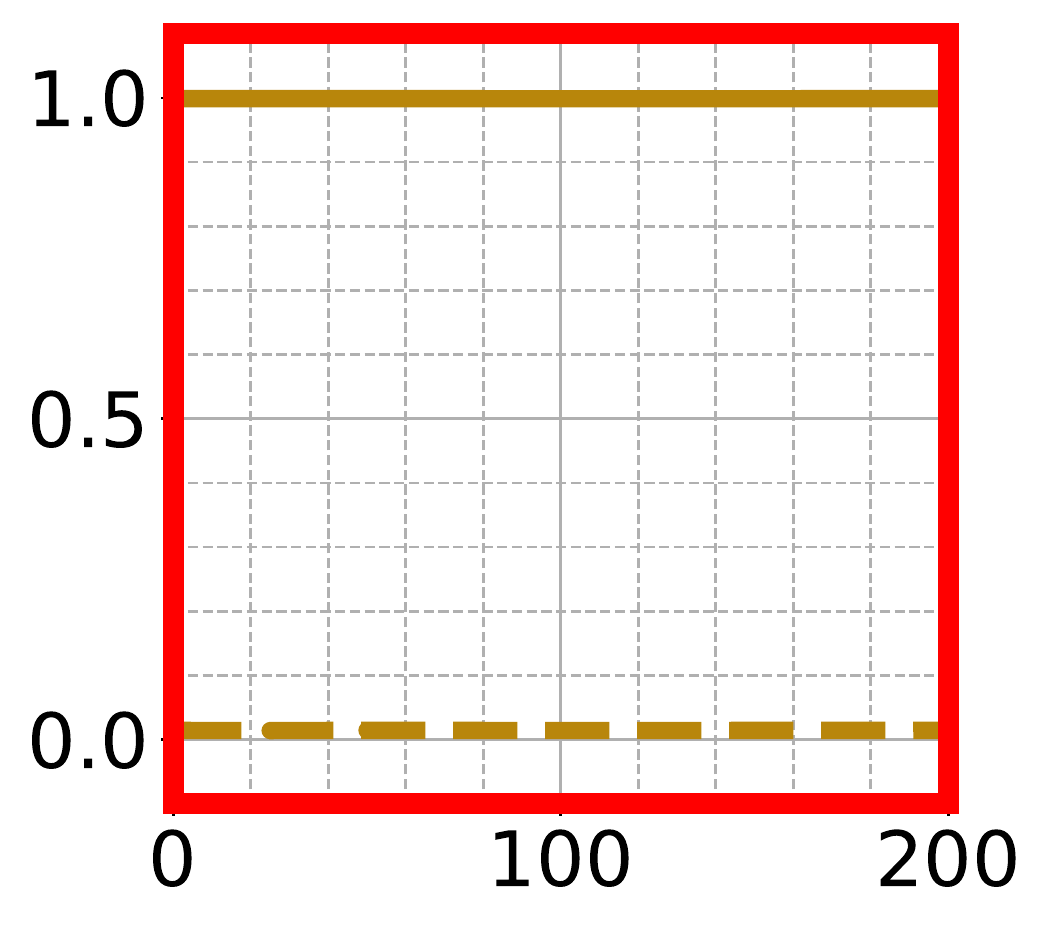}} &
		\fbox{\includegraphics{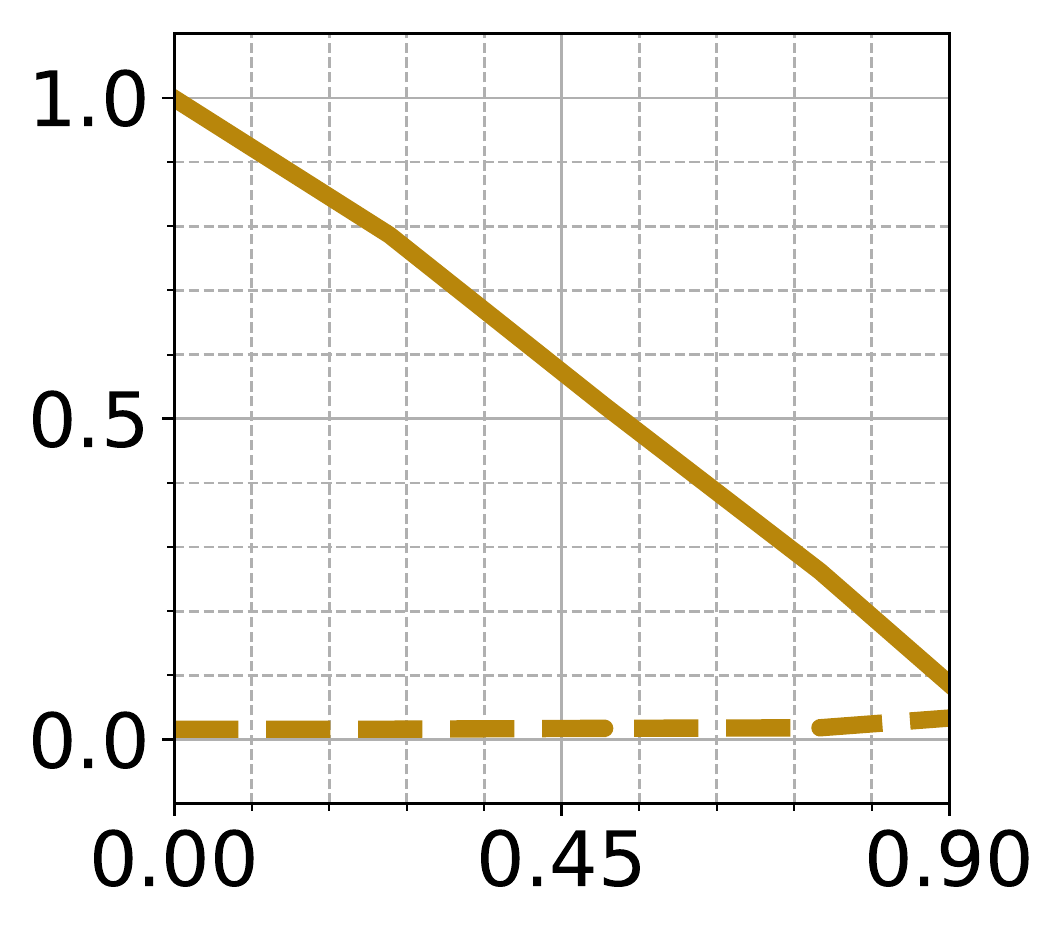}} \\ 					
		
		\raisebox{4.5cm}{\rotatebox[origin=t]{90}{\HUGE  \SEGM{}}} &
		\fbox{\includegraphics{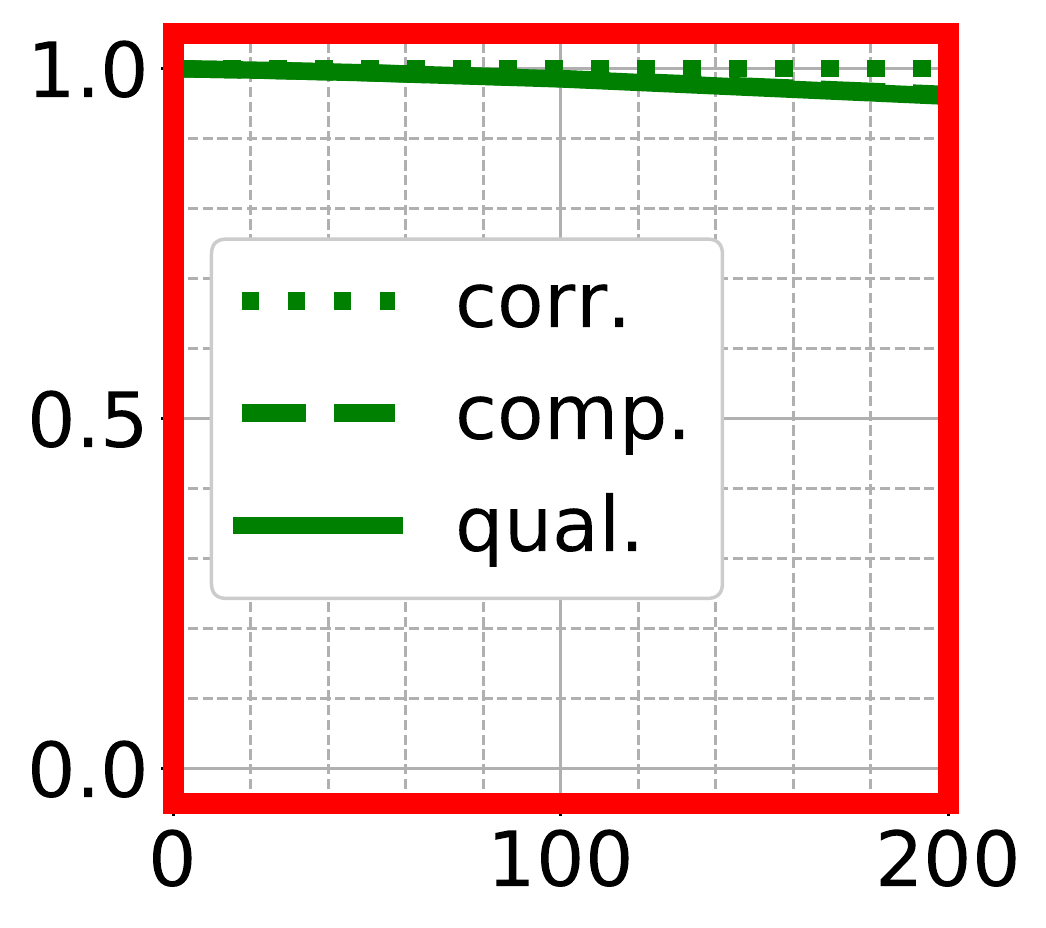}} & 
		\fbox{\includegraphics{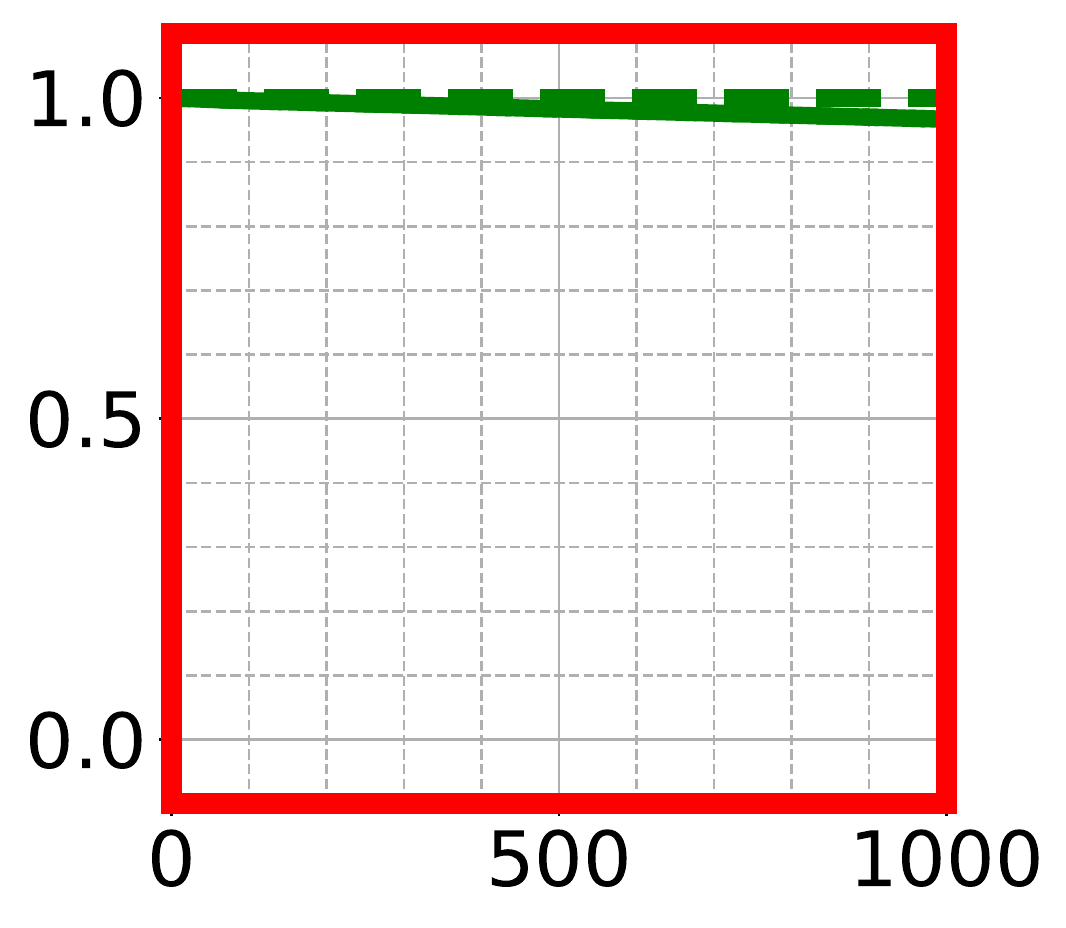}} &
		\fbox{\includegraphics{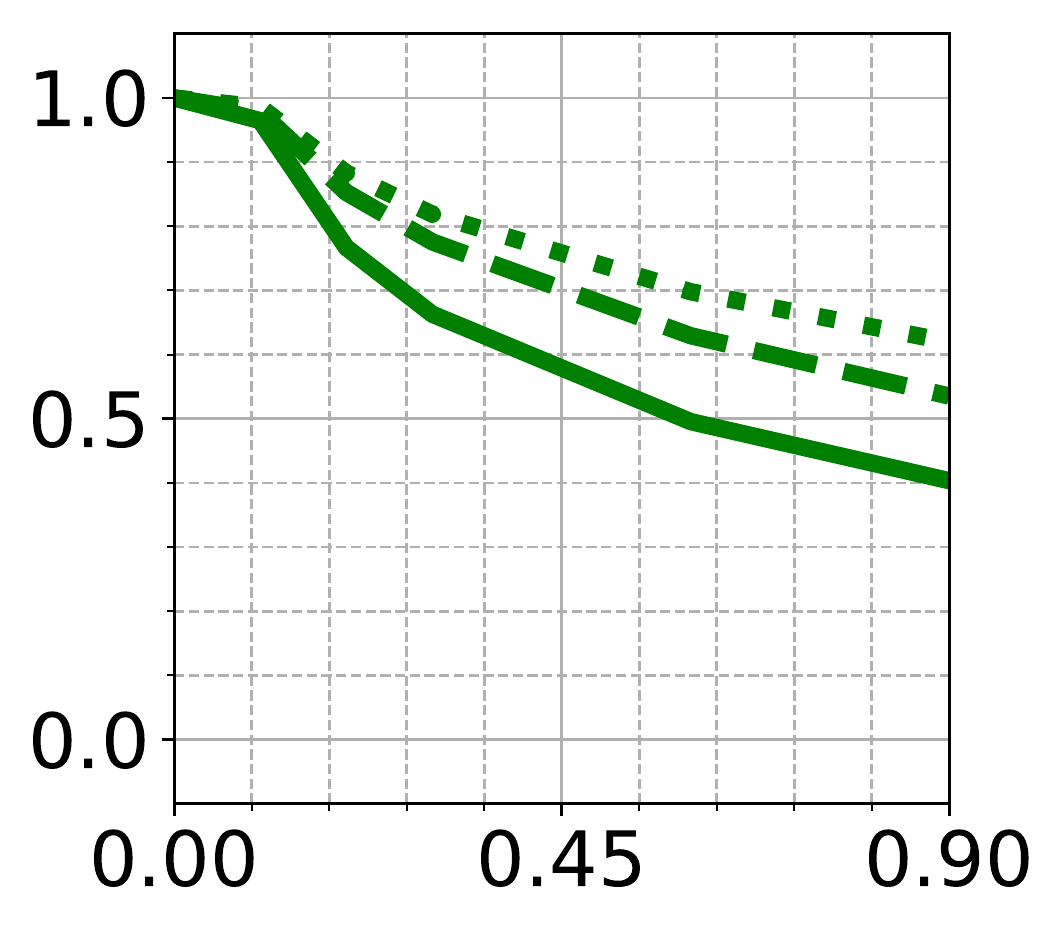}} &
		\fbox{\includegraphics{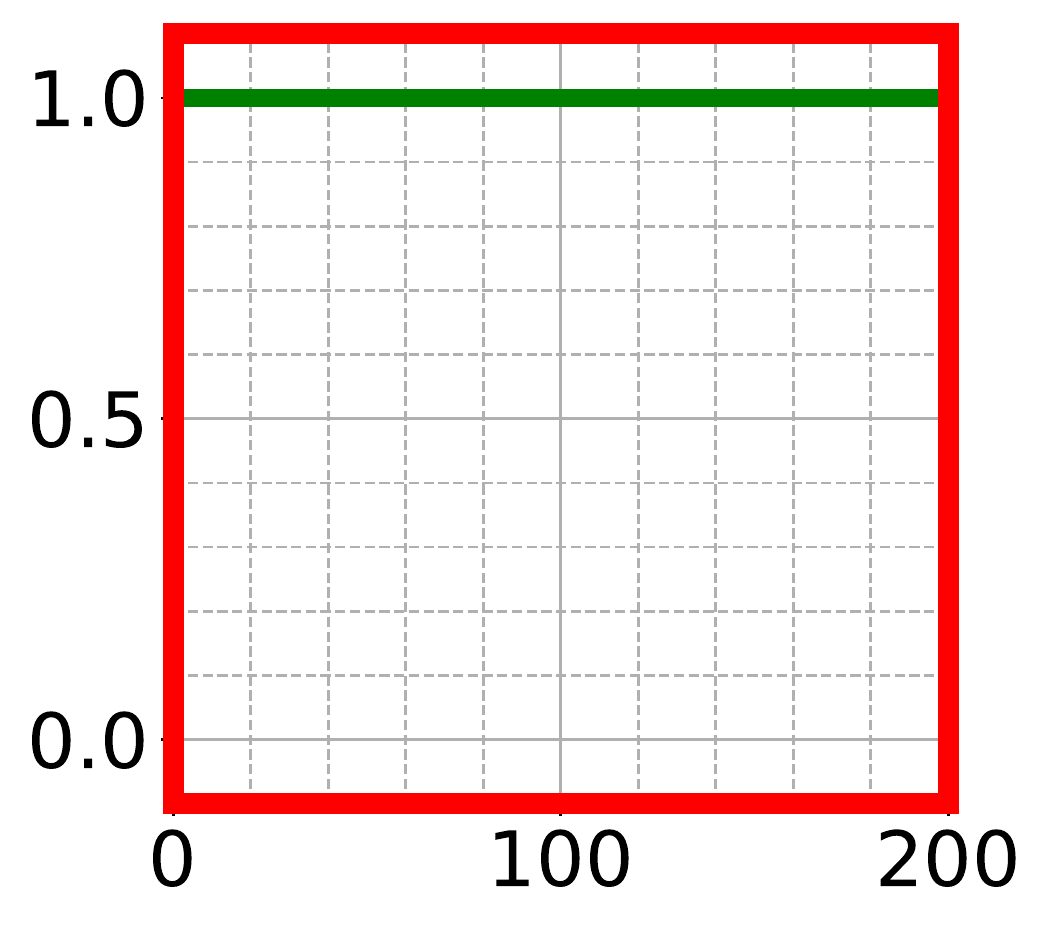}} &
		\fbox{\includegraphics{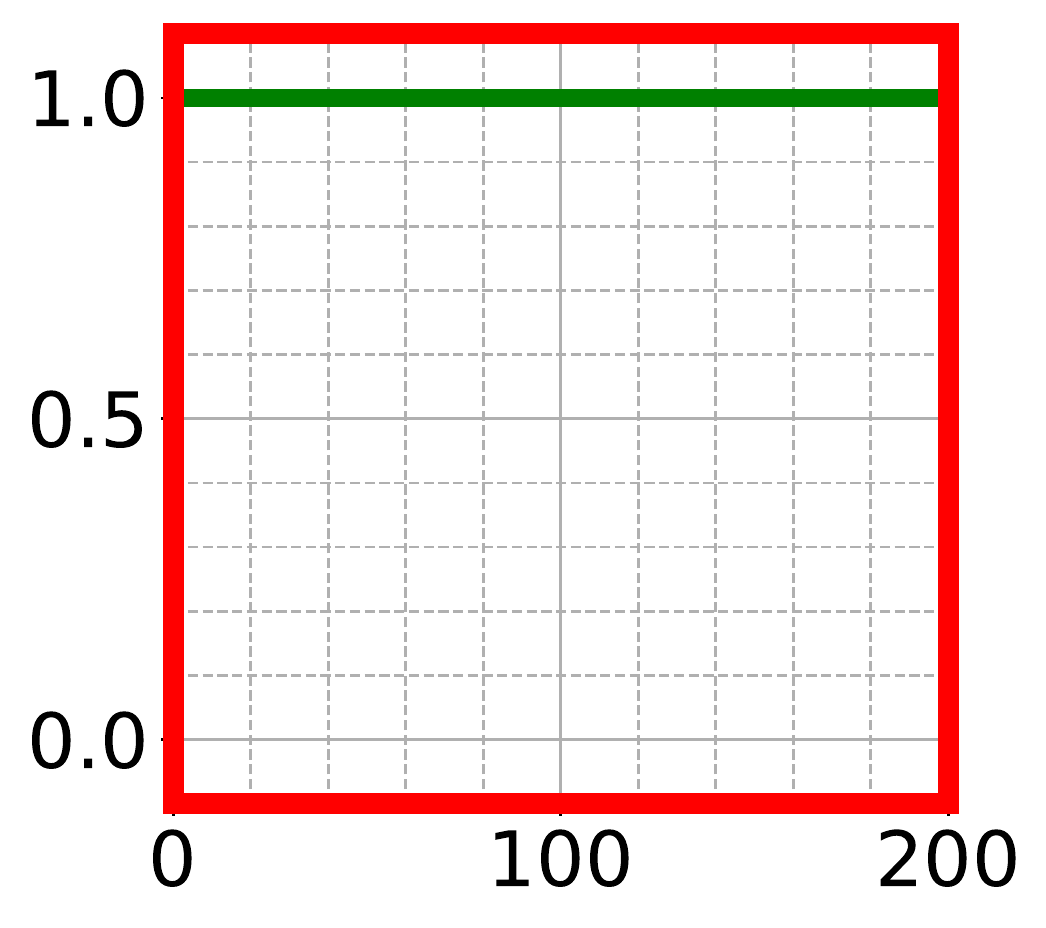}} &
		\fbox{\includegraphics{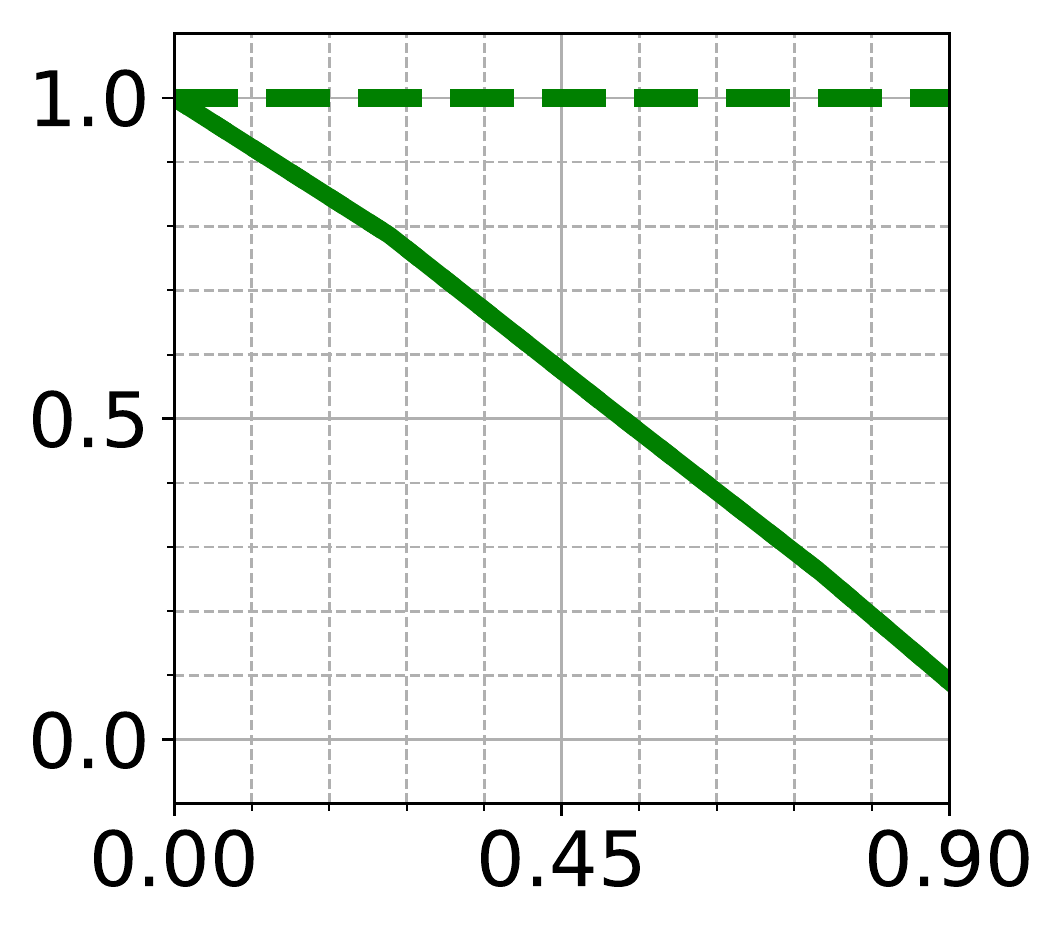}} \\			
                \multicolumn{7}{c}{\multirow{2}{*}{\HUGE \qquad new scores}} \\
                \cmidrule(r{3cm}){2-4}
                \cmidrule(l{2.5cm}){5-7}
                \\
		
		\raisebox{4.5cm}{\rotatebox[origin=t]{90}{\HUGE  \PNEW{}}} &
		\fbox{\includegraphics{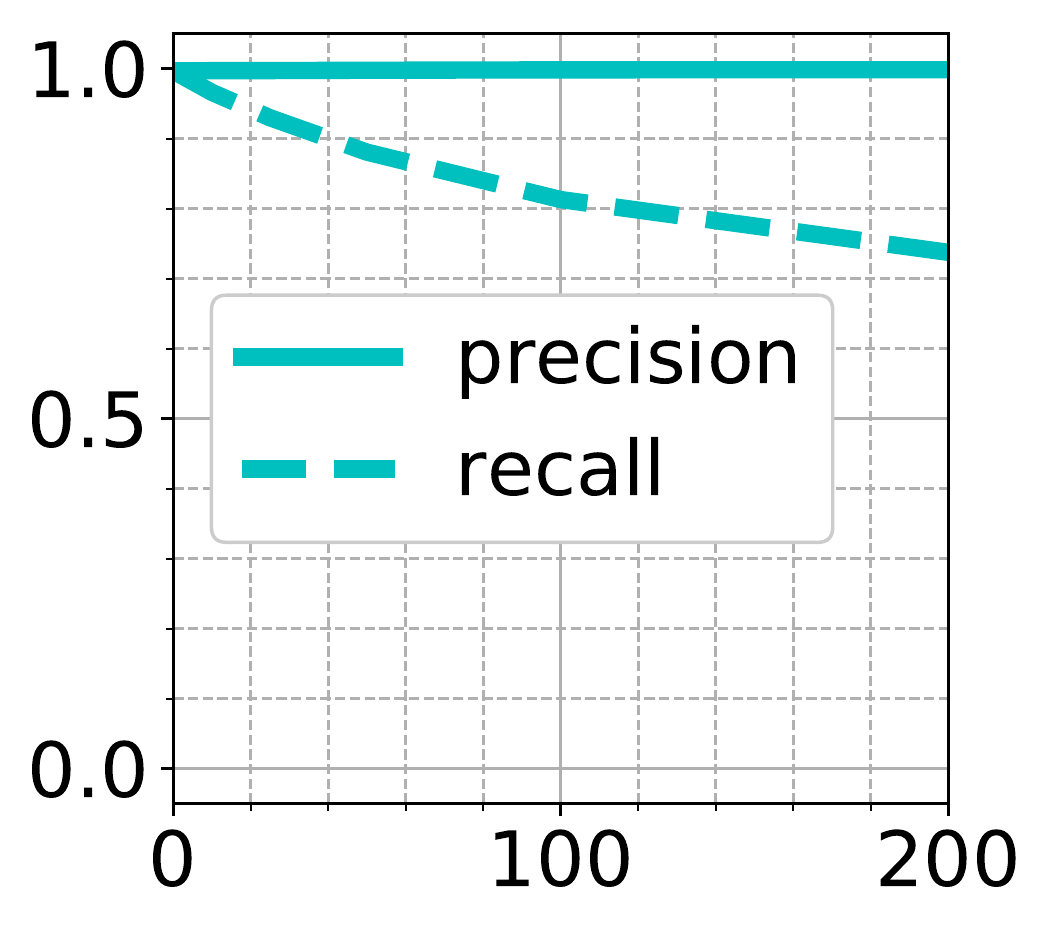}} & 
		\fbox{\includegraphics{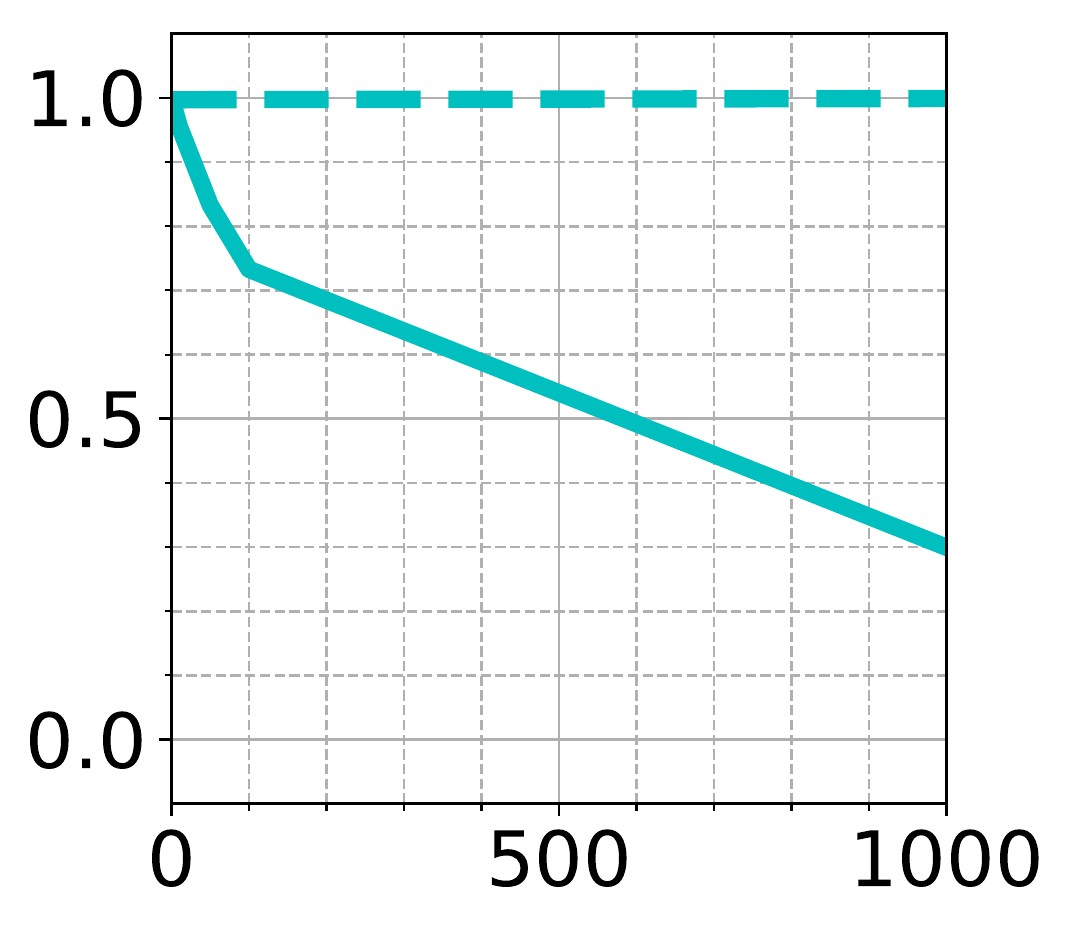}} &
		\fbox{\includegraphics{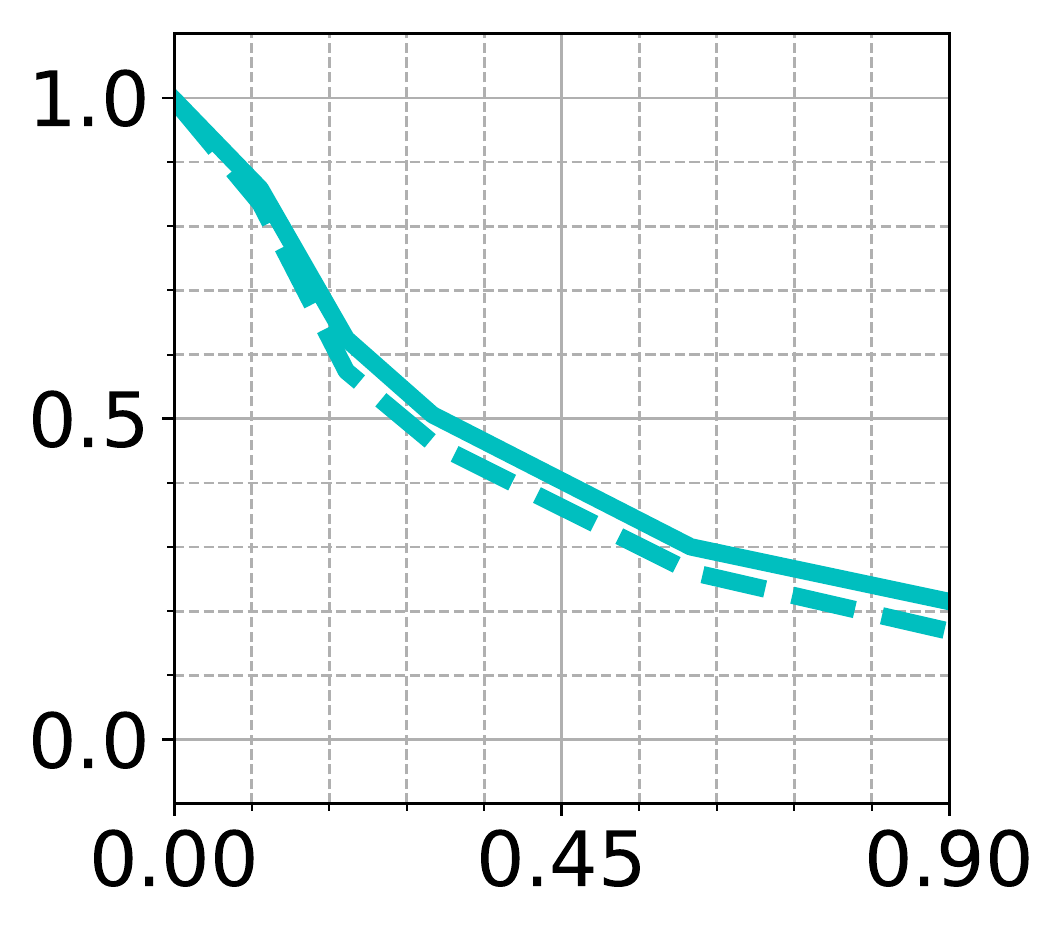}} &
		\fbox{\includegraphics{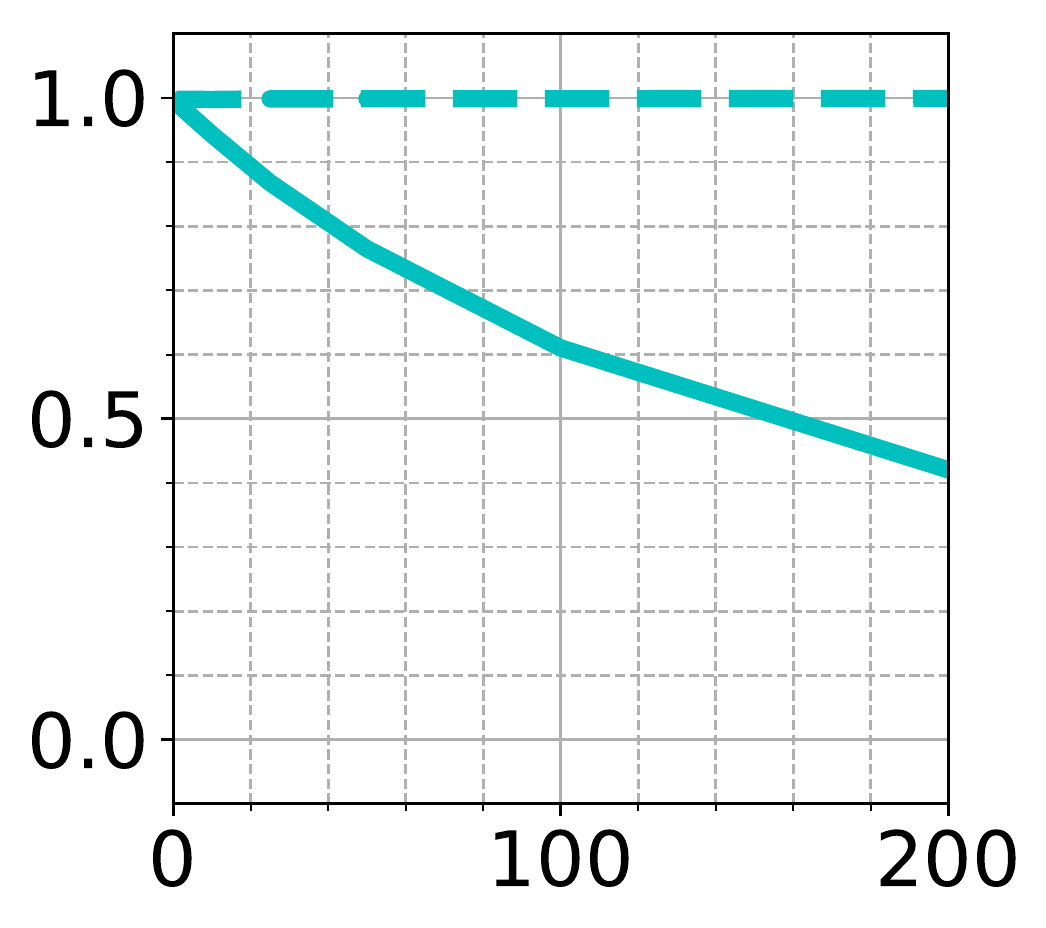}} &
		\fbox{\includegraphics{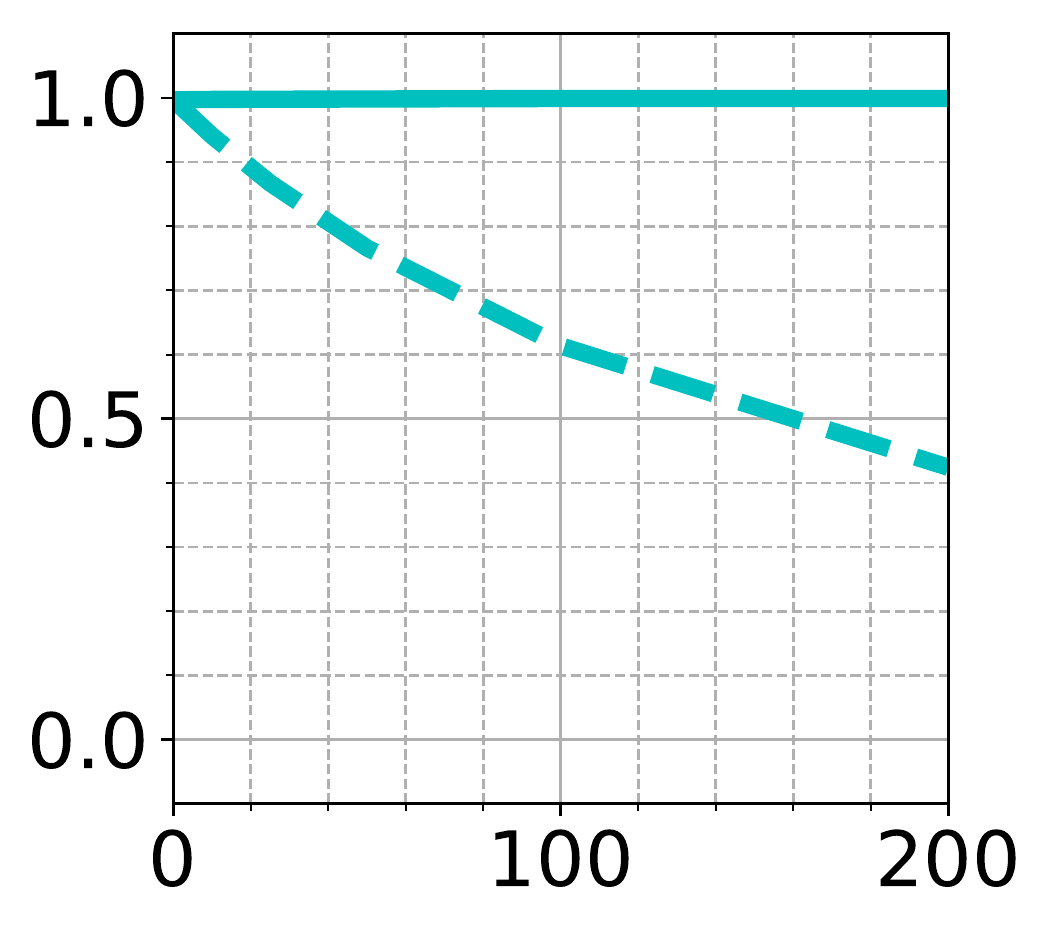}} &
		\fbox{\includegraphics{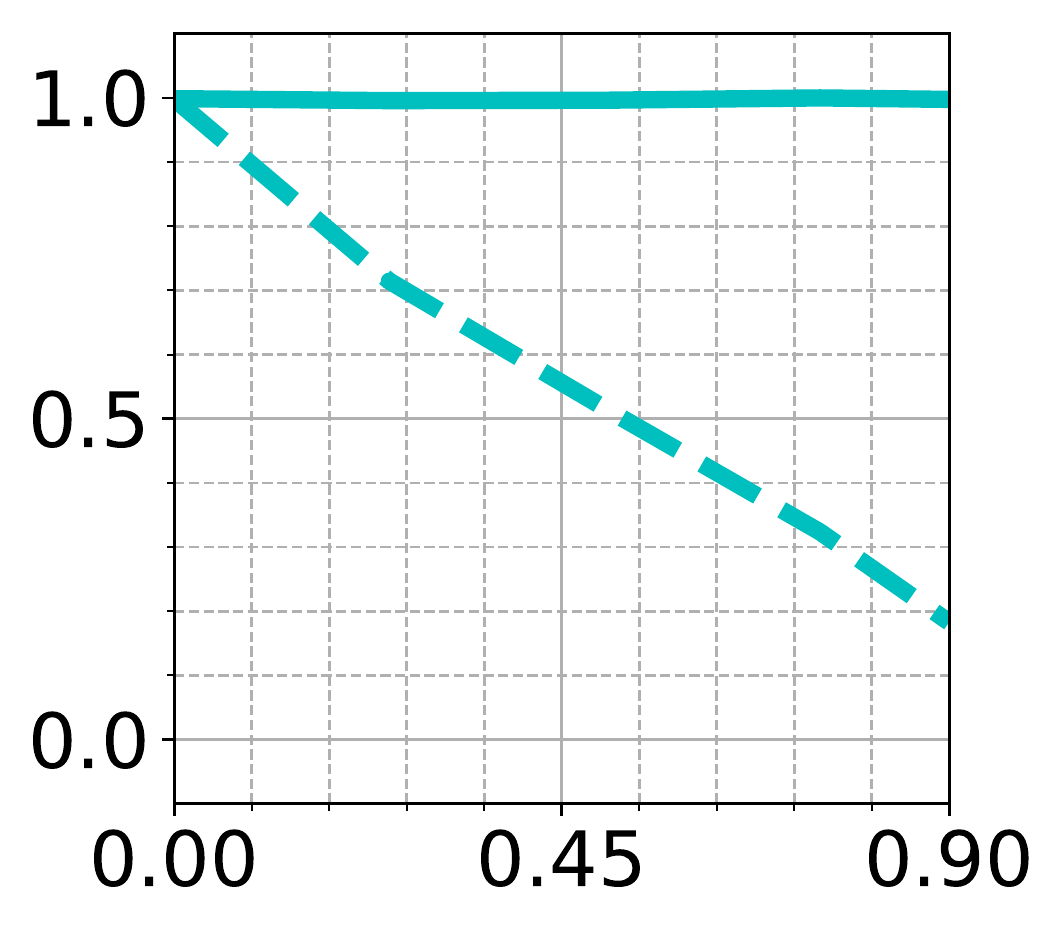}} \\
		
		\raisebox{4.5cm}{\rotatebox[origin=t]{90}{\HUGE  \JNEW{}}} &
		\fbox{\includegraphics{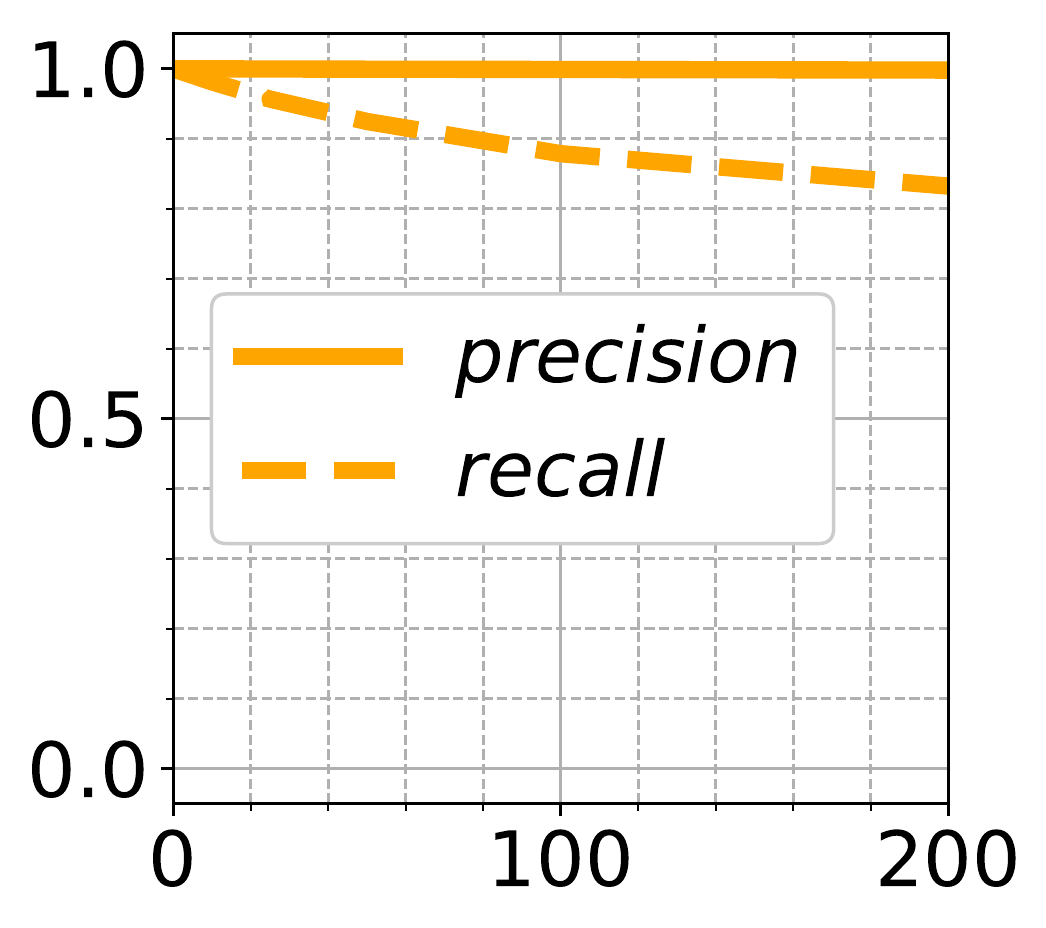}} &
		\fbox{\includegraphics{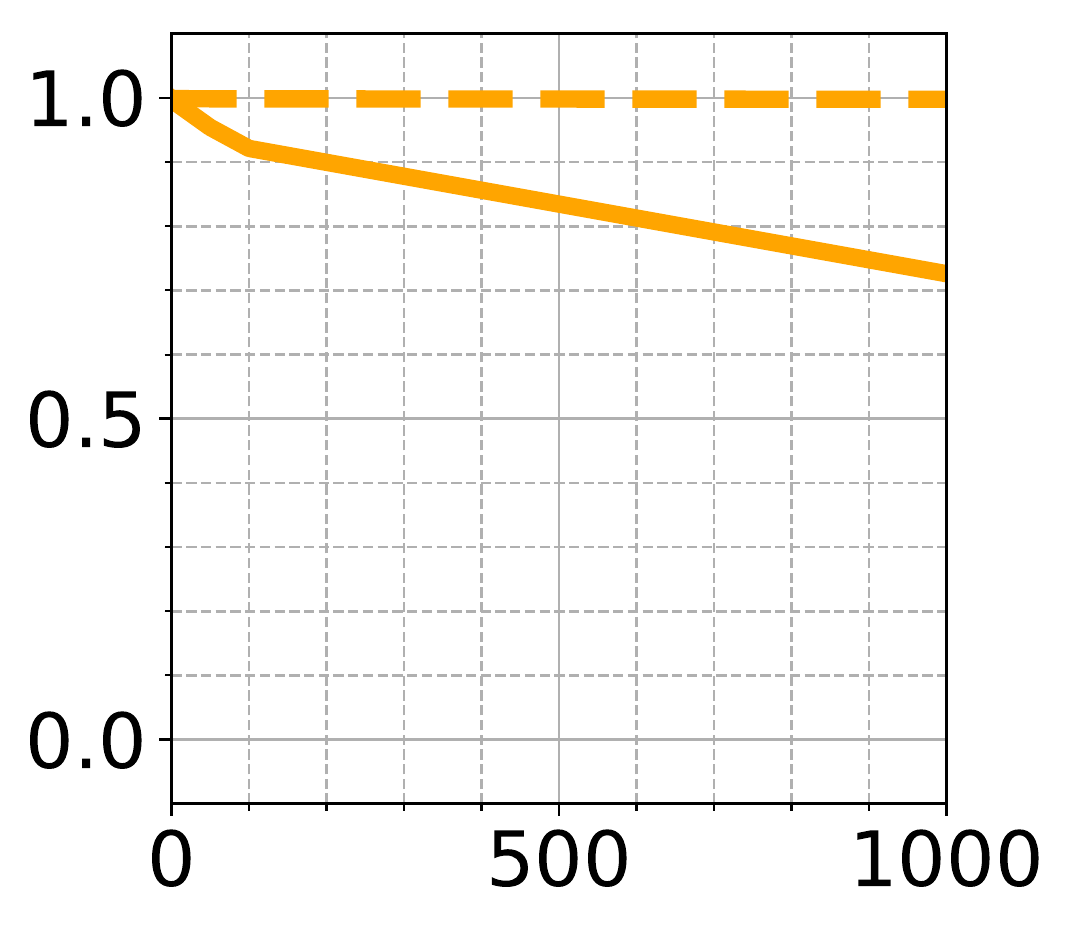}} &
		\fbox{\includegraphics{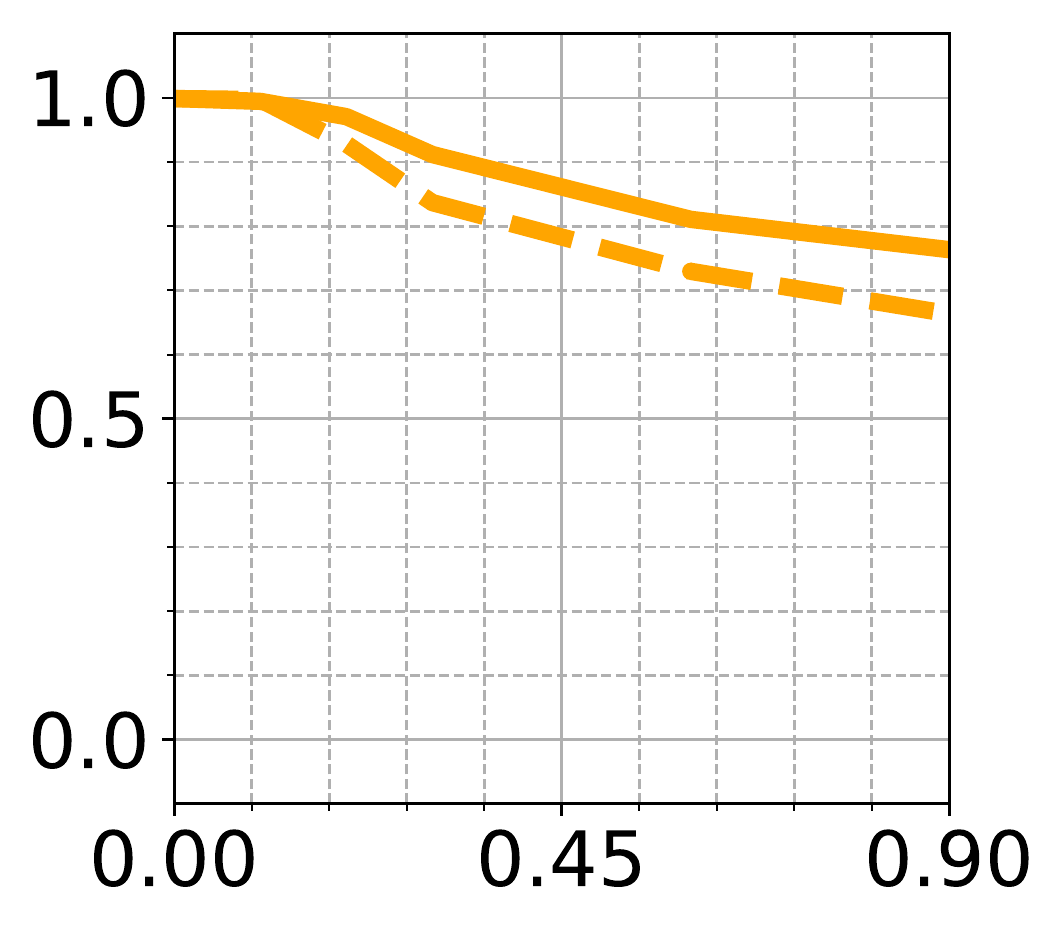}} &
		\fbox{\includegraphics{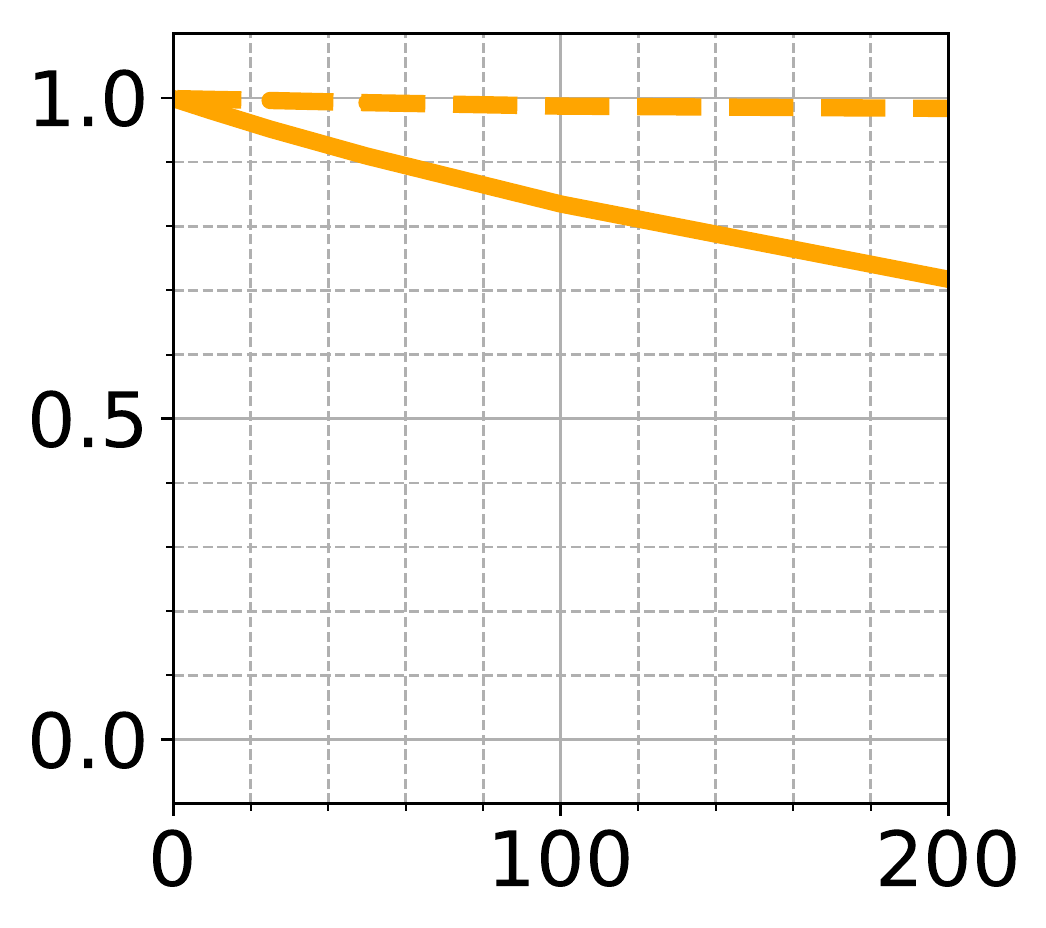}} &
		\fbox{\includegraphics{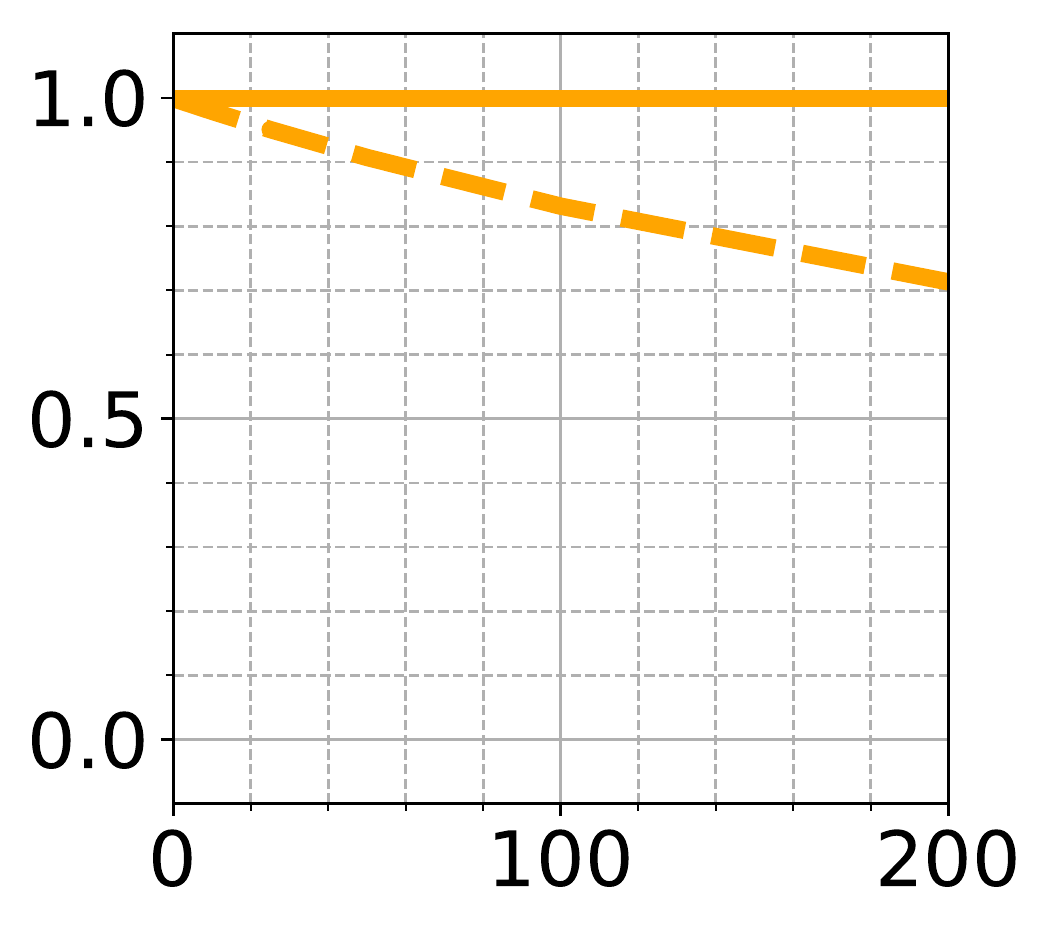}} &
		\fbox{\includegraphics{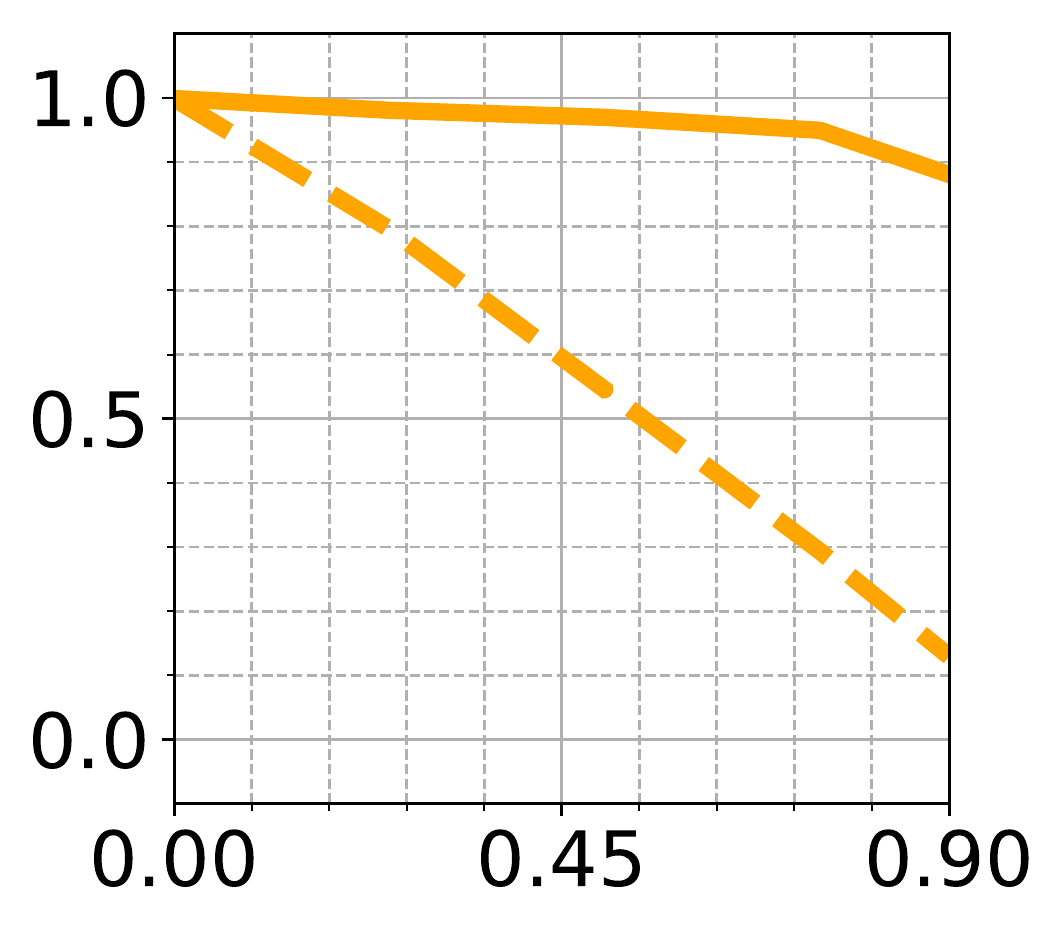}} \\		
		
		\raisebox{4.5cm}{\rotatebox[origin=t]{90}{\HUGE  \GNEW{}}} &
		\fbox{\includegraphics{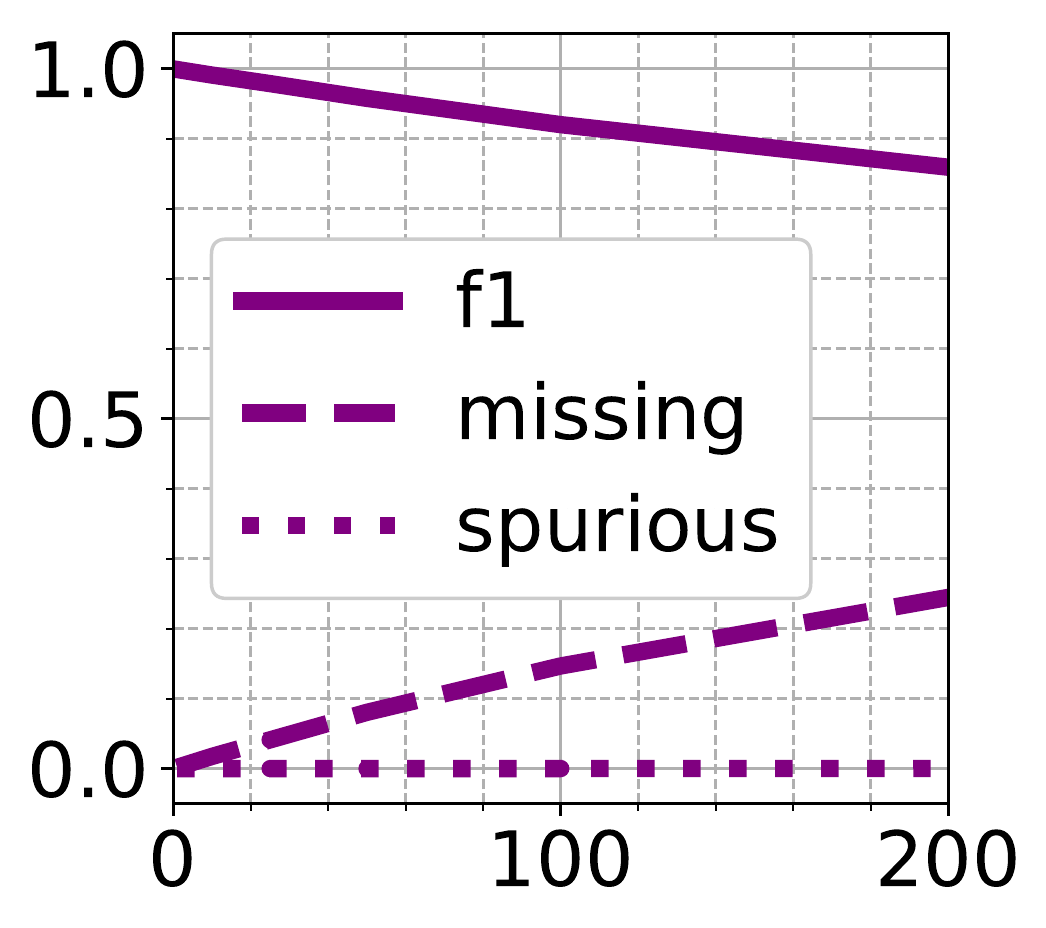}} &
		\fbox{\includegraphics{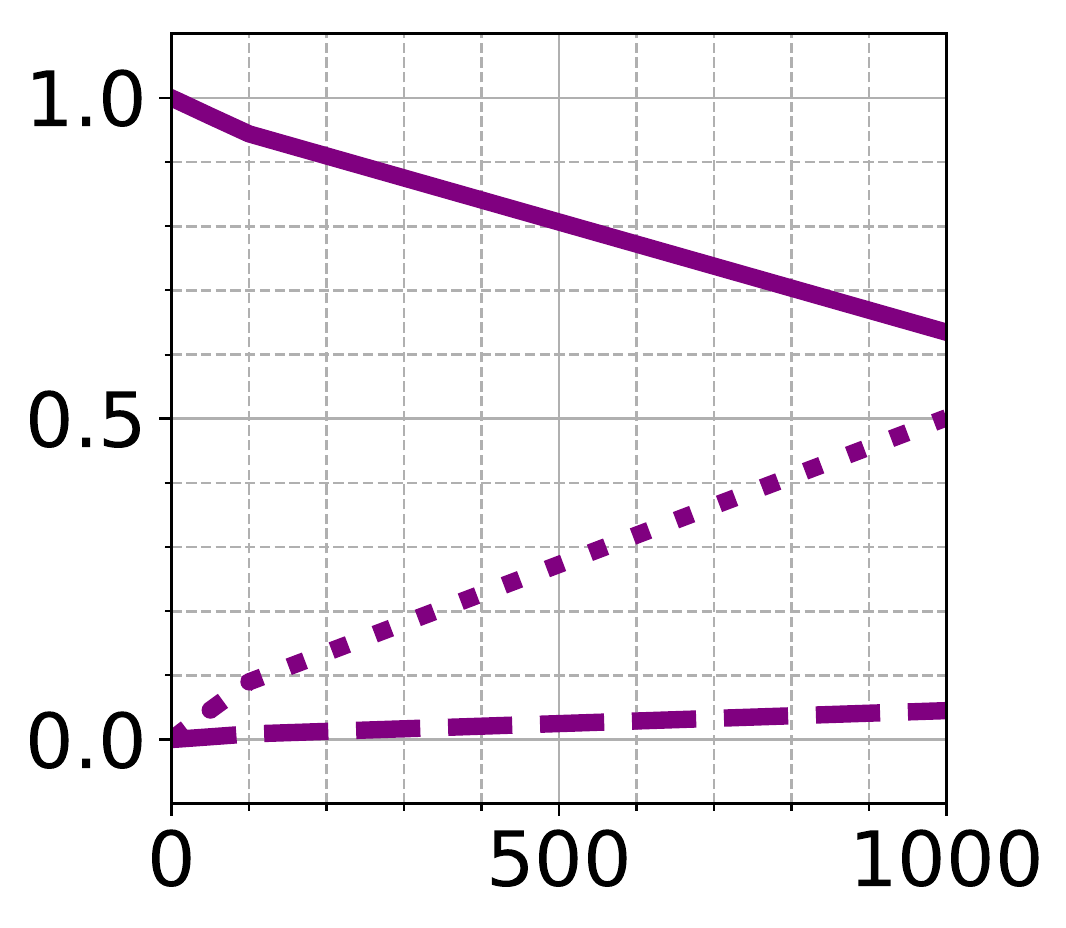}} &
		\fbox{\includegraphics{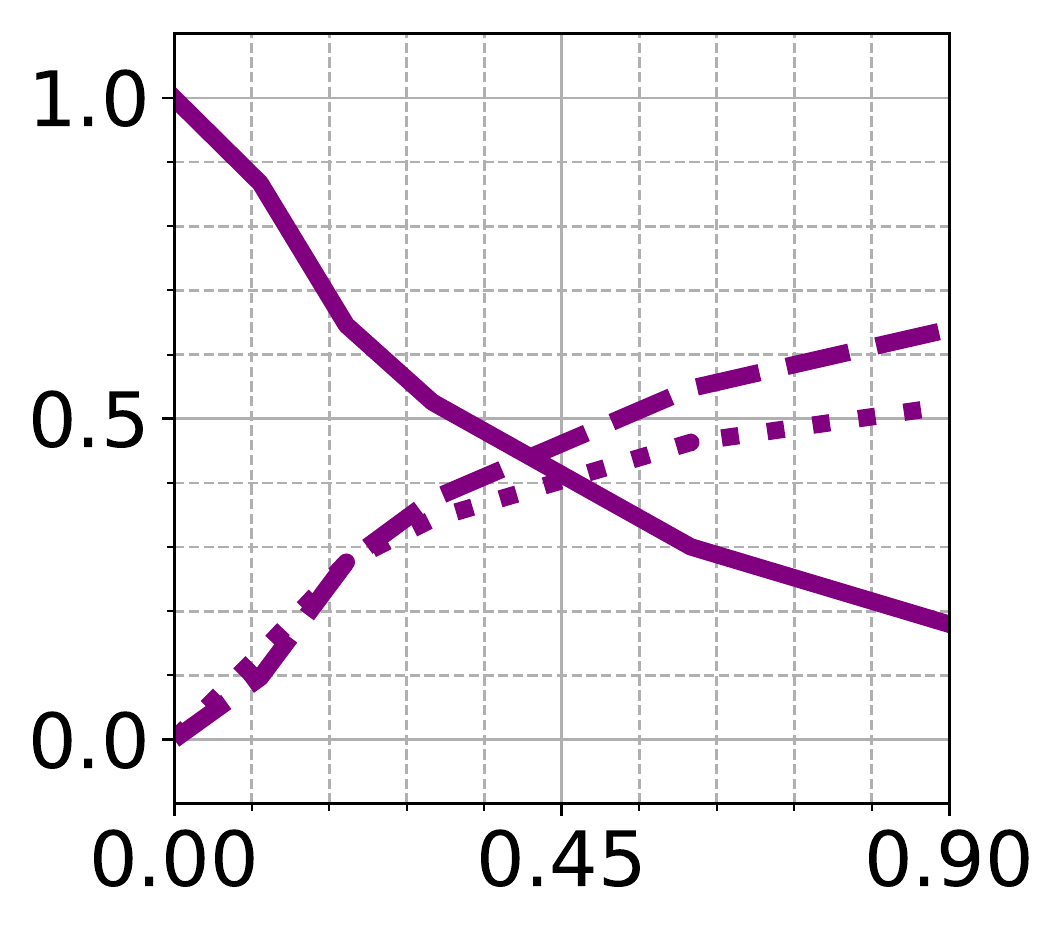}} &
		\fbox{\includegraphics{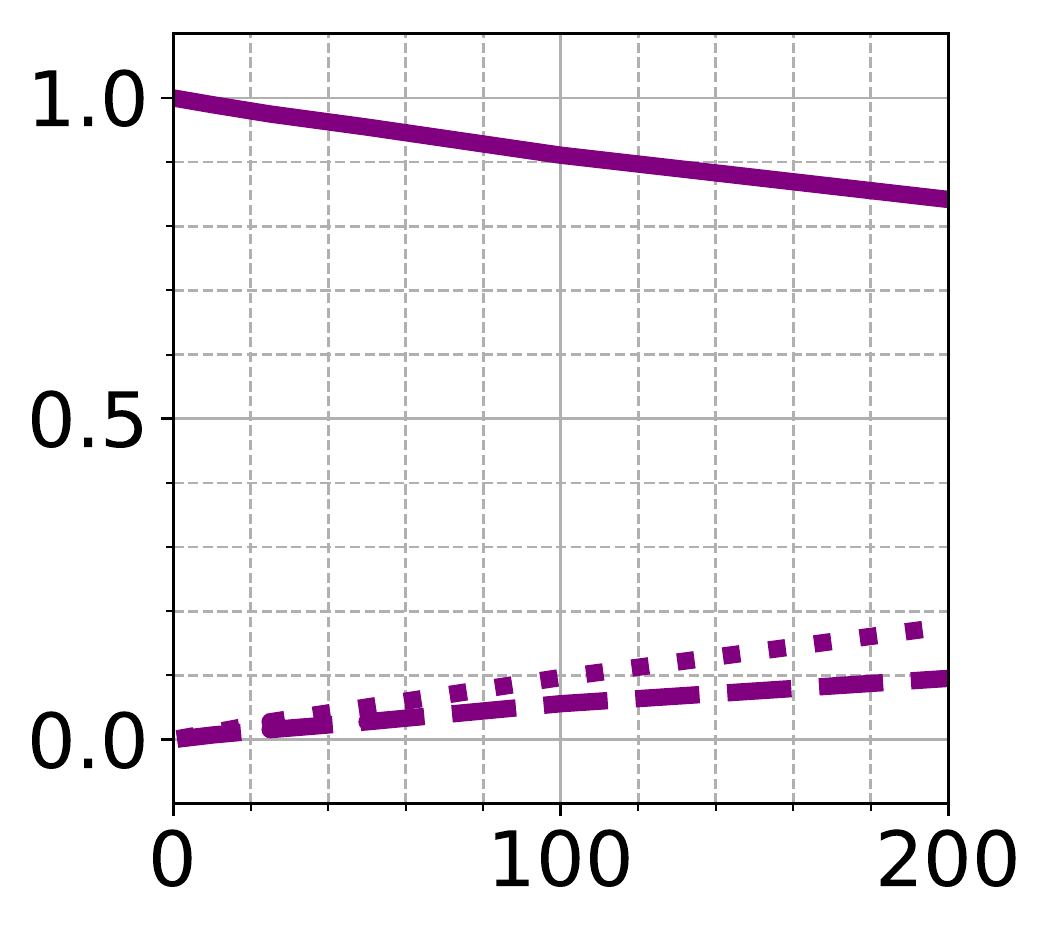}} &
		\fbox{\includegraphics{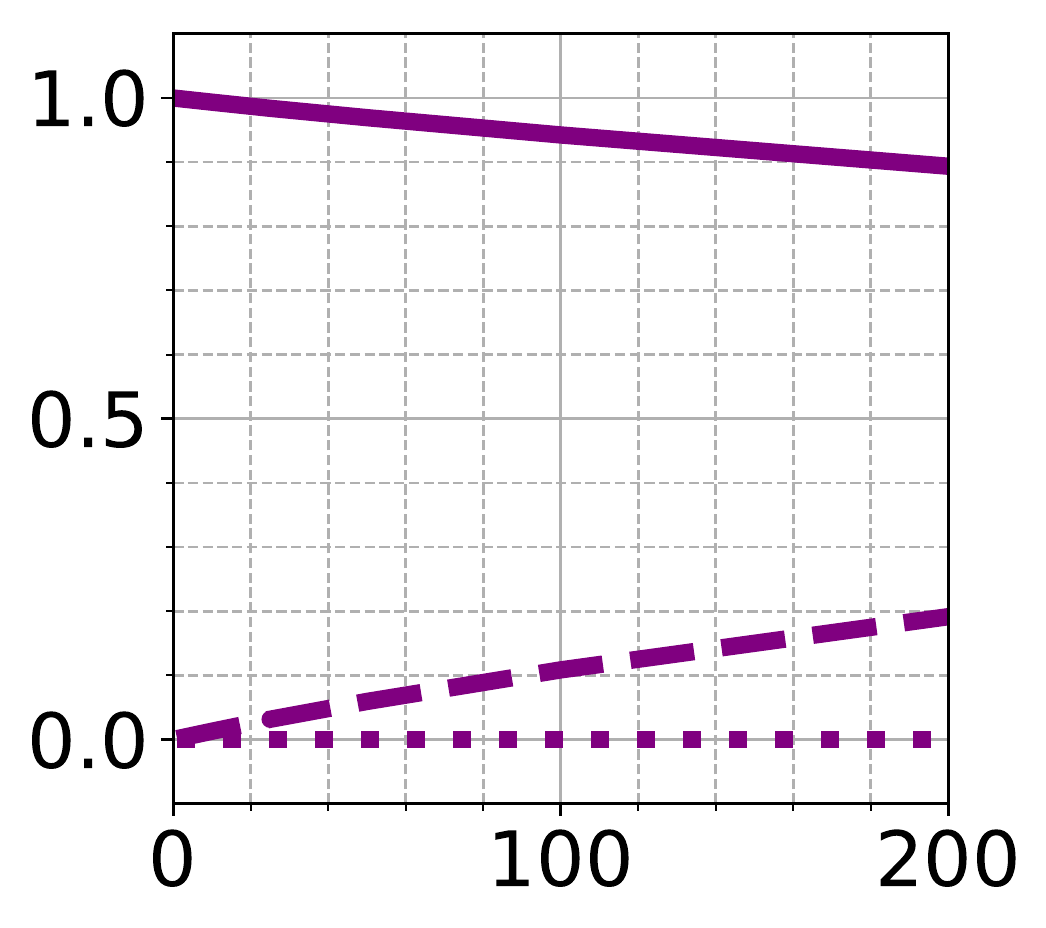}} &
		\fbox{\includegraphics{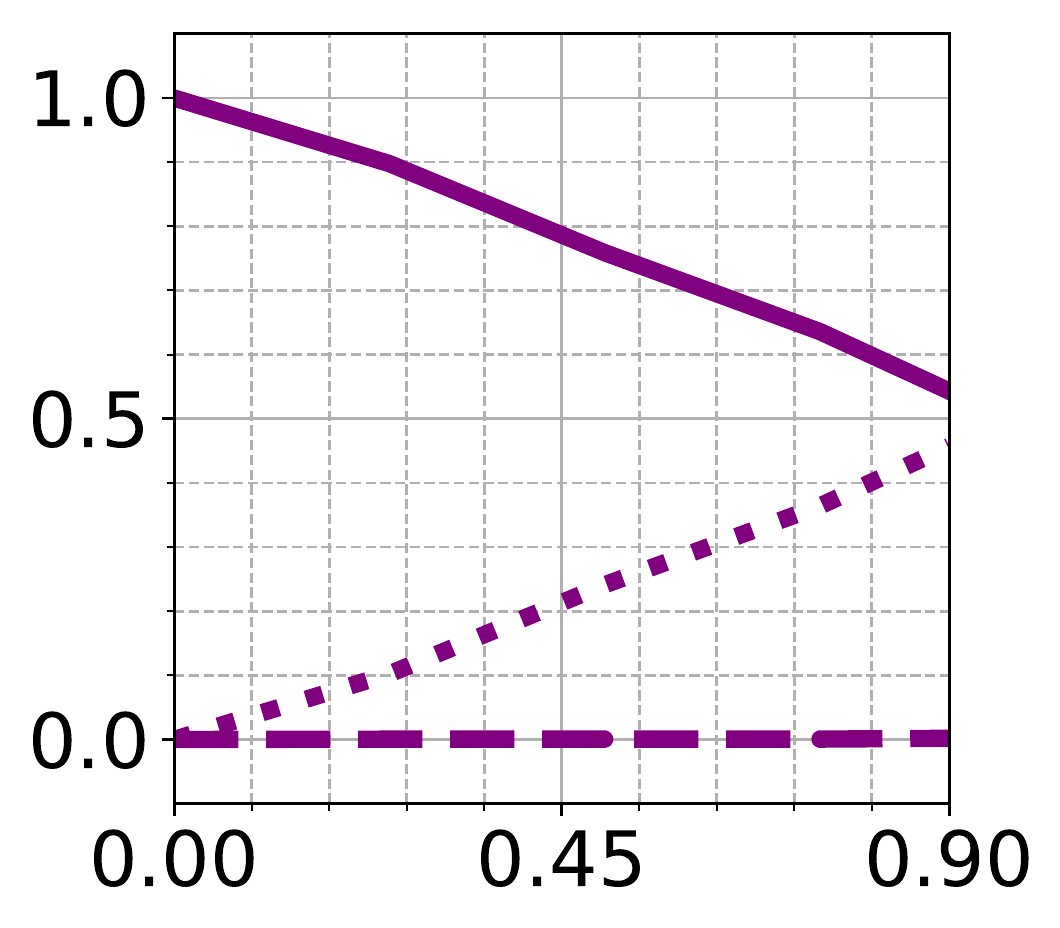}} \\									 

                \HUGE\bf x-axis: &
                \HUGE \quad no.\ interruptions & 
                \HUGE \; no.\ additional connections & 
                \HUGE \: perturbation magnitude & 
                \HUGE \quad no.\ doubled roads & 
                \HUGE \quad no.\ missed roads & 
                \HUGE \quad frac.\ map removed  
                \\
	      
	  \end{tabular}
  }
  \caption{Sensitivity of the existing and the new scores to different types of errors. 
           The plots that demonstrate lack of sensitivity of existing errors are outlined in red. 
           See section~\ref{sec:synth} for details. Best viewed in color. \label{fig:benchmark_plots}}
  
\end{figure*}

\begin{table*}[t]
\centering
\resizebox{\linewidth}{!}{%
	\setlength{\tabcolsep}{3pt}
	\begin{tabular}{@{}  l l p{0.2cm}  ccc c ccc c cc c c c c p{0.2cm} c c ccc c ccc @{} }

& & &
\multicolumn{14}{c}{existing scores} & &
\multicolumn{9}{c}{new scores} \\
\cmidrule{4-17} 	
\cmidrule{19-27}

& & &
\multicolumn{3}{c}{\small \SEGM{}} & &
\multicolumn{2}{c}{\small \TLTS{}} &&
\multicolumn{1}{c}{\small \makebox[\widthof{ABC}][c]{\APLS{}}} & &
\multicolumn{3}{c}{\small \JOLD{}} & &
\multicolumn{1}{c}{\small \makebox[\widthof{ABC}][c]{\GOLD{}}} &&
\multicolumn{3}{c}{\small \PNEW{}} &&
\multicolumn{3}{c}{\small \JNEW{}} &&
\multicolumn{1}{c}{\small \makebox[\widthof{ABC}][c]{\GNEW{}}} \\

\multicolumn{2}{l}{\rotatebox[origin=b]{90}{$\Lsh$}dataset} & &
corr. & comp. & qual. & &
corr. & 2l+2s & &
& &
$F_{cor}$ & $F_{err}$ & f1 & &
f1 && 
pre. & rec. & f1 && 
 pre. & rec. & f1  & &
f1 \\
\cmidrule{1-2} 	
\cmidrule{4-17} 	
\cmidrule{19-27}

\multirow{5}{*}{\rotatebox{90}{RoadTracer}}&
\RTracer{} \cite{Bastani18} & &
0.682 &       0.543 &       0.431 &&  
0.286 &       0.130 && 
0.536 & & 
0.732 &  \bf{0.111} &  \bf{0.803} &&  
0.647 & & 
0.600 &       0.388 &       0.472 &&  
0.777 &       0.687 &       0.729 & & 
0.618 \\ 

&
\Segm{}   \cite{Bastani18}   & &
0.774 &       0.581 &       0.493 & & 
0.208 &  \bf{0.116} & & 
0.589 & & 
0.713 &       0.120 &       0.788 & & 
0.661 & & 
0.636 &       0.420 &       0.506 &&  
0.804 &       0.703 &       0.750 & & 
0.647 \\ 

&
\SegPath  \cite{Mosinska19}  & &
0.627 &  \bf{0.738} &       0.515 & & 
\bf{0.313} &       0.257 & & 
\bf{0.693} & & 
\bf{0.940} &    0.331 &     0.782 & & 
0.658 & & 
0.421 &  \bf{0.571} &       0.485 &&  
0.621 &  \bf{0.857} &       0.720 & & 
0.619 \\ 

&
\DRoad{}  \cite{Mattyus17} & &
\bf{0.845} &    0.468 &     0.423 & & 
0.048 &       0.215 & & 
0.251 & & 
0.461 &       0.201 &       0.585 & & 
0.449 & & 
0.624 &       0.262 &       0.369 &&  
0.792 &       0.518 &       0.626 & & 
0.434 \\ 

&
\RCNN{}  \cite{Yang19}          & &
0.763 &       0.657 &  \bf{0.542} & & 
0.182 &       0.334 & & 
0.486 & & 
0.716 &       0.129 &       0.786 & & 
\bf{0.672} & & 
\bf{0.682} &       0.482 &  \bf{0.565} &&  
\bf{0.805} &       0.709 &  \bf{0.754} & & 
\bf{0.648} \\ 

\cmidrule{1-2} 	
\cmidrule{4-17} 	
\cmidrule{19-27}

\multirow{3}{*}{\rotatebox{90}{DGlobe}}&
\LinkN{}~\cite{Batra19} & &
0.778 &     0.803 &       0.653 & & 
0.632 &       0.107 & & 
0.660 & & 
0.699 &     0.180 &       0.755 & & 
0.735 & & 
0.599 &       0.781 &       0.678 &&  
0.743 &       0.789 &       0.766 & & 
0.716 \\ 

&
\MultiB{}~\cite{Batra19} & &
\bf{0.804} &       0.826 &  \bf{0.687} & & 
0.684 &  \bf{0.101} & & 
\bf{0.699} & & 
0.751 &   \bf{0.143} &  \bf{0.801} & & 
\bf{0.757} & & 
\bf{0.648} &       0.812 &  \bf{0.720} &&  
\bf{0.777} &       0.811 &  \bf{0.794} & & 
\bf{0.744} \\ 

&
\Segm{} \cite{Ronneberger15}    & &
0.545 &  \bf{0.841} &       0.495 & &  
\bf{0.720} &       0.138 & &  
0.618 & & 
\bf{0.925} &       0.458 &       0.683 & &  
0.675 & & 
0.394 &  \bf{0.874} &       0.543 &&  
0.532 &  \bf{0.863} &       0.658 & & 
0.658 \\ 

\cmidrule{1-2} 	
\cmidrule{4-17} 	
\cmidrule{19-27}
\end{tabular}
}
\caption{
Values of the existing and the new scores computed for road networks reconstructions by different methods on the RoadTracer and DeepGlobe datasets.
Our scores rank the methods much more consistently.
\label{tab:results}
}
\end{table*}


\begin{figure*}[t]
  \centering
  \setlength{\tabcolsep}{20pt}
	  \begin{tabular}{@{}lr@{}}	    
		\raisebox{-0.0cm}{\includegraphics[scale=0.6]{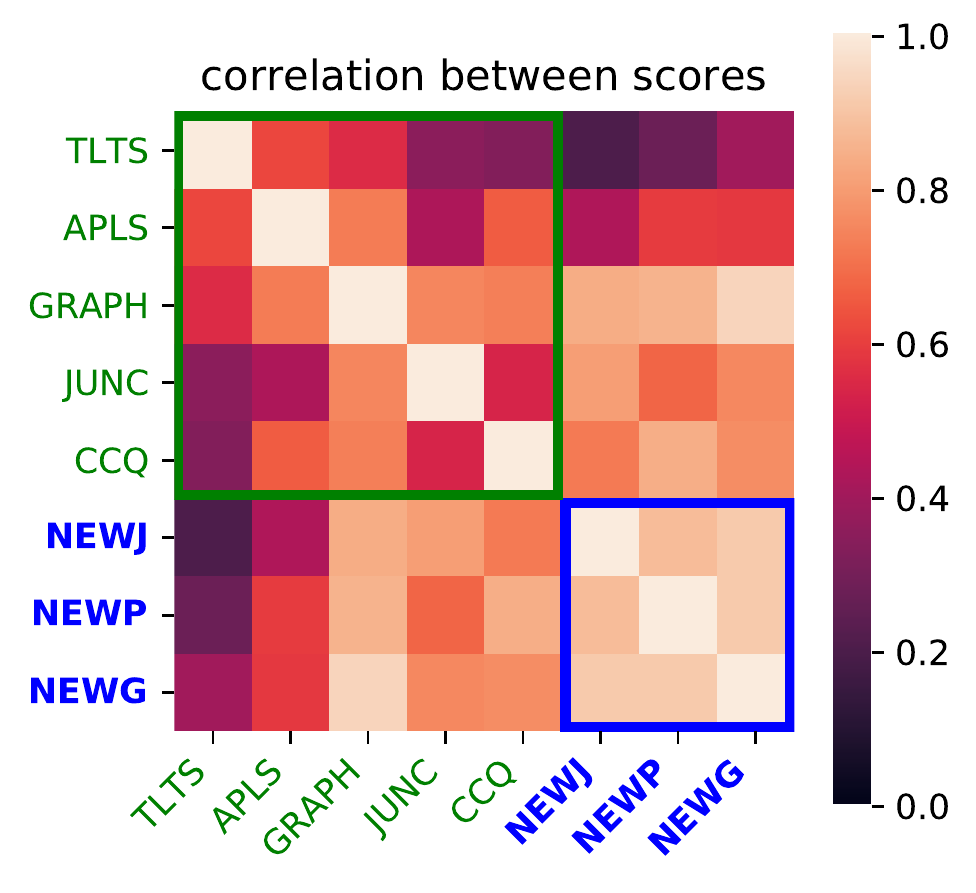}}   &
		\includegraphics[scale=0.6,trim={0cm, 0cm, 0cm, 0cm},clip]{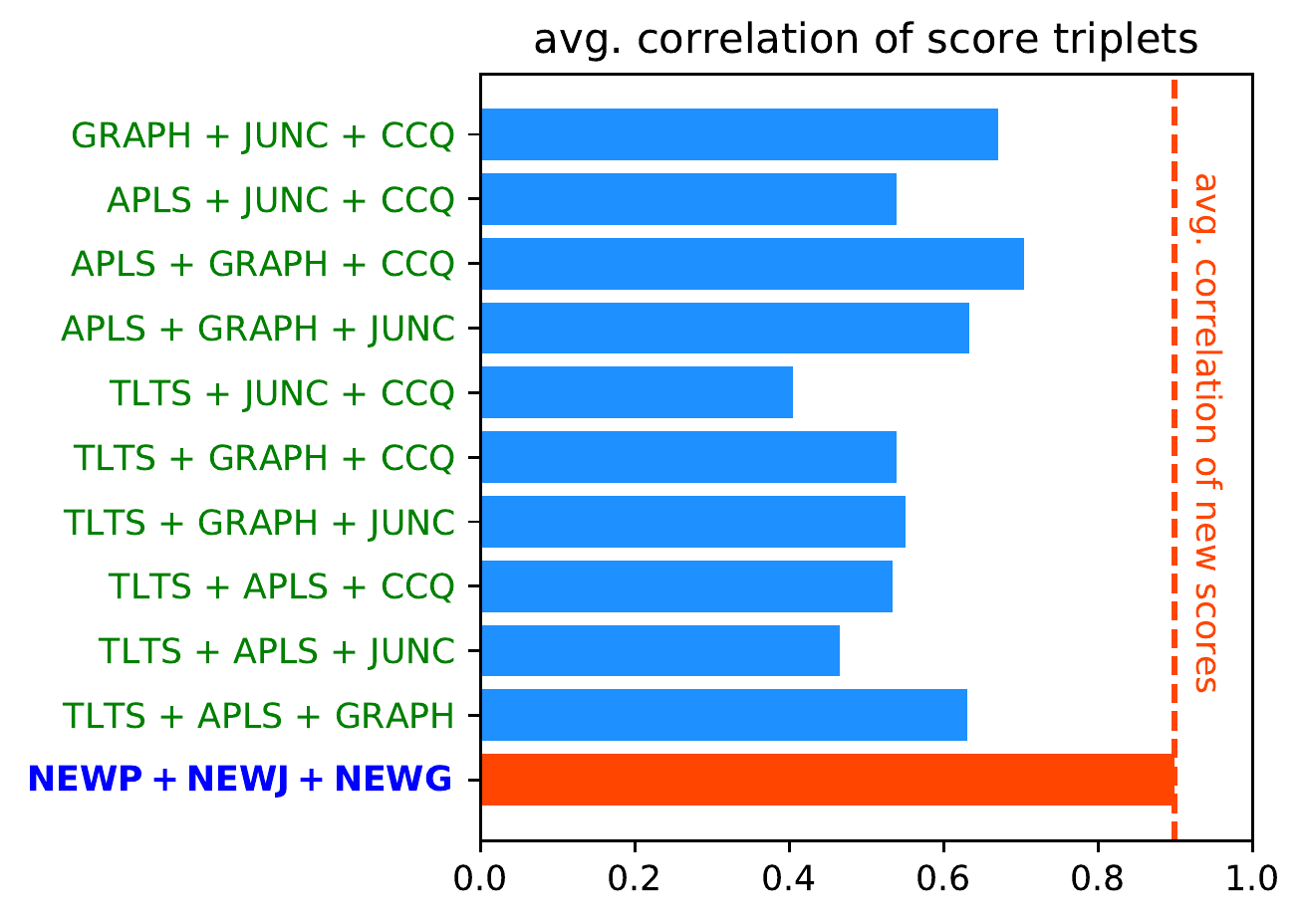} \\  
	  \end{tabular}
  \caption{
    \label{fig:correlation_plots_combined} 
Our three new scores correlate better than the existing scores.
\emph{Left}: A matrix of correlations of the scores computed for the maps reconstructed by different methods on the roadtracer dataset. The correlation coefficients of the old scores are outlined in green, the correlation coefficients of the new scores in blue.
\emph{Right}: The average correlation of all possible existing score triplets (blue bars) against the average correlation of the three new scores (dashed red line).
  }
\end{figure*}


\begin{figure}[t]
  \centering
  \setlength{\tabcolsep}{0pt}
  \renewcommand{\arraystretch}{0} 
  \setlength{\fboxsep}{0pt}
  \setlength{\fboxrule}{0pt}
	  \begin{tabular}{@{}r@{}}	    
		\fbox{\includegraphics[trim={0 0.8cm 0 0},clip,scale=0.5]{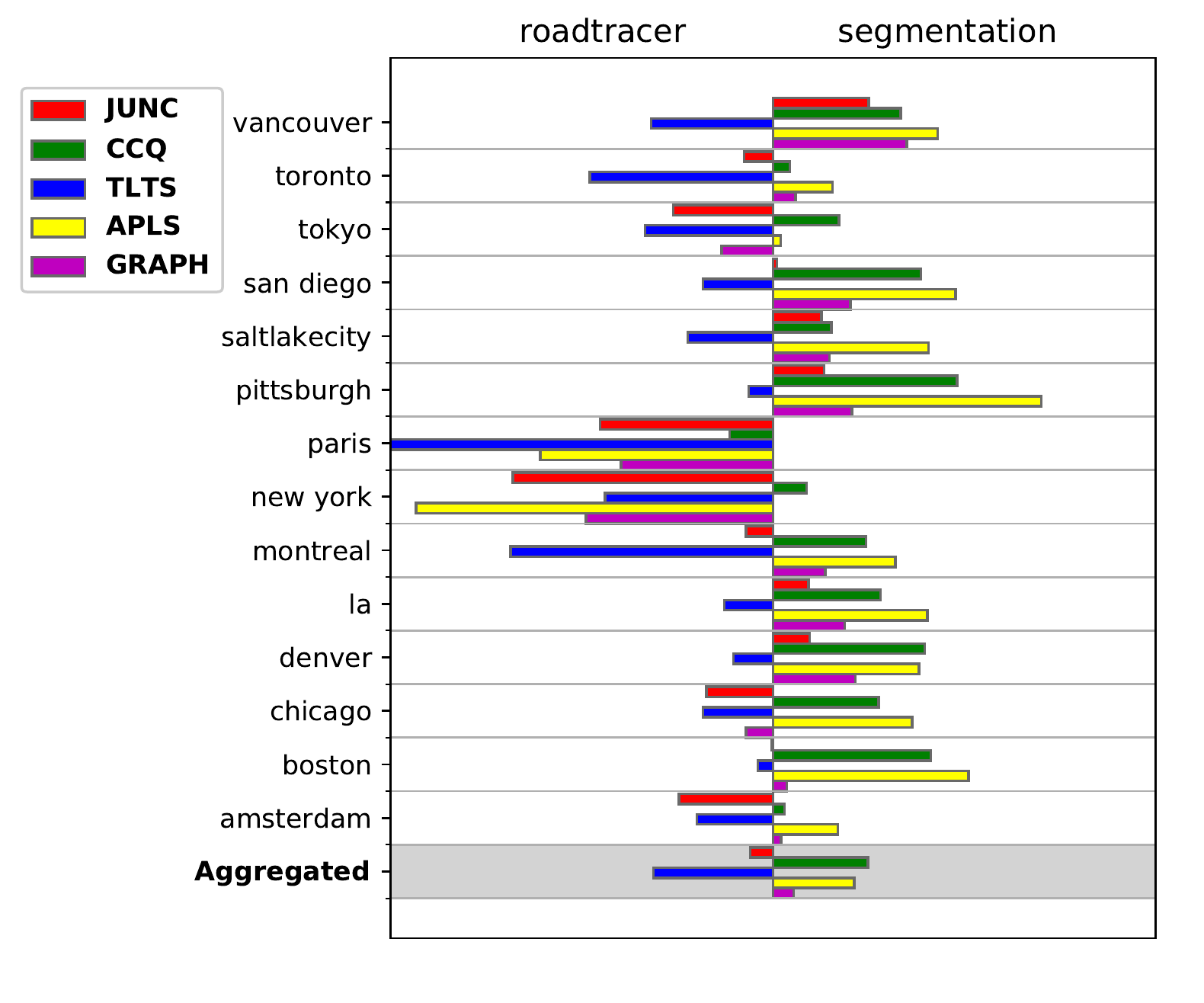}} 
                \\  
		\fbox{\includegraphics[trim={0 0 0 0.2cm},clip,scale=0.5]{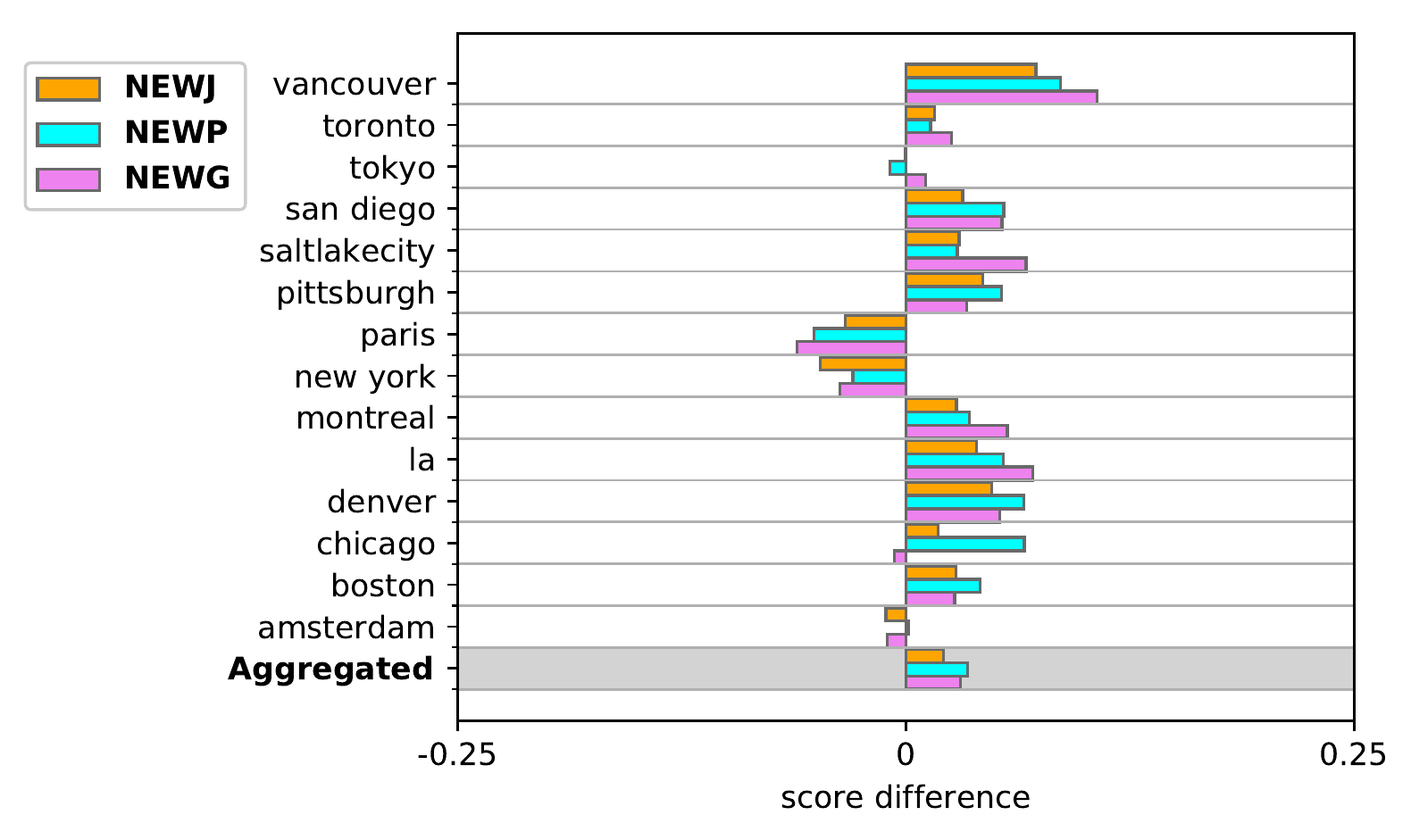}} \\  

	  \end{tabular}
  \caption{
    \label{fig:methods_plots} 
Differences of scores computed for selected cities from the RoadTracer test set.
Bars extending to the right express preference for the \Segm{}, ones extending to the left indicate the \RTracer{} scored higher.
  }
  
\end{figure}

\begin{figure}[t]
  \centering
  \includegraphics[trim=0.2cm 3.25cm 0.6cm 0, clip,width=\columnwidth]{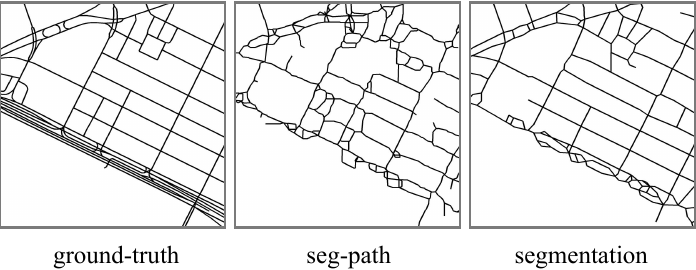}
  \includegraphics[trim=0.2cm 0.8cm 0.3cm 3.1cm, clip,width=\columnwidth]{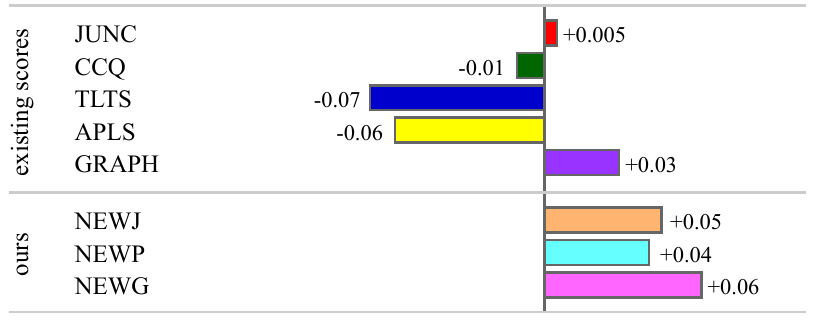}
\caption{
\emph{Top}: Crops of a road network of Pittsburgh and its reconstructions from aerial images by \emph{seg-path}~\cite{Mosinska19} and \emph{segmentation}.
\emph{Bottom}: Differences of the metrics for the two reconstructions. Bars that extend to the right favor \emph{segmentation}, ones that extend to the left favor \emph{seg-path}. 
\label{fig:visual_comp}
}
\end{figure}

In this section, we first use synthetic data to confirm that the flaws we have noted in existing metrics do have an impact on their ability to spot errors and that our new metrics are indeed more sensitive to those. 
We then compute the metrics for real data and observe that our metrics tend to be much more consistent than the existing ones, even though they follow very different approaches to attributing scores.

\subsection{Synthetic Data}
\label{sec:synth}

We created a synthetic benchmark dataset from selected crops of road networks from~\cite{Bastani18}.
We copied each crop to create pairs of `ground truth' and `predicted' networks and injected them with a controlled number of errors.
We selected the simplest errors that expose the faults of the scores.
\begin{itemize}

 \item Interruptions: Unwarranted breaks in roads.
 
 \item Overconnections: Spurious additional roads connecting randomly selected pairs of points.
 
 \item Perturbations: displacing graph nodes from their true locations without disconnecting the roads.
 
 \item Doubled roads: Spurious copies of road segments shifted slightly and connected to the originals. 
 
 \item Doubled roads-ground truth: Same as above, but the copies are added to the ground-truth. 
 
 \item False positives far away from true roads: To simulate them, we removed from the ground truth.

\end{itemize}
Fig.~\ref{fig:benchmark} depicts example graphs from our dataset. 
As shown in Fig.~\ref{teaser} and~\ref{fig:visual_comp}, similar errors appear in real reconstructions.
In Fig.~\ref{fig:benchmark_plots}, we plot the behavior of all the metrics as a function of the severity of each perturbation. 
As expected, each of the existing metrics is insensitive to at least one of them, while our metrics respond to all of them.

\subsection{Real Data}
\label{sec:real}

Instead of using synthetic data, we now turn to the recent road delineation algorithms, and analyze their predictions for the publicly available Roadtracer~\cite{Bastani18} and DeepGlobe~\cite{DeepGlobe18} datasets. 
We used algorithms implementations of which were made publicly available, and ones whose authors kindly shared with us the delineation results:
\begin{itemize}
 \item \Segm{}. Segmentation-based approach where the output probability map is thresholded and skeletonized. We use the prediction provided in \cite{Bastani18} for the Roadtracer dataset and our own implementation of UNet~\cite{Ronneberger15} for DeepGlobe.
 \item \RTracer{}. Iterative graph construction where node locations are selected by a CNN~\cite{Bastani18}.
 \item \SegPath{}. Unified approach to segmenting linear structures and classifying potential connections.~\cite{Mosinska19}
 \item \DRoad{}. Image segmentation followed by post-processing focused at fixing missing connections~\cite{Mattyus17}. 
 \item \RCNN{}. Recursive image segmentation with post-processing for graph extraction~\cite{Yang19}.
 \item \MultiB{}. A recursive architecture co-trained in road segmentation and orientation estimation~\cite{Batra19}.
 \item \LinkN{}. An encoder-decoder architecture~\cite{Chaurasia17} co-trained in segmentation and orientation estimation~\cite{Batra19}.
\end{itemize}
We present aggregated results in Table~\ref{tab:results}.
On the RoadTracer dataset, existing metrics favor \RTracer{}, \RCNN{}, or \SegPath{}. 
Such inconsistency would be fully justified if the scores were intended to measure different, possibly uncorrelated, qualities of interest.
This is however not the case. 
The scores serve as substitutes for a truly `bulletproof' method of comparing road networks, and are used for quantifying performance of competing algorithms.
Their inability to provide consistent evaluation makes reliable comparison of algorithms impossible.
By contrast, the proposed metrics consistently point to \RCNN{}.
Moreover, all of them rank \Segm{} second and \SegPath{} and \RTracer{} compete for the third place with very similar scores in all our metrics.
This level of consistency is reassuring -- the scores are computed using very different algorithms and their agreement attests to the reliability of the evaluation.
As seen in the bottom part of Table~\ref{tab:results} this also holds for the DeepGlobe data.
The existing scores are less inconsistent than for the RoadTracer dataset, with \TLTS{} favoring segmentation while other scores prefer \MultiB{}, but the new scores consistently prefer \MultiB{}. 
Note also the recurring patterns of precision- and recall-related scores for \JNEW{} and \PNEW{}.

It could be argued that the good agreement of our three measures is simply an artifact of selecting the right three measures. 
We show that this is not true by computing the correlations between pairs of scores on the RoadTracer dataset.
To evaluate consistency of score triplets, we average correlations of all pairs within a triplet.
We present the correlation matrix in the left part of Fig.~\ref{fig:correlation_plots_combined}.
In its right part we present the average correlations of all possible triplets of existing scores compared to the average correlation of our new scores.
The new scores clearly correlate better than the old ones. 

The RoadTracer dataset comprises images of several cities, with different street layout and appearance, and some methods might perform better in specific cities. 
Could our scores expose this difference?
To answer this question, we group the evaluation results for individual cities and expose them in Fig.~\ref{fig:methods_plots}. 
Due to space constraints, we only present the plots for \RTracer{} and \Segm{} here, and move the remaining plots to the supplement.
As can be seen at the top of the figure, the existing metrics favor different reconstructions of the same city and fail to highlight the difference clearly.
With the new metrics, the picture becomes much clearer, as shown as the bottom of the figure. 

In Fig.~\ref{fig:visual_comp}, we visualize fragments of two predicted networks, whose comparison by the existing scores is inconclusive,
but which are ranked consistently by our scores. We provide more examples in the supplement.


\section{Conclusion}
\label{sec:conclusion}

We were surprised to discover that \emph{all} the existing scores for evaluation of road network reconstructions suffer from design faults that make them insensitive to particular types of errors.
Our experiments show that the concerns this rises about the reliability of evaluation by means of these scores are justifiable -- one could overturn the results of a study by carefully selecting the score used for evaluation.
We have demonstrated that the flaws of existing metrics can be corrected -- our three new metrics are much more coherent than the old ones, despite the fact that each of them is computed in a different way.

We have focused on road network reconstructions, but the proposed scores can be used for comparing any curvilinear networks.
In future, we plan to apply them to the evaluation of reconstructions of vasculature from magnetic resonance angiography volumes and cell membranes from electrone microscopy images.

{\small
\bibliographystyle{ieee_fullname}
\bibliography{string,vision,learning,biomed,misc}

\begin{thebibliography}{10}\itemsep=-1pt

\bibitem{Ahmed15a}
M. Ahmed, B. Fasy, K. Hickmann, and C. Wenk.
\newblock A path-based distance for street map comparison.
\newblock {\em ACM Trans. Spatial Algorithms Syst.}, 1(1):3:1--3:28, July 2015.

\bibitem{Bajcsy76a}
R. Bajcsy and M. Tavakoli.
\newblock {Computer Recognition of Roads from Satellite Pictures}.
\newblock {\em IEEE Transactions on Systems, Man, and Cybernetics},
  SMC-6(9):623--637, 1976.

\bibitem{Bastani18}
F. Bastani, S. He, M. Alizadeh, H. Balakrishnan, S. Madden, S. Chawla, S.
  Abbar, and D. Dewitt.
\newblock {Roadtracer: Automatic Extraction of Road Networks from Aerial
  Images}.
\newblock In {\em Conference on Computer Vision and Pattern Recognition}, 2018.

\bibitem{Batra19}
A. Batra, S. Singh, G. Pang, S. Basu, C. Jawahar, and M. Paluri.
\newblock {Improved Road Connectivity by Joint Learning of Orientation and
  Segmentation}.
\newblock In {\em Conference on Computer Vision and Pattern Recognition}, June
  2019.

\bibitem{Biagioni12}
James Biagioni and Jakob Eriksson.
\newblock Inferring road maps from global positioning system traces.
\newblock {\em Transportation Research Record: Journal of the Transportation
  Research Board}, 2291:61--71, 12 2012.

\bibitem{Chaurasia17}
A. Chaurasia and E. Culurciello.
\newblock {Linknet: Exploiting Encoder Representations for Efficient Semantic
  Segmentation}.
\newblock {\em CoRR}, abs/1707.03718, 2017.

\bibitem{Cheng17}
G. Cheng, Y. Wang, S. Xu, H. Wang, S. Xiang, and C. Pan.
\newblock {Automatic Road Detection and Centerline Extraction via Cascaded
  End-To-End Convolutional Neural Network}.
\newblock {\em IEEE Trans. Geoscience and Remote Sensing}, 55(6):3322--3337,
  2017.

\bibitem{Chu19}
H. Chu, D. Li, D. Acuna, A. Kar, M. Shugrina, X. Wei, M. Liu, A. Torralba, and
  S. Fidler.
\newblock {Neural Turtle Graphics for Modeling City Road Layouts}.
\newblock In {\em International Conference on Computer Vision}, 2019.

\bibitem{DeepGlobe18}
I. Demir, K. Koperski, D. Lindenbaum, G. Pang, J. Huang, S. Basu, F. Hughes, D.
  Tuia, and R. Raskar.
\newblock {Deepglobe 2018: A Challenge to Parse the Earth through Satellite
  Images}.
\newblock In {\em Conference on Computer Vision and Pattern Recognition}, June
  2018.

\bibitem{VanEtten18}
A.~Van Etten, D. Lindenbaum, and T. Bacastow.
\newblock Spacenet: {A} remote sensing dataset and challenge series.
\newblock {\em CoRR}, abs/1807.01232, 2018.

\bibitem{Fischler81b}
M. Fischler, J.M. Tenenbaum, and H.C. Wolf.
\newblock {Detection of Roads and Linear Structures in Low-Resolution Aerial
  Imagery Using a Multisource Knowledge Integration Technique}.
\newblock {\em Computer Vision, Graphics, and Image Processing},
  15(3):201--223, March 1981.

\bibitem{Karagiorgou12}
S. Karagiorgou and D. Pfoser.
\newblock On vehicle tracking data-based road network generation.
\newblock In {\em Proceedings of the 20th International Conference on Advances
  in Geographic Information Systems}, SIGSPATIAL '12, pages 89--98, New York,
  NY, USA, 2012. ACM.

\bibitem{Li18f}
Y. Li, X. Zhang, and D. Chen.
\newblock {CSRNet: Dilated Convolutional Neural Networks for Understanding the
  Highly Congested Scenes}.
\newblock In {\em Conference on Computer Vision and Pattern Recognition}, 2018.

\bibitem{Mnih13}
V. Mnih.
\newblock {\em {Machine Learning for Aerial Image Labeling}}.
\newblock PhD thesis, University of Toronto, 2013.

\bibitem{Mnih10}
V. Mnih and G.E. Hinton.
\newblock {Learning to Detect Roads in High-Resolution Aerial Images}.
\newblock In {\em European Conference on Computer Vision}, pages 210--223,
  2010.

\bibitem{Mosinska19}
A. Mosi{\'n}ska, M. Kozinski, and P. Fua.
\newblock {Joint Segmentation and Path Classification of Curvilinear
  Structures}.
\newblock {\em IEEE Transactions on Pattern Analysis and Machine Intelligence},
  2019.

\bibitem{Mosinska18}
A. Mosi{\'n}ska, P. Marquez-Neila, M. Kozinski, and P. Fua.
\newblock {Beyond the Pixel-Wise Loss for Topology-Aware Delineation}.
\newblock In {\em Conference on Computer Vision and Pattern Recognition}, pages
  3136--3145, 2018.

\bibitem{Mattyus17}
G. {Máttyus}, W. {Luo}, and R. {Urtasun}.
\newblock {Deeproadmapper: Extracting Road Topology from Aerial Images}.
\newblock In {\em International Conference on Computer Vision}, pages
  3458--3466, 2017.

\bibitem{Quam78}
L.H. Quam.
\newblock {Road Tracking and Anomaly Detection}.
\newblock In {\em DARPA Image Understanding Workshop}, pages 51--55, May 1978.

\bibitem{Ronneberger15}
O. Ronneberger, P. Fischer, and T. Brox.
\newblock {{U-Net}: Convolutional Networks for Biomedical Image Segmentation}.
\newblock In {\em Conference on Medical Image Computing and Computer Assisted
  Intervention}, pages 234--241, 2015.

\bibitem{Vanderbrug76}
G. {Vanderbrug}.
\newblock Line detection in satellite imagery.
\newblock {\em IEEE Transactions on Geoscience Electronics}, 14(1):37--44, Jan
  1976.

\bibitem{Wegener05}
I. Wegener and R. Pruim.
\newblock {\em Complexity Theory: Exploring the Limits of Efficient
  Algorithms}.
\newblock Springer-Verlag, Berlin, Heidelberg, 2005.

\bibitem{Wegner13}
J.D. Wegner, J.A. Montoya-Zegarra, and K. Schindler.
\newblock {A Higher-Order CRF Model for Road Network Extraction}.
\newblock In {\em Conference on Computer Vision and Pattern Recognition}, pages
  1698--1705, 2013.

\bibitem{Wiedemann98}
C. Wiedemann, C. Heipke, H. Mayer, and O. Jamet.
\newblock {Empirical Evaluation of Automatically Extracted Road Axes}.
\newblock In {\em Empirical Evaluation Techniques in Computer Vision}, pages
  172--187, 1998.

\bibitem{Yang19}
X. {Yang}, X. {Li}, Y. {Ye}, R.~Y.~K. {Lau}, X. {Zhang}, and X. {Huang}.
\newblock Road detection and centerline extraction via deep recurrent
  convolutional neural network u-net.
\newblock {\em IEEE Transactions on Geoscience and Remote Sensing}, pages
  1--12, 2019.

\end{thebibliography}
}

\end{document}